\documentclass[final]{UD-Class}%[seceqn,secfloat,secthm]

\usepackage{graphicx}
\usepackage[sf,SF]{subfigure}

\usepackage{times}%
\usepackage{lastpage}
\usepackage[sort]{natbib}

\usepackage{url}
\usepackage{amssymb}

\usepackage{xspace}
\usepackage{xcolor}
\usepackage{setspace}
\usepackage{caption}

\usepackage{amsmath}
\usepackage{amssymb}
\usepackage[thmmarks]{ntheorem}

\usepackage{xcolor}
\usepackage{xspace}
\usepackage{color}

\newcommand{\mymodels}{\rotatebox[origin=c]{-180}{$\models$}}

\usepackage{algorithmic} % Check if correct package for  LLNCS
\usepackage{algorithm} % Check if correct package for  LLNCS

\usepackage{mathabx}
\usepackage{skull}
\usepackage{stmaryrd}
\usepackage{amsopn}

\usepackage{amsmath}
\usepackage{amssymb}
\usepackage[thmmarks]{ntheorem}
\newtheorem{Definition}{Definition}[section]
\newtheorem{Example}{Example}[section]

\usepackage[bookmarksopen=true,bookmarksnumbered=true,pdftitle={DesignSpace},pdfauthor={Mehul Bhatt},linkbordercolor=blue,urlcolor=red]{hyperref}

\firstpage{1}
\lastpage{\pageref{LastPage}}
\volume{}
\pubyear{2013}
\newcommand{\myurl}[1]{{\upshape\footnotesize\texttt{\url{#1}}}}

\newcommand{\key}[1]{{\emph{#1}}}

\newcommand{\OPRAm}[0]{\ensuremath{\mathcal{OPRA}_m}}

\newcommand{\bulf}{{\small$\blacktriangleright~$}}

\def\RelSpace{\ensuremath{\mathcal{R}}}

\newcommand{\msf}[1]{\ensuremath{\mathsf{\small#1}}}

\newcommand{\exts}[1]{\ensuremath{exists(\mathsf{\small#1})}}

\newcommand{\cons}[1]{\ensuremath{\llbracket #1 \rrbracket}}

\newcommand{\eod}{\hfill$\square$}

\begin{document}
\begin{frontmatter}                           % The preamble begins here.
%
%\pretitle{{\normalsize \emph{Mental Model Ascription by Language-Enabled Intelligent Agents}. \textbf{CogSci 2013}, Berlin, Germany}.}

\pretitle{{\normalsize\emph{ ISPRS International Journal of Geo-Information (ISSN 2220-9964); Special Issue on:\\Geospatial Monitoring and Modelling of Environmental Change}. \textbf{IJGI}\\{\upshape\footnotesize\texttt{\color{blue!50!black}http://www.mdpi.com/journal/ijgi/special\_issues/geospatial\_monitoring}}\\Editor: \emph{Duccio Rocchini}}}

\vspace{0.3in}

%\title{Between Sense and Sensibility}%\\
\title{Geospatial Narratives and their Spatio-Temporal Dynamics\thanks{This article is a draft of the review version submitted to the special issue of IJGI on Geospatial Monitoring and Modelling of Environmental Change.}}%\\

\subtitle{Commonsense Reasoning for High-level Analyses in Geographic Information Systems}

%\subtitle{Emerging Perspectives on Assistance for Architecture and Urban Design}

\runningtitle{Geospatial Narratives and their Spatio-Temporal Dynamics}

\medskip

\author{\bf\fnms{Mehul} \snm{Bhatt}}
%\thanks{Corresponding author: First Author, address of first author.}},
%and
%\author{\fnms{Christian} \snm{Freksa}}

\runningauthor{Bhatt, Wallgruen}
\address{Cognitive Systems (CoSy)\\
Spatial Cognition Research Center (SFB/TR 8)\\
University of Bremen, Germany\\ 
{\upshape\footnotesize\texttt{\color{blue}bhatt@informatik.uni-bremen.de}}\\
}

\author{\bf\fnms{Jan Oliver} \snm{Wallgruen}} %
%\thanks{Corresponding author: First Author, address of first author.}},
%and
%\author{\fnms{Christian} \snm{Freksa}}

\runningauthor{Bhatt, Wallgruen}

\address{GeoVISTA Center, Department of Geography,\\
Penn State University, USA\\ 
{\upshape\footnotesize\texttt{\color{blue}wallgruen@informatik.uni-bremen.de}}\\
}

\vfill

{\textbf{subject keywords}
$~$\\$~$\\\it\small computer science; cognitive science; artificial intelligence; cognitive systems; human-computer interaction; geography; environmental science; geospatial modelling; monitoring; remote sensing
}

\medskip

{\textbf{general keywords}
$~$\\$~$\\\it\small geographic information systems\sep spatio-temporal dynamics\sep computational models of narrative\sep events and objects in GIS\sep geospatial analysis\sep ontology\sep  qualitative spatial modelling and reasoning\sep spatial assistance systems\sep decision-support systems

}

\end{frontmatter}

$~$
\newpage

$~$

\begin{center}
\huge{\textsc{Abstract}}
\end{center}

\medskip

The modelling, analysis, and visualisation of dynamic geospatial phenomena has been identified as a key developmental challenge for next-generation Geographic Information Systems (GIS). In this context, the envisaged paradigmatic extensions to contemporary foundational GIS technology raises fundamental questions concerning the ontological, formal representational, and (analytical) computational methods that would underlie their spatial information theoretic underpinnings.\\\\We present the conceptual overview and architecture for the development of high-level semantic and qualitative analytical capabilities for dynamic geospatial domains. Building on formal methods in the areas of commonsense reasoning, qualitative reasoning, spatial and temporal representation and reasoning, reasoning about actions and change, and computational models of narrative, we identify concrete theoretical and practical challenges that accrue in the context of formal reasoning about \emph{space, events, actions}, and \emph{change}. With this as a basis, and within the backdrop of an illustrated scenario involving the spatio-temporal dynamics of urban narratives, we address specific problems and solutions techniques chiefly involving \emph{qualitative abstraction}, \emph{data integration and spatial consistency}, and \emph{practical geospatial abduction}.\\\\From a broad topical viewpoint, we propose that next-generation dynamic GIS technology demands a transdisciplinary scientific perspective that brings together Geography, Artificial Intelligence, and Cognitive Science.

\newpage

\section{\textsc{Introduction}}	

\noindent Geographic Information Systems (GIS) are confronted with massive quantities of micro and macro-level spatio-temporal data. In conventional GIS systems, this data takes the form of spatio-temporal databases of precise measurements pertaining to environmental features, aerial imagery, and more recently, sensor network databases that store real-time information about natural and artificial processes and phenomena. Within next-generation GIS systems, the fundamental information theoretic modalities are envisioned to undergo radical transformations: high-level ontological entities such as \key{objects}, \key{events}, \key{actions} and \key{processes} and the capability to model and reason about these is expected to be a native feature of next-generation GIS. Indeed, one of the crucial developmental goals in GIS systems of the future is a fundamental paradigmatic shift in the underlying `\key{spatial informatics}' of these systems.

\noindent \textbf{Time and GIS}.\quad Integrating time with GIS is necessary toward the development of GIS capable of monitoring and analysing successive states of spatial entities \citep{Claramunt_Theriault_95_Managing,DBLP:books/daglib/0020516,citeulike:10114141}. Such capability, necessitating the representation of instances of geographic entities and their change over time rather than change to layers or scenes, is the future of GIS and has been emphasized in the National Imagery and Mapping Agency's (NIMA; now the National Geospatial-Intelligence Agency (NGA)) vision for Integrated Information Libraries \citep{NIMA:2000:GIS-Vision}. A (temporal) GIS should, in addition to accounting for spatial changes, also consider the events behind changes and the facts which enable observation of these changes \citep{Beller:1991:ST-Events}. In the words of \citet{Claramunt_Theriault_95_Managing}:
	
\begin{quote}{%\small
`\emph{To respond adequately to scientific needs, a TGIS should explicitly preserve known links between events and their consequences. Observed relationships should be noted (e.g., entities $A$ and $B$ generate entity $C$) to help scientists develop models that reproduce the dynamics of spatio-temporal processes. Researchers will thus be able to study complex relationships, draw conclusions and verify causal links that associate entities through influence and transformation processes}'.}
\end{quote}

\noindent Clearly, this facility necessitates a formal approach encompassing events, actions and their effects toward representing and reasoning about dynamic spatial changes. Such an approach will be advantageous in GIS applications concerned with retrospective analysis or diagnosis of observed spatial changes involving either fine-scale object level analysis or macro-level (aggregate) analysis of dynamic geospatial phenomena. For instance, within GIS, spatial changes could denote (environmental) changes in the geographic sphere at a certain temporal granularity and could bear a significant relationship to natural events and human actions, e.g., changes in land-usage, vegetation, cluster variations among aggregates of demographic features, and wild-life migration patterns. 

\smallskip

\noindent \textbf{Geospatial Semantics}.\quad Conceptual models for representing geospatial events and processes in general have been the focus of extensive research efforts in the last decade. Research in the area of geospatial semantics, taxonomies of geospatial events and processes, and basic ontological research into the nature of processes in a specific geospatial context has garnered specific interest from several quarters \citep{Grenon_Smith_04_SNAP,DBLP:conf/giscience/WorboysH04,Renolen_00_Modelling,journals/ao/GaltonM09,Stewart-Hornsby_Cole_07_Modeling}. Fundamental epistemological aspects concerning, for instance,  event and object identity have received special attention in the community \citep{DBLP:conf/kr/Bennett02,Hornsby_Egenhofer_00_Identity}. This has mainly been spurred by the realization that purely snapshot-based temporal GIS does not provide for an adequate basis for analyzing spatial events and processes and performing spatio-temporal reasoning. Event-based and object-level reasoning at the spatial level could serve as a basis of explanatory analyses within a GIS \citep{galton04-interpolation,journals/jucs/MondoSCT10,journals/gis/Worboys05,Couclelis-Cosit09}. 
For instance, a useful reasoning mechanism that applications may benefit from could be the task of causal explanation, which is the process of retrospective analysis by the extraction of an event-based explanatory model from available spatial data. Indeed, the explanation would essentially be an event-based history of the observed spatial phenomena defined in terms of both domain-independent and domain-dependent occurrences.

\noindent \textbf{Narrative as a Model of Perceptual Sense Making}. Researchers in computational logics of action and change have interpreted narratives in several ways (e.g., in the context of formalisms such as the situation calculus and event calculus) \citep{MillerS94,OccurNarraSC:Pinto:1998,DBLP:journals/lalc/Mueller07,McCarthy-98-kr-combining-narrative,McCarthy:2000:concepts-logical-AI}:

\begin{quote}{\small
``\emph{a sequence of events about which we may have incomplete, conflicting or incorrect information}'' %\hfill \citep{MillerS94,OccurNarraSC:Pinto:1998}

``\emph{accounts of sets of events, not necessarily given as sequences; a narrative is an account of what happened}'' %\hfill \citep{McCarthy-98-kr-combining-narrative}
}
\end{quote}

The significance of \emph{narratives} in everyday discourse, interpretation, interaction, belief formation, and decision-making has been acknowledged and studied in a range of scientific, humanistic, and artistic disciplines. Narrativisation of everyday perceptions by humans, and the significance of narratives, e.g., in communication and interaction, has been investigated under several frameworks, and through several interdisciplinary initiatives involving the arts, humanities, and social sciences, e.g., the narrative paradigm \citep{narrative-paradigm}, narrative analysis \citep{narrat-analysis}, narratology \citep{narratolog-prince-1982,narratology-marie-laure,meister-narratology-hdbk-narrato}, discourse analysis and computational narratology \citep{Roland-1975-narrative-structural,CMN-Mani-2012,Mani-comp-narratology,Goguen-course-compu-narratology}. %The study of narratives has attracted attention from several quarters, most prominently in disciplines such as literature, linguistics, anthropology, semiotics, cultural studies, geography, psychology, cognitive science, logic, and computer science. 

%The study of narratives has attracted attention from several quarters, most prominently in disciplines such as literature, linguistics, anthropology, semiotics, cultural studies, geography, psychology, cognitive science, logic, and computer science. 

We regard narratives and high-level processes of (computational) narrativisation emanating therefrom as a general underlying structure serving the crucial function of \emph{perceptual sense-making} --- i.e.,  as a link between problem-specific perceptual sensing (i.e., data) and the (computational) formation of sensible impressions concerned with interpretation and analytical tasks. The particular form of the proposed narrative structure is that of cognitively inspired \emph{computational model of narrativisation} involving high-level commonsense reasoning with \emph{space, events, actions, change, and interaction} \citep{Bhatt:RSAC:2012}. We posit that computational narrativisation pertaining to space, actions, and change provides a useful model of  \emph{visual} and \emph{spatio-temporal thinking} within a wide-range of problem-solving tasks and application areas, with geospatial dynamics being the focus of this paper.

Computational models of geospatial narratives therefore, by definition, are aimed at making sense of massive quantities of micro and macro-level spatio-temporal data pertaining to environmental, socio-economic and demographic processes operating in a geospatial context. Such narratives are constructed on the basis of spatio-temporal databases of precise measurements about environmental features, aerial imagery, sensor network databases with real-time information about natural and artificial processes and phenomena etc. Geospatial narratives typically span a temporal horizon encompassing generational change, but these could also pertain to the scale of everyday `\emph{life in the city}', natural environmental processes, etc.

% \citep{Bhatt_Wallgruen_11_Analytical}

% \citet{Grenon_Smith_04_SNAP} introduced the distinction between the SNAP ontological perspective concerned with continuants and conceptualizing spatio-temporal knowledge as a set of snapshots indexed by time, and the SPAN perspective concerned with occurrents depicting reality as processes unfolding in space and time. 

%\section*{Citations}
%
%\begin{itemize}
%%
%	\item - Helen C.
%%	
%	\item \noindent \citep{Chaudet:2006:EEC} -- \cite{Jiang_Worboys_09_Event}
%%
%%	
%\end{itemize}
%
%

\noindent \textbf{Narratives and High-Level Analytical Interpretation in GIS}.\quad The core development goal for constructing a narrative-centred representational and computational apparatus for next-generation GIS is indeed aimed at \emph{making-sense} of enormous quantities geographic data. This has to be done in order to empower analysts and decision-makers at all levels of the socio-political and economic policy-making hierarchy in both public as well as private spheres.

Advances in formal methods in the areas of commonsense reasoning, qualitative spatio-temporal representation and reasoning, reasoning about space, actions and change, and spatio-temporal dynamics \citep{Bhatt:RSAC:2012,bhatt2011-scc-trends,Cohn_Renz_07_Qualitative} provide interesting new perspectives for the development of the foundational spatial informatics underlying next-generation GIS systems. The basic requirements within these systems encompass:

\begin{itemize}

	\item \emph{Knowledge engineering, semantics, and modelling}: Introduction of the capability to include \key{object}, \key{event} and \key{process} based abstractions of spatio-temporal phenomena as native, first-class entities, enshrined with rich semantic characterizations within the ontology and conceptual model of the GIS system in a manner that is interoperable across systems and implementations.

	\item \emph{Analytical reasoning}: From a computational viewpoint, generic \key{high-level reasoning mechanisms} that leverage upon the semantics of the formally modelled or axiomatised properties of domain-independent and dependent aspects are necessary. These mechanisms could be used to ground and model environmental (natural and human) phenomena from domains such as \emph{epidemiology, urban dynamics, vegetation monitoring, wild-life biology, transportation dynamics, cultural heritage} and so forth (Section \ref{sec:application-areas}).
\end{itemize}
	
Indeed, it is expected that these knowledge representation and reasoning capabilities will provide a basis for high-level analytical and decision-making tasks, either individually or in conjunction with other forms of analytical techniques from the field of spatial statistics, or quantitative analysis in GIS.

%BeyonDeparture from State of the Art
\subsection*{Contributions and Organization of the Paper.}\quad This paper aims to bring formal methods concerning Knowledge Representation and Reasoning (KR) into the domain of Geographic Information Systems. In this context, and with a particular focus on the use of KR-based formal commonsense reasoning methods, this paper:

\begin{itemize}
	\item demonstrates basic representation and computational challenges pertaining to \emph{space, actions, and change}
	\item presents an overarching framework for high-level modelling and (explanatory) analysis for the geospatial domain, and
	\item addresses concrete representational and computational problems that accrue in this context and provides a unified view of a consolidated architecture in the backdrop of an illustrated application scenario from the domain of urban dynamics.

 \end{itemize}
 
% \commentJOW{departure from state of the art to be added here}
%
%\begin{enumerate}
%	\item QSR - departure
%	
%	\item Commonsense departure
%	
%	\item Etc
%\end{enumerate}
%

\noindent \emph{The paper is organised as follows}:\quad 

{ 

\medskip

\textbf{Section \ref{sec:geo-dynamics-apps}} presents application-guided perspectives from several domains where the notion of geospatial dynamics is recognised as being applicable; we also provide two concrete motivating scenarios concerning the spatio-temporal dynamics of urban narratives. 

\smallskip

\textbf{Section \ref{sec:csc}} presents a brief overview of formal methods in commonsense, qualitative spatial representation and reasoning, and may be skipped by readers familiar with the topic.

\smallskip

\textbf{Section \ref{sec:spat-informatiks}} presents an intuitive overview of the core spatial informatics---representational and computational challenges---that accrue whilst modelling and reasoning with dynamic geospatial phenomena. 

\smallskip

\textbf{Section \ref{sec:geo-formal-framework}} contextualizes the dynamic geospatial spatial informatics by way of a consolidated framework: We describe the overall architecture  and its components using a running example, and  illustrate how basic representational and computational challenges may be met within the formal theory of space, events, actions and change. 

\smallskip

\textbf{Section \ref{sec:disc-outlook}} concludes the paper with a  discussion of our research perspective and summary of contributions.

}

\section{\textsc{Geospatial Dynamics: Application Perspectives}}\label{sec:geo-dynamics-apps}

\noindent In recent years, modelling and analysis of dynamic geospatial phenomena and the integration of time in GIS have emerged as major research topics within the GIS community. Although at present the representational and analytical apparatus to examine the dynamics of such phenomena is nascent at best, the issue has been considered as a major research priority in GIS \citep{RschAgendaUCGIS:2004}. 

Here, we briefly indicate a select range of domains where the notion of geospatial dynamics is applicable, and also provide motivating scenarios from the field of urban dynamics and environmental development.

\subsection{Application Areas}\label{sec:application-areas}

\noindent A wide-range of priority areas where high-level analytical ability is crucial come to the fore:

\noindent\bulf \textbf{Epidemiology}. \quad This is a classic application domain, which from a spatial perspective, involves the study of diffusive processes (e.g. spread of disease) with either point-based or aggregate entities in space and time.

\noindent\bulf \textbf{Moving Data Analysis}.\quad This domain involves the analysis of (typically people-centered) motion data for purposes of prediction and explanation. For instance, studies involving vehicles / people trajectories,  transportation data, crime statistics have found significant attention in this area.

\noindent\bulf \textbf{Land-use Analysis}.\quad This corresponds to the analysis of land-use patterns, e.g., in urban areas, on the basis of remote sensing and other ground data. For instance, one objective here could be to study the nature of land-use dynamics either together or in isolation with data involving socio-economic dimensions.

\noindent\bulf \textbf{Disaster Management}.\quad This corresponds to \emph{assistive technologies} that provide managerial and analytical capabilities both before / after and in times of natural and man-made calamities (e.g., fire, flooding, hurricanes, tornado, landslides, earthquakes).

\noindent\bulf \textbf{Environmental Modelling, Wildlife Biology}.\quad These domains involve modelling and analysis of environmental phenomena at the ecological level, e.g., integrated systems and relationships involving flora and fauna. Typical studies involve vegetation monitoring (e.g., forestry / deforestation), climatic change (e.g., glaciers, sea level change), and monitoring of pollution, soil, air quality, water quality etc.

\noindent\bulf \textbf{Archeology, Cultural Heritage}.\quad GIS technology is employed
by archaeologists to reconstruct historical events and developments as well as 
to predict sites of potential archeological interest. Resulting archeological records
are made accessible to the public in the form of cultural heritage portals. To facilitate
intuitive access to cultural heritage information, for instance by
tourists, a spatio-temporal ontology of changes in political and administrative regions is required.

\subsection{Urban Narratives, and their Spatio-Temporal Dynamics: An Example}\label{sec:urban-dynamics} % as a Narrative

%Urban Narratives \textbf{TODO}

Urbanization and high-level narratives of urban dynamics can be interpreted with respect to the sum total of a range of demographic, environmental (both natural and artificial), sociological, and economic processes. Indeed, urban dynamics, and `\emph{the urban narrative}'  may not be trivialised as being strictly as such, but for the present discussion, this interpretation suffices.

\medskip

\begin{figure}%[th]
	\center
	\subfigure[1984]{
		\label{fig:vegas1}
		\scalebox{0.22}{\includegraphics{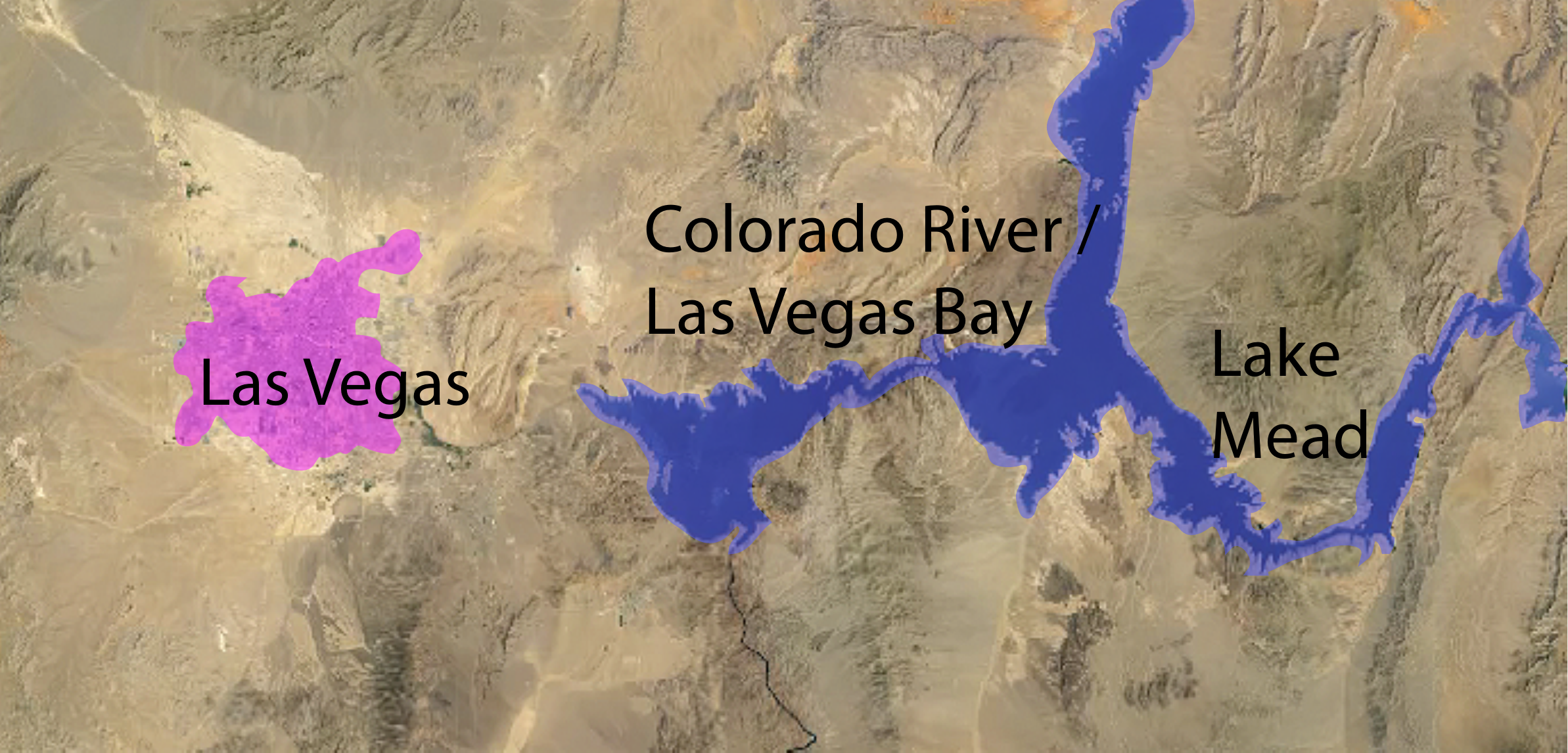}}}
%		\subfigure[1986]{
%		\label{fig:vegas2}
%		\scalebox{0.18}{\includegraphics{Google-Time-Lapse/Growth-Las_Vegas/Raw-Edited/Vegas-1986-2.pdf}}}
		\subfigure[1989]{
		\label{fig:vegas3}
		\scalebox{0.22}{\includegraphics{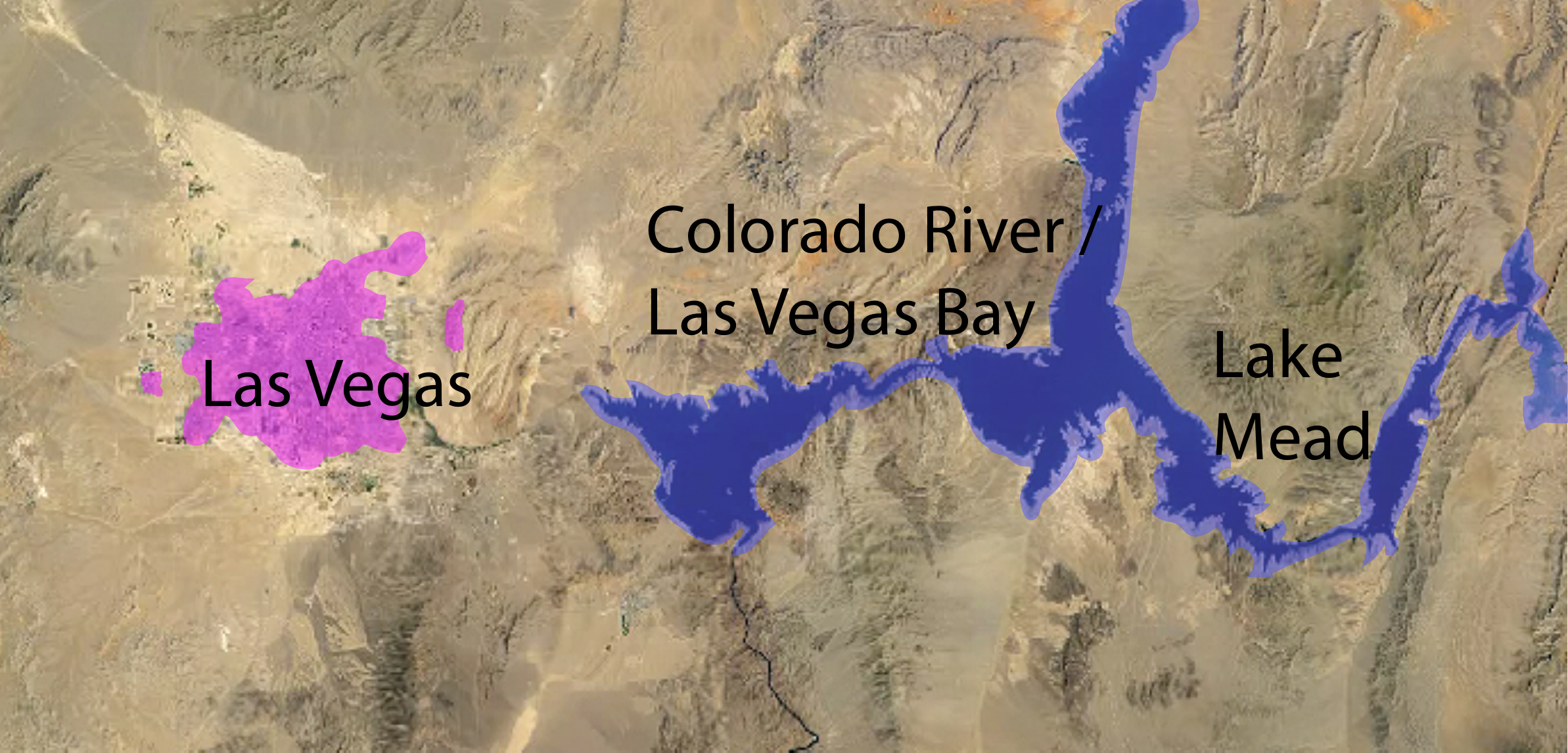}}}
		\subfigure[1994]{
		\label{fig:vegas4}
		\scalebox{0.22}{\includegraphics{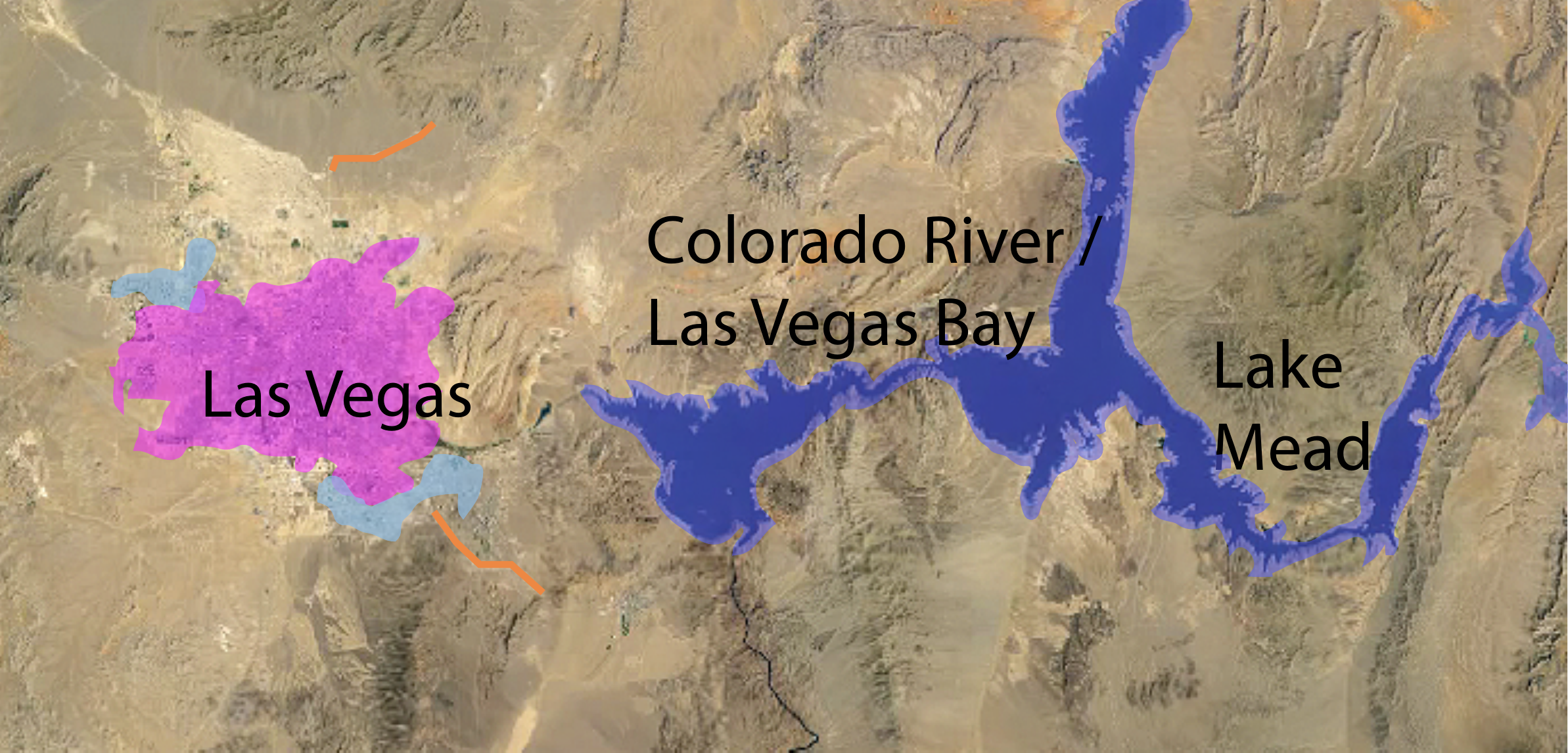}}}
\subfigure[2000]{
		\label{fig:vegas5}
		\scalebox{0.22}{\includegraphics{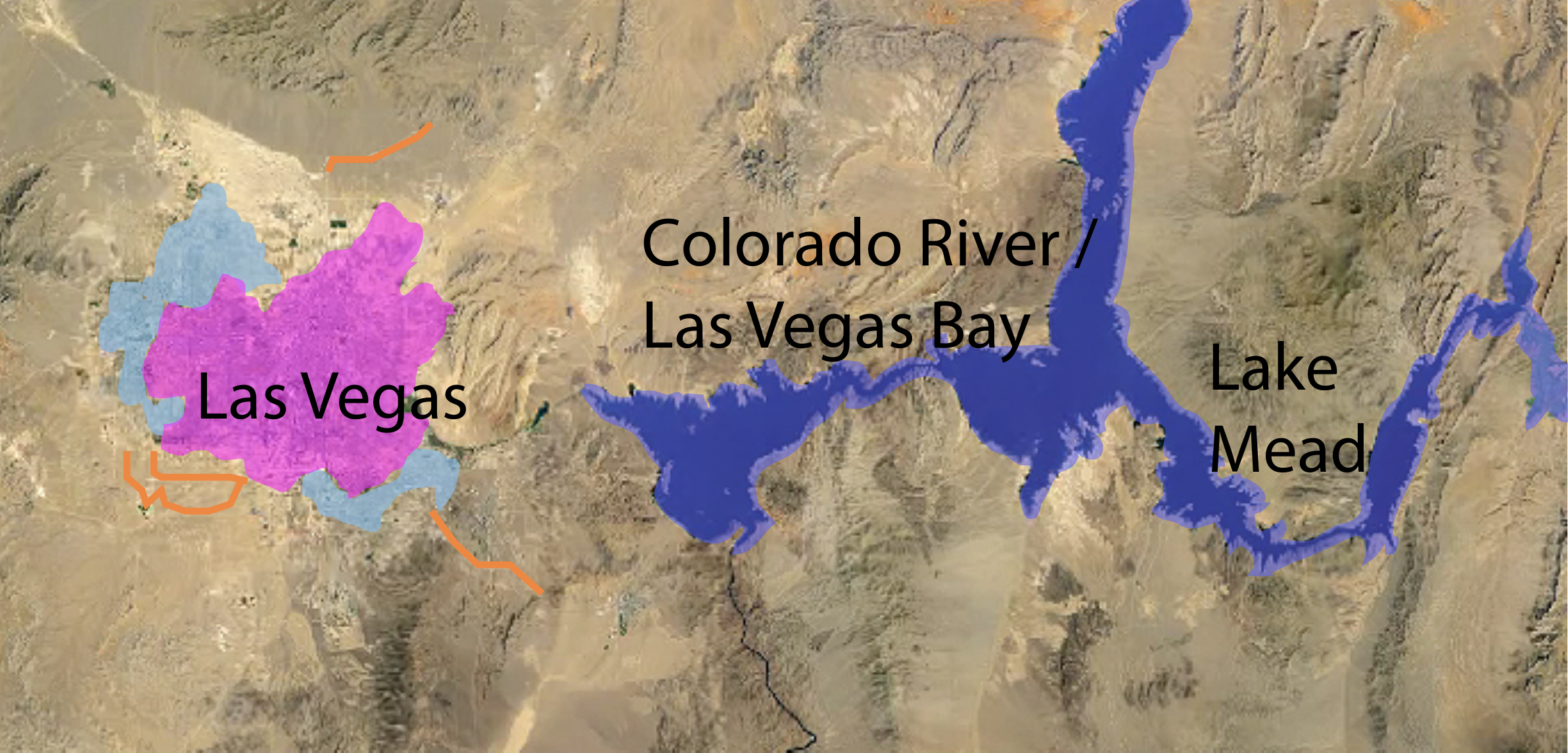}}}
%		\subfigure[2002]{
%		\label{fig:vegas6}
%		\scalebox{0.18}{\includegraphics{Google-Time-Lapse/Growth-Las_Vegas/Raw-Edited/Vegas-2002-5.pdf}}}
		\subfigure[2004]{
		\label{fig:vegas7}
		\scalebox{0.22}{\includegraphics{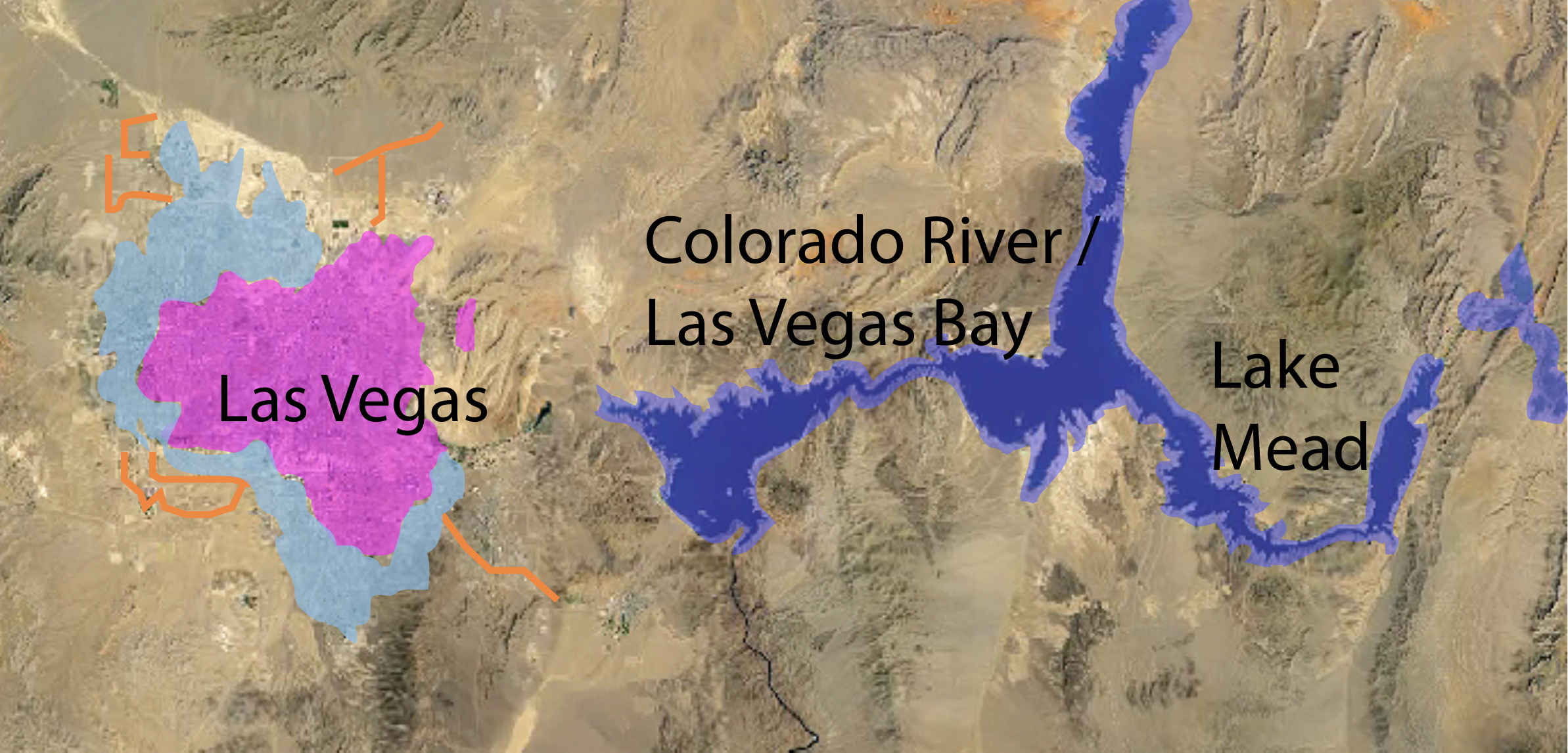}}}
		\subfigure[2010]{
		\label{fig:vegas8}
		\scalebox{0.22}{\includegraphics{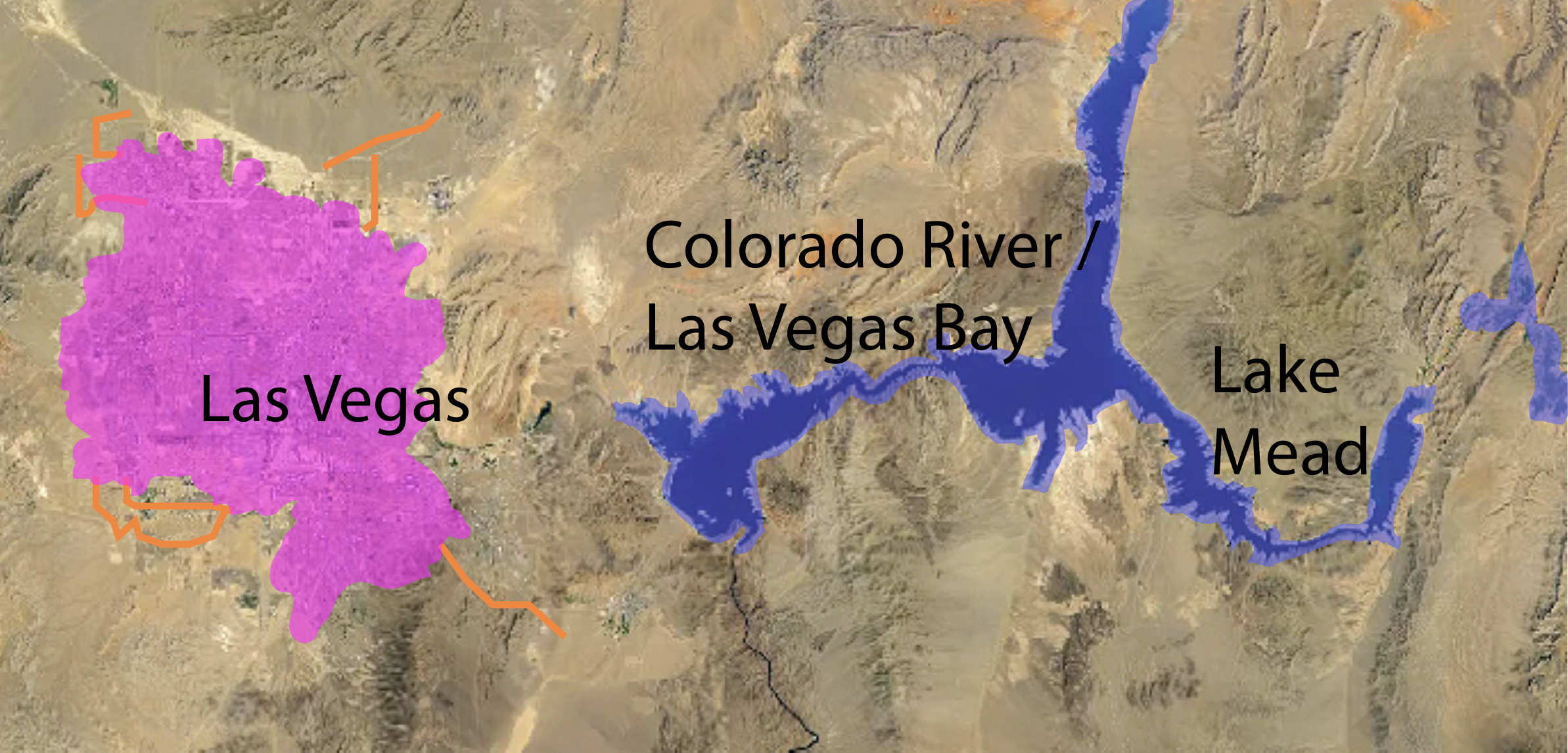}}}
	\caption{Urban expansion in Las Vegas, Texas, USA}
	\label{fig:vegas}
\end{figure}

\begin{figure}%[th]
	\center
	\subfigure[1984]{
		\label{fig:dubai1}
		\scalebox{0.19}{\includegraphics{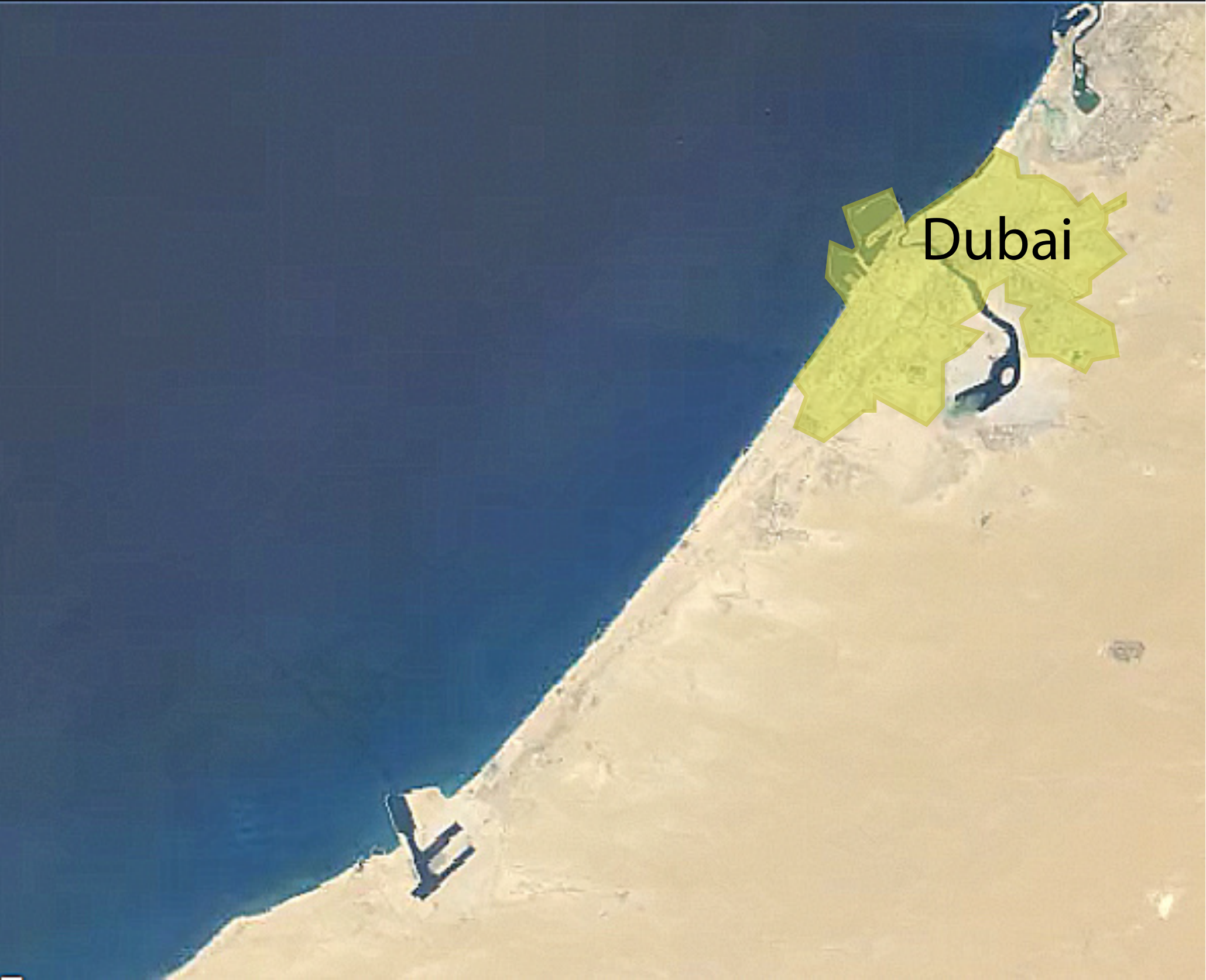}}}
	%\subfigure[1992]{
	%	\label{fig:dubai2}
	%	\scalebox{0.2}{\includegraphics{Google-Time-Lapse/Dubai-Coastal-Expansion/Raw-Edited/Dubai-Coastal-1992_mod.pdf}}}
	\subfigure[1998]{
		\label{fig:dubai3}
		\scalebox{0.19}{\includegraphics{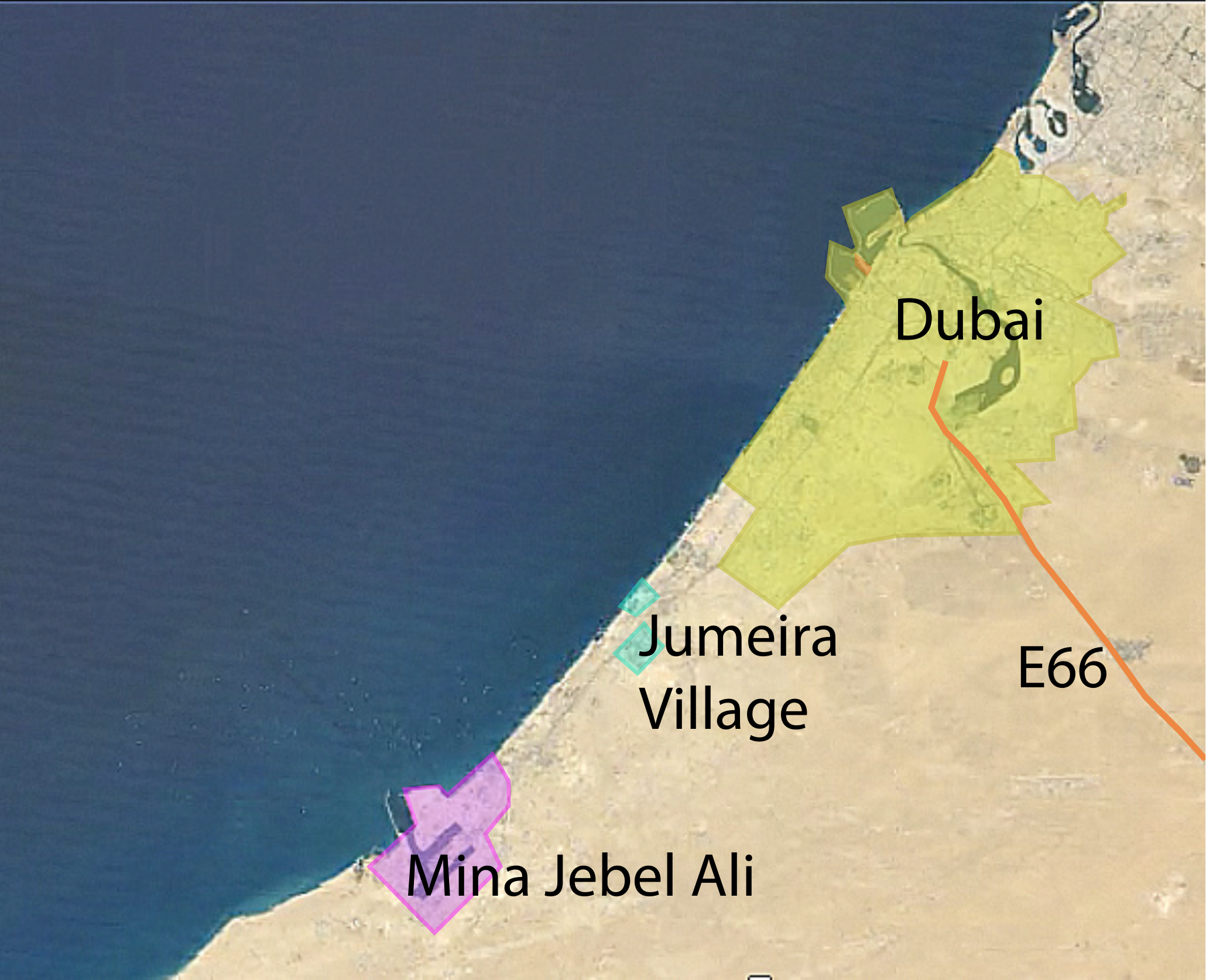}}}
	%	\subfigure[2001]{
	%	\label{fig:dubai4}
	%	\scalebox{0.2}{\includegraphics{Google-Time-Lapse/Dubai-Coastal-Expansion/Raw-Edited/Dubai-Coastal-2001_mod.pdf}}}
		\subfigure[2002]{
		\label{fig:dubai5}
		\scalebox{0.19}{\includegraphics{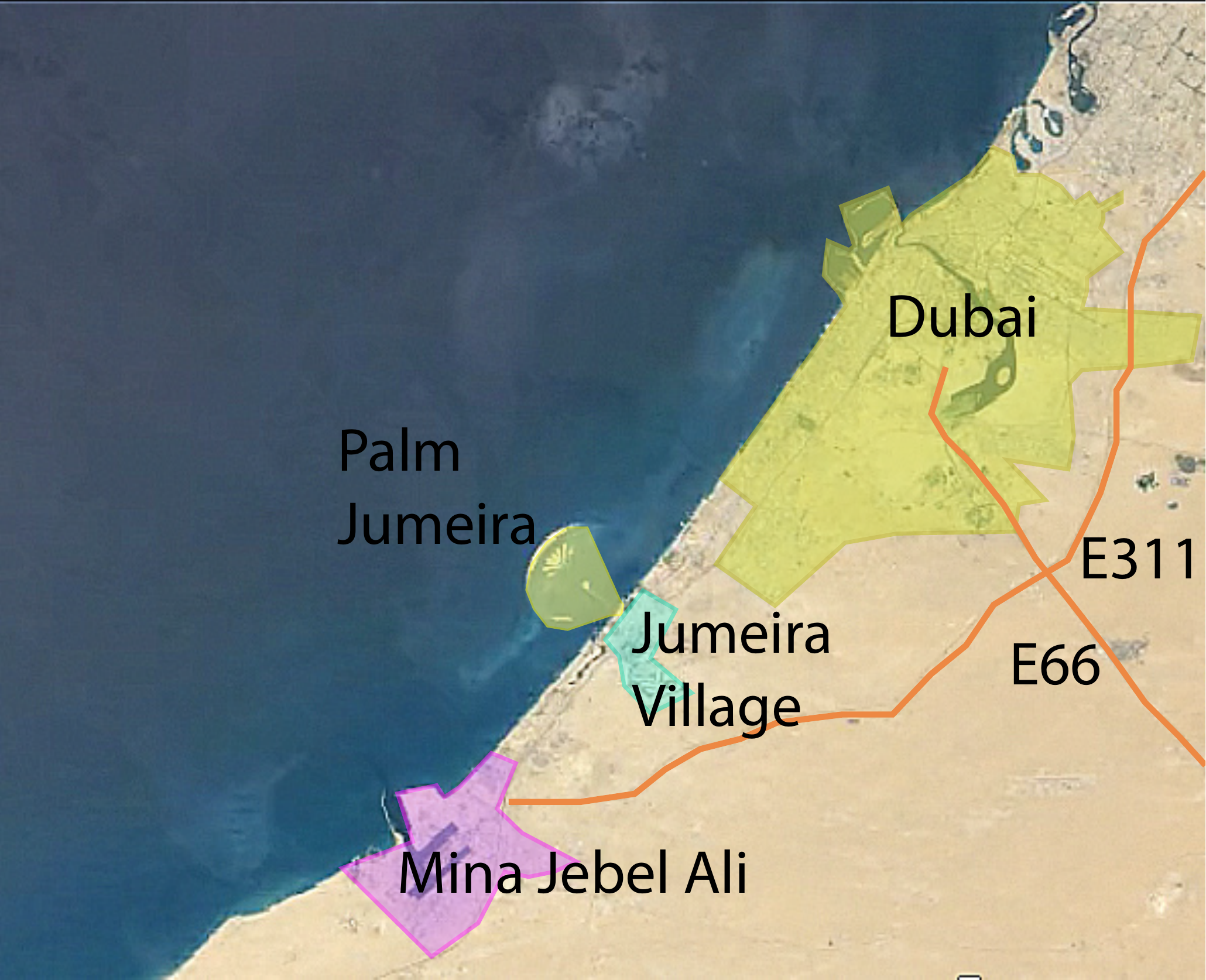}}}
	%	\subfigure[2003]{
	%	\label{fig:dubai6}
	%	\scalebox{0.2}{\includegraphics{Google-Time-Lapse/Dubai-Coastal-Expansion/Raw-Edited/Dubai-Coastal-2003_mod.pdf}}}
\subfigure[2004]{
		\label{fig:dubai7}
		\scalebox{0.19}{\includegraphics{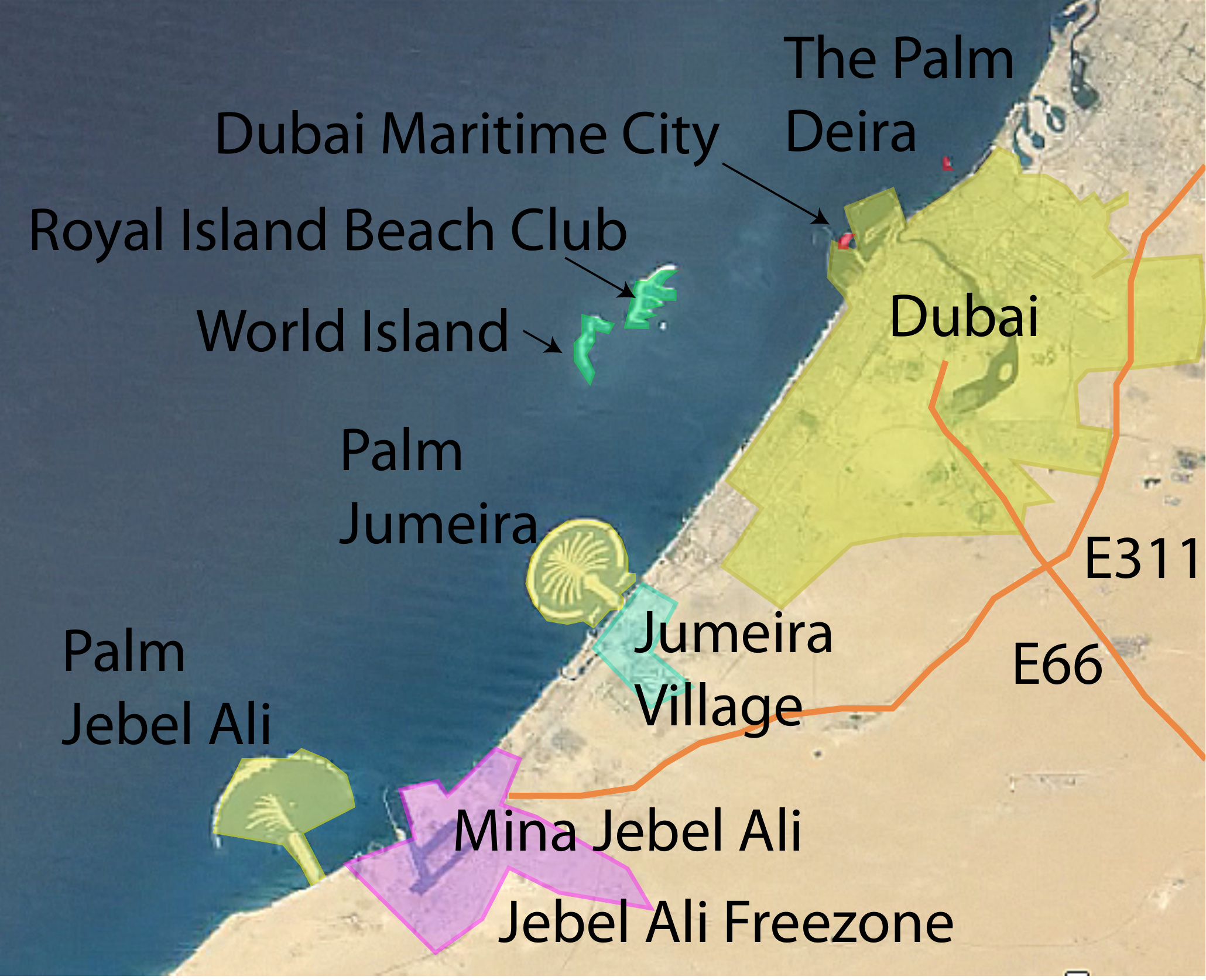}}}
		\subfigure[2005]{
		\label{fig:dubai8}
		\scalebox{0.19}{\includegraphics{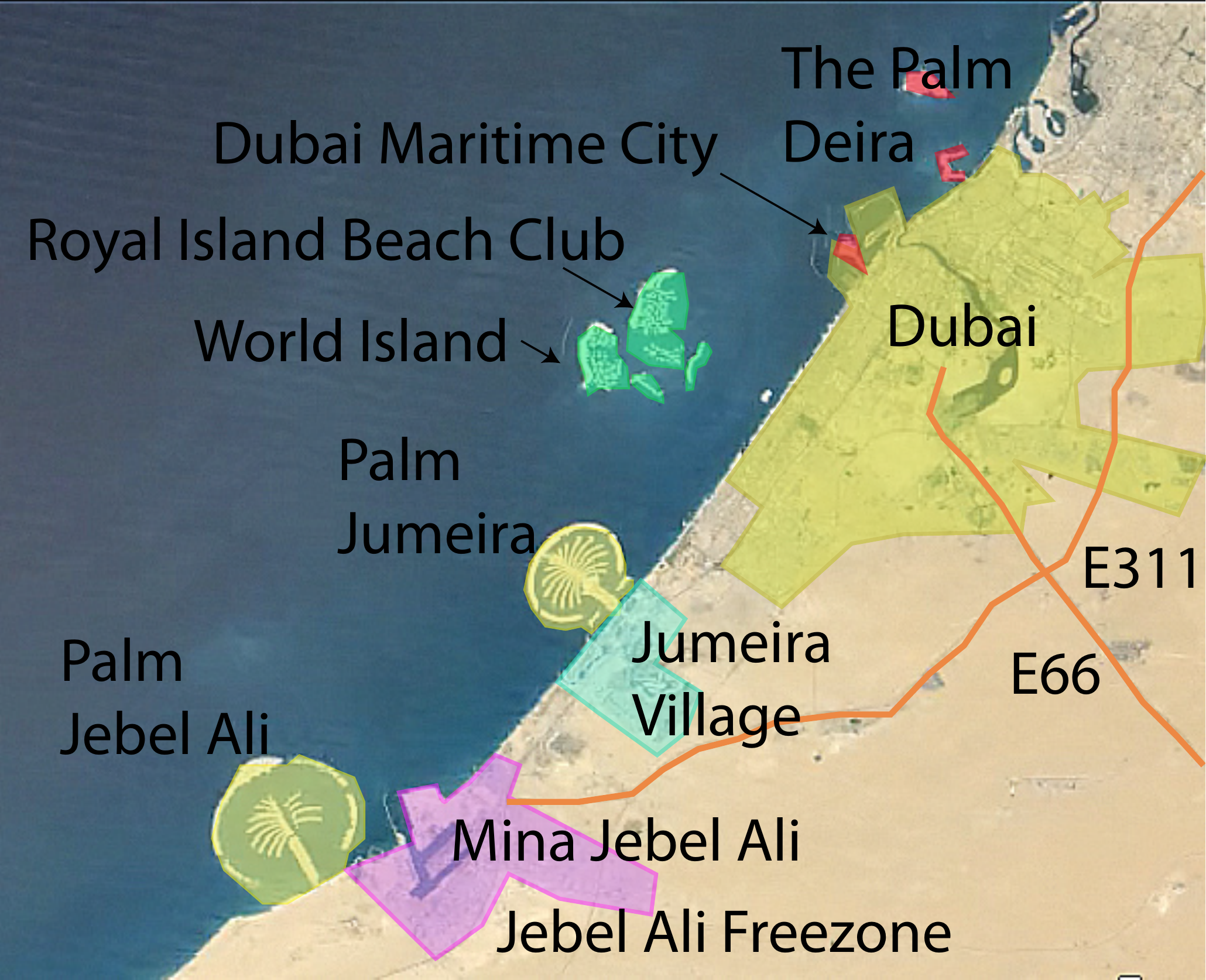}}}
	%	\subfigure[2006]{
	%	\label{fig:dubai9}
	%	\scalebox{0.2}{\includegraphics{Google-Time-Lapse/Dubai-Coastal-Expansion/Raw-Edited/Dubai-Coastal-2006_mod.pdf}}}
		\subfigure[2007]{
		\label{fig:dubai10}
		\scalebox{0.19}{\includegraphics{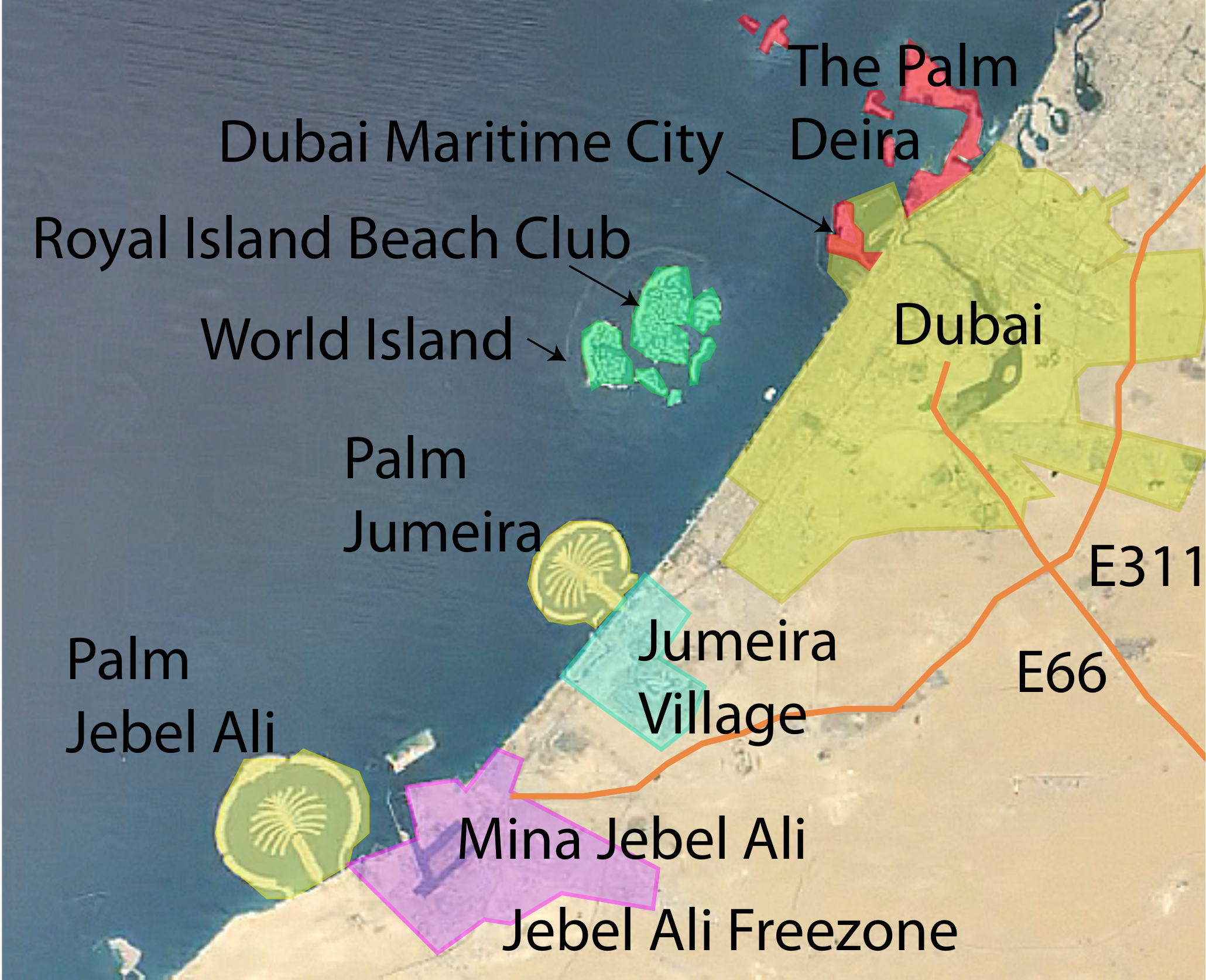}}}
	%	\subfigure[2008]{
	%	\label{fig:dubai11}
	%	\scalebox{0.2}{\includegraphics{Google-Time-Lapse/Dubai-Coastal-Expansion/Raw-Edited/Dubai-Coastal-2008_mod.pdf}}}
		\subfigure[2009]{
		\label{fig:dubai12}
		\scalebox{0.19}{\includegraphics{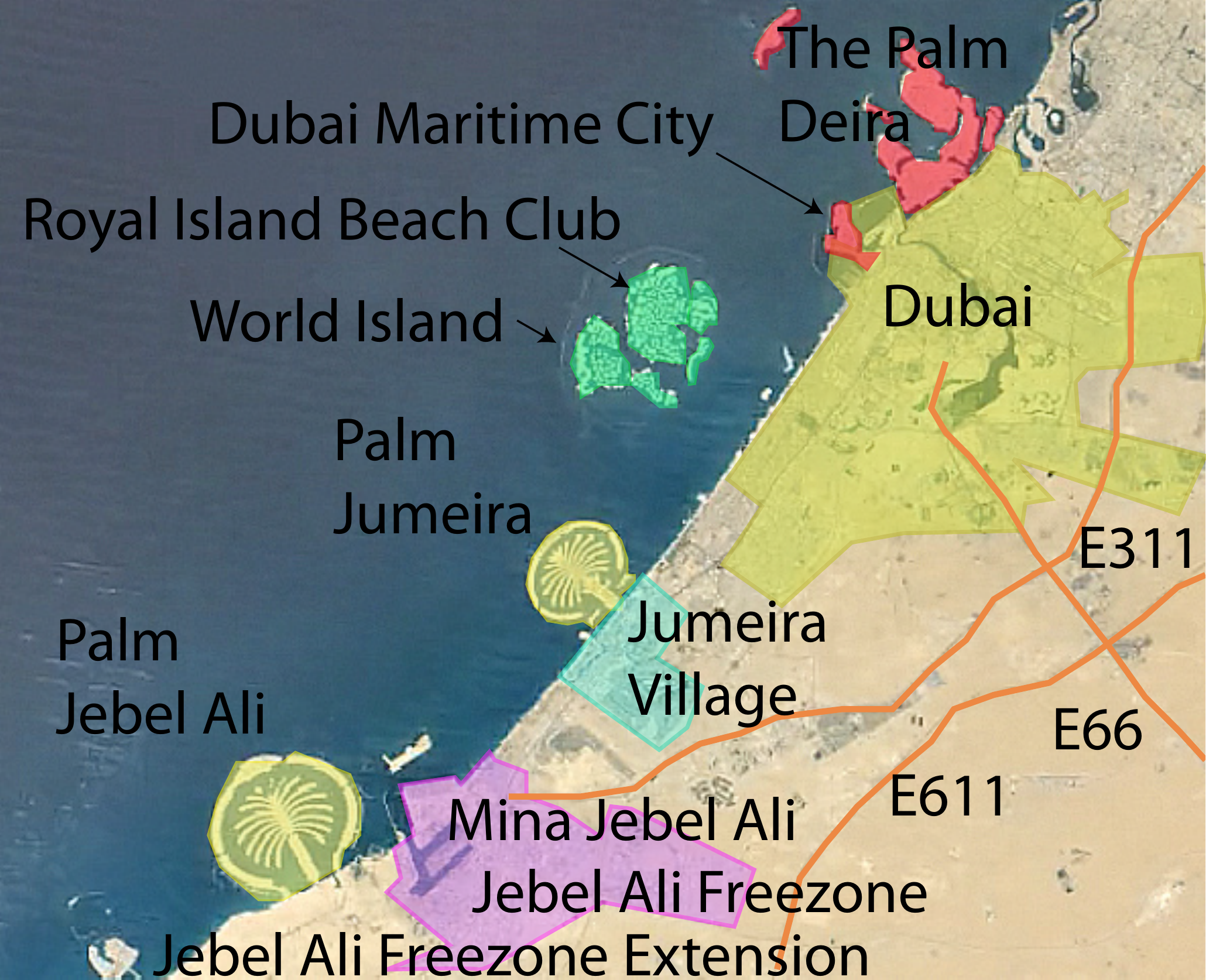}}}
	%	\subfigure[2010]{
	%	\label{fig:dubai13}
	%	\scalebox{0.2}{\includegraphics{Google-Time-Lapse/Dubai-Coastal-Expansion/Raw-Edited/Dubai-Coastal-2010_mod.pdf}}}
	%	\subfigure[2011]{
	%	\label{fig:dubai14}
	%	\scalebox{0.2}{\includegraphics{Google-Time-Lapse/Dubai-Coastal-Expansion/Raw-Edited/Dubai-Coastal-2011_mod.pdf}}}
		\subfigure[2012]{
		\label{fig:dubai15}
		\scalebox{0.19}{\includegraphics{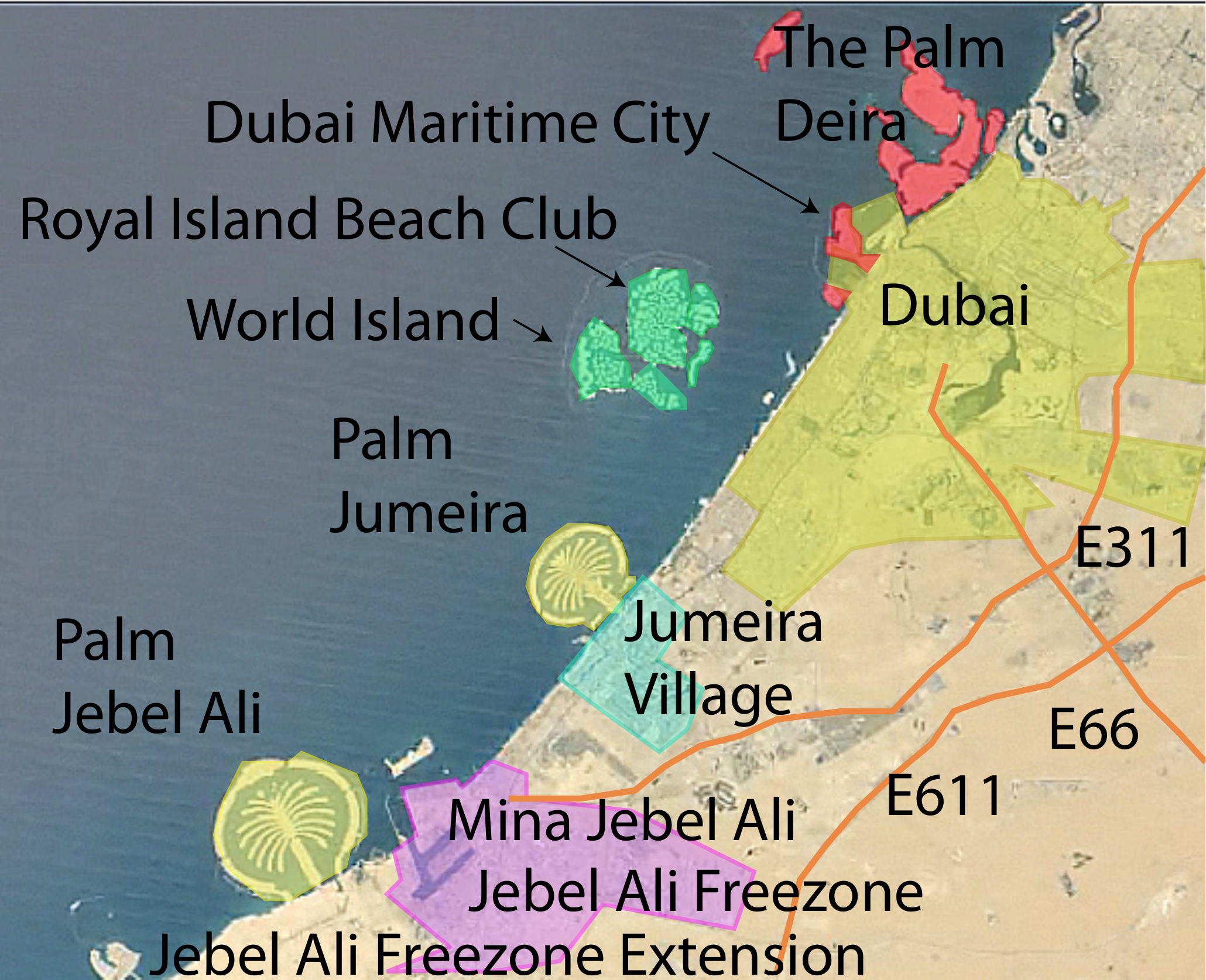}}}
	\caption{Urban expansion in Dubai, United Arab Emirates}
	\label{fig:dubai}
\end{figure}

\textbf{Urbanization over 28 Years}.\quad As an example, consider the phenomena of \emph{urbanisation} during 1984-2012 for the cities of \emph{Las Vegas} (USA) and \emph{Dubai} (UAE); the following expert analyses in (N1--N2) describes the high-level geospatial, demographic, economic, environmental and other related processes pertaining to urbanisation. The strictly spatio-temporal determinants of urbanisation in these cities are depicted in Fig. \ref{fig:vegas} and Fig. \ref{fig:dubai} respectively. The data and analyses have been sourced via the publicly available {\sffamily TimeLapse} initiative (Listing 1).

\begin{center}
\colorbox{gray!18}{
\begin{minipage}{0.95\textwidth}
%\centering\small
{\footnotesize
$~$\hfill\textbf{Listing 1}.

\smallskip

\textbf{\small TimeLapse.}.\quad TimeLapse is a collaborative project involving Google, the U.S. Geological Survey (USGS), NASA, TIME, and Carnegie Mellon University's CREATE Lab. TimeLapse has recently released an interactive animation constructed from satellite images of the Earth; the satellite images, sourced from the Landsat program, represent images dating back to 1984 detailing a year-by-year progression of changes to the surface of the Earth. 

\smallskip

Preliminary view of the generated data depicts phenomena such as deforestation in the Amazon, the effects of coal mining in Wyoming, the urban expansion of Shanghai and Las Vegas, and the drying of the largest lake in the Middle East, Lake Urmia. We use some of these publicly available examples from TimeLapse to establish a context for the overall context of this paper, and introduce the idea of narrative-centred interpretation in the geospatial domain.

{\hfill\textbf{Google, Nasa, Time}.\quad\footnotesize\texttt{http://world.time.com/timelapse/}}
}
\label{list:nasa}
\end{minipage}}
\end{center}
%\caption{\label{list:nasa}...}

\begin{quote}
{\small\textbf{N1}.}\quad\textbf{Las Vegas}\quad 

{\small``\sffamily{Throughout the 1990s and much of the 2000s, the \underline{boundaries of metro Las Vegas kept} \underline{expanding}, as \underline{new housing developments} were thrown up to accommodate the throngs of Americans who wanted to take advantage of the \underline{regionÕs booming economy}. From 2000 to 2010, the city's \underline{population grew by} nearly 50\% -- a rate thatÕs hard to find outside the developing world.

But if \underline{Las Vegas boomed along with the housing sector during} the first several years of the 21st century, \underline{it went bust when the recession hit}. The city was ground zero for the foreclosure crisis. As late as 2012, Las Vegas had \underline{one foreclosure filing for every 99 housing units}, good for the \underline{fourth highest rate in the country}. And as economically unsustainable as Las Vegas' growth has proved to be over the past several years, it may be even more environmentally unsustainable. The \underline{city receives almost no rain}, and most of its \underline{water comes from} \underline{nearby Lake Mead}. But as can be clearly seen in the TimeLapse images, \underline{Lake Mead is} \underline{drying up}, the victim of a \underline{prolonged drought} -- potentially abetted by climate change -- and the \emph{increasing demand placed} on it by \underline{Las Vegas' growing population}. Lake MeadÕs \underline{water level has fallen} from a little over 1,200 ft. (365 m) to 1,125 ft. (343 m) now. In recent years, officials in Las Vegas have taken admirable steps to reduce water waste, but \underline{if Lake Mead keeps shrinking}, \underline{Sin City will stop growing}.}''}
\end{quote}

\begin{quote}
{\small\textbf{N2}.}\quad\textbf{Dubai}.\quad 

{\small``\sffamily{In the mid-1980s, [...], \underline{Dubai was} a \underline{small desert city} of about 300,000 people, \underline{overshadowed} \underline{by nearby Abu Dhabi}, the capital of the United Arab Emirates. What growth Dubai had experienced was mostly recent; \underline{in the 1950s it was little more than a village}, with pearl diving its chief industry. Today, DubaiÕ's population exceeds 2.1 million, and the metropolis has asserted itself as the financial center of the Middle East.

Dubai is a city that seemed to grow almost overnight, like a desert oasis made real. It has the worldÕs tallest skyscraper -- the Burj Khalifa, ..., -- as well as its largest mall, its biggest theme park and its longest indoor ski run..... Not content with simply building in the desert, \underline{over the past couple of decades Dubai has built out into the sea}. \underline{Sand dredged} \underline{from the seafloor} has been \underline{used to create artificial islands of recognizable shapes} -- including a pair of palm trees. In the lower-right corner of the Timelapse images, \underline{areas of empty} \underline{sand} are \underline{filled up with new buildings}, as the \underline{city grows further and further away from the sea}, pushing into the desert. That breakneck \underline{pace of development has slowed somewhat} in recent years, as \underline{Dubai was hit hard by the global recession} of 2008.}''}
\end{quote}

\medskip

In general, high-level expert analysis encompassing commonsense, qualitative interpretation  (e.g., the underlined parts in N1--N2 above) of urban / geospatial processes may be identified from measurable low-level spatial and temporal features, themselves obtainable from a range of data sources such as \emph{satellite imagery and remote sensing, land-surveys, physical environmental sensing a la sensor networks}, etc. In particular, the complex dynamics underlying the identification of urbanisation processes may encompass several data sources such as:

\begin{enumerate}
{%\footnotesize
	\item Satellite imagery% (SI)

	\item Remote sensing %(RS)

	\item Land use statistics and databases, Gazetteers %(LU)

	\item Demographic data\\(e.g., from census surveys) 
		
	\item Economic data\\(income, growth, economic activity, currency and stock market performance, etc.)
}
\end{enumerate}

%\textbf{Scope of this paper}.\quad 
The focus of the narrative-centred model presented in this paper is strictly on the spatio-temporal aspects of the dynamic geospatial phenomena that underlie perceivable geospatial change at the \emph{object} or \emph{feature} level. The spatio-temporal aspects can be co-related with other kinds of quantitative and qualitative data (e.g., \emph{economic and demographic measures}, \emph{census studies}); however, a formal treatment of such correlations is beyond the scope of this paper. We emphasise  that modelling and reasoning about such correlations would indeed be possible, and also be within the scope of the overall narrative based analytical framework for GIS that has been proposed in this paper.

\section{\textsc{Qualitative Spatial Representation and Reasoning}}\label{sec:csc}
%\section{Commonsense, Space, Change: A Brief Overview of Formal Models}\label{sec:csc}

The field of qualitative spatio-temporal  representation and reasoning (QSTR) seeks to define formal models of spatial and temporal relations dealing with different aspects of space such \emph{topology, direction, distance, size}, etc. QSTR has evolved as a specialised discipline within Artificial Intelligence \citep{IntervalCalc:Allen:1983,cosyFreksa1991a,Cohn_Renz_07_Qualitative,renz-nebel-hdbk07,bhatt2011-scc-trends}.
Formal methods in QSTR provide a commonsensical interface to abstract and reason about quantitative spatial information.

The common characteristic of the developed formal models, often termed qualitative calculi, is that, in contrast to quantitative approaches, just a small number 
of basic relations is distinguished. Qualitative spatial / temporal calculi are relational-algebraic systems pertaining to one or more aspects of space. They abstract from metrical details and focus on properties that make a difference in a  particular application domain. This allows for an analysis of spatio-temporal data on a high-level of abstraction and directly with respect to human spatio-temporal concepts and commonsense reasoning. As such, they provide one means to represent and analyse 
data in an abstract way that is more natural to humans,  a key challenge identified for future GIS (e.g., \cite{5906,Mennis00:conceptual,DBLP:journals/tgis/Gahegan96}).
The basic tenets in QSTR consist of constraint based reasoning algorithms over an infinite (spatial) domain to solve \emph{consistency} problems in the context of spatial calculi. The key idea here is to partition an infinite quantity space into finite disjoint categories, and utilise the special relational properties of such a partitioned space for reasoning purposes.

In general, qualitative spatial calculi can be classified into two groups: topological and positional calculi. With topological calculi such as the \emph{Region Connection Calculus} (RCC), the primitive entities are spatially extended regions of space, and could possibly even be 4D spatio-temporal histories, e.g., for \emph{motion-pattern} analyses. Alternatively, within a dynamic domain involving translational motion, point-based abstractions with orientation calculi suffice. Examples of orientation calculi include \citep{Cohn_Renz_07_Qualitative}: the Oriented-Point Relation Algebra ({\small\OPRAm}), the Double-Cross Calculus, and the line-segment based \emph{Dipole Calculus}.

\begin{figure}
\centering
\includegraphics[width=0.99\columnwidth]{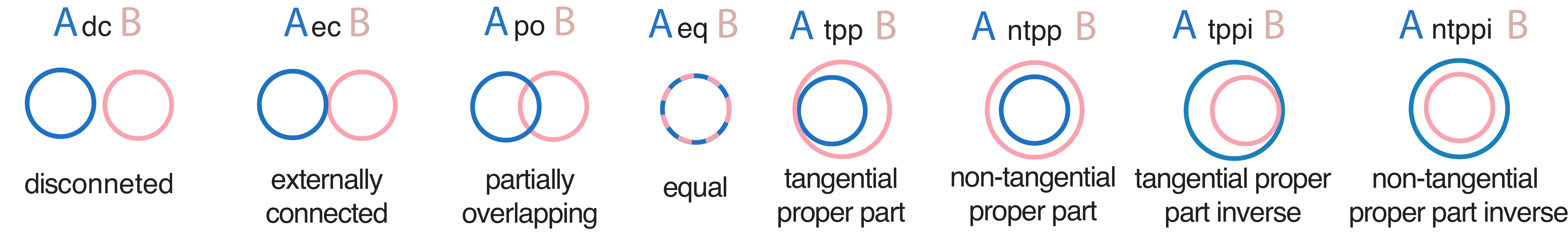}
\caption{Basic relations of the topological RCC-8 calculus}
\label{fig:rcc8}
\end{figure}

Similar to these works, which are situated within an Artificial Intelligence / Knowledge Representation (KR) context, many crucial advances have accrued from other communities concerned with the development of formalisms and algorithms for modelling and reasoning about spatial information, a prime example here being the domain of spatial information theory for Geographic Information Systems (GIS) \citep{Egenhofer:1991:pointset,citeulike:4160678}. Most widely adopted and applied qualitative spatial (topological) calculi are the RCC-8 calculus \cite{Randell_Cui_Cohn_92_A} and \emph{9-Intersection Model} \cite{Egenhofer_91_Reasoning} which both essentially distinguish the same eight basic topological relations between two spatial regions as shown in Fig.~\ref{fig:rcc8}.\footnote{In this paper, we will use RCC-8 in all our examples.}

\textbf{Relevant Applications in GIS, and Urban Planning}.\quad Qualitative spatial calculi have, for instance, been utilized in the GIS domain to describe spatial relationships in query and retrieval scenarios \cite{Clementini_Sharma_Egenhofer_94_Modeling,Abdelmoty_09_Supporting}, to formalize (geo)spatial concepts and processes\cite{Claramunt_97_A,Klippel_Worboys_Duckham_08_Identifying,Jiang_Worboys_08_Detecting,Duckham_10_Decentralized}, and to specify background knowledge and integrity constraints in the context of spatial and spatio-temporal database applications \cite{DBLP:conf/dlog/HaarslevM97,Kahn_Schneider_10_Topological,Rodriguez:2010:MCR:1869790.1869818}. The notion of conceptual neighborhood \cite{Freksa_91_Conceptual,Egenhofer_Al-Taha_92_Reasoning}  has been introduced to describe spatial change on the level of qualitative spatial relations and forms the basis to perform temporal reasoning in the form of simulation, interpolation, and planning.

\textbf{Tools}.\quad Spatial reasoning techniques manifest themselves in several ways as practical tools aimed at providing general spatial abstraction, reasoning, consistency and constraint satisfaction tasks, prime examples here being systems \emph{CLP(QS)} \citep{bhatt-et-al-2011,DBLP:conf/ecai/SchultzB12}, \emph{SparQ} \citep{cosy:sparq-sc06}, \emph{GQR} \citep{GQR:2009:AAAI-Sym}, and the generic toolkit \emph{QAT} for n-ary calculi \citep{DBLP:conf/time/CondottaSL06}.

\section{\textsc{The Spatial Informatics of Geospatial Dynamics}}	\label{sec:spat-informatiks}

The spatial information theoretic challenges underlying the development of high-level analytical capability in dynamic GIS consist of fundamental representational and computational problems pertaining to: the semantics of spatial occurrences, practical abduction in GIS, and to support these, problems of data abstraction, integration,  and spatial consistency.

%\citet{Claramunt_Theriault_95_Managing} define a taxonomy of spatio-temporal processes distinguishing three main types: (1) processes that concern the evolution of a single entity again classified into basic processes (e.g., appearance, disappearance), transformation processes (expansion, contraction, deformation), and movement processes (displacement, rotation), (2) processes concerning the functional relationship between two entities (!!!examples\commentJOW{look up}!!!), and (3) processes concerning the evolution of spatial structures involving several entities (split, union, reallocation). Based on these, an architecture for an event-oriented temporal GIS that supports queryies about spatial, temporal, and thematic aspects is developed. 

\begin{figure}
\centering
\includegraphics[width=0.6\textwidth]{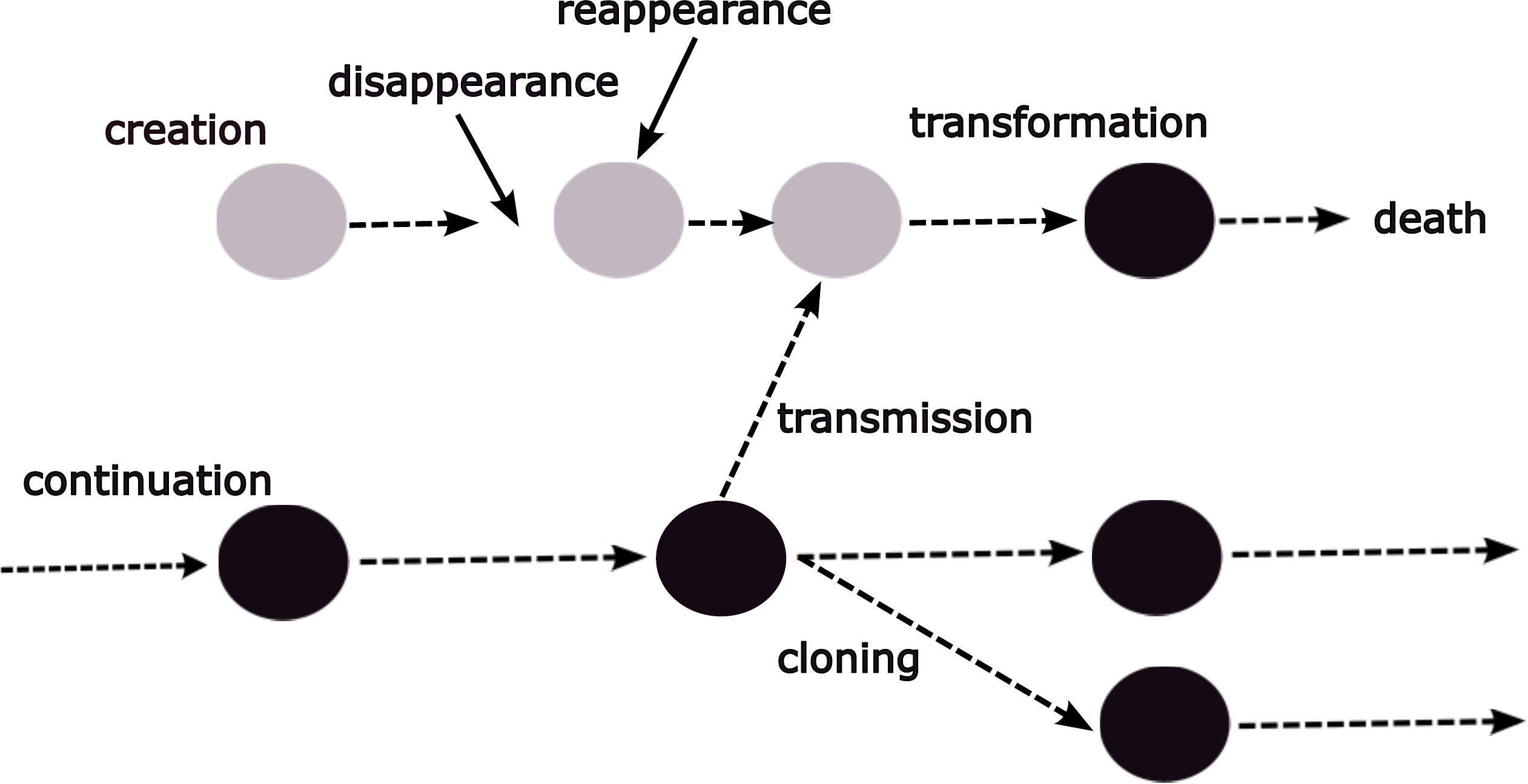}
\caption{{\sffamily Object Change History, Source: \citep{journals/gis/Worboys05}}}
\label{fig:worboys-object-change-hist1}
\end{figure}

\subsection{Spatial Occurrences: Analyses with Events and Objects}\label{sec:occurrences}

Our objective is to develop the functionality that enables reasoning about spatio-temporal narratives consisting of events and processes at the geographic scale. We do not attempt an elaborate ontological characterisation of events and processes, a topic of research that has been addressed in-depth in the state-of-the-art. For the purposes of this paper, we utilise a minimal, yet rich, conceptual model consisting of a range of events such that it may be used to qualitatively ground metric geospatial datasets consisting of spatial and temporal footprints of human and natural phenomena at the geographic scale.

Spatial occurrences may be defined at two levels: (I). \emph{domain-independent}, and (II). \emph{domain-dependent}:

\subsubsection*{\bfseries I. Domain Independent Spatial Occurrences.} 
These occurrences are those that may be semantically characterized within a general theory of space and spatial change. These may be grounded with respect to either a qualitative theory, or an elaborate typology of geospatial events (e.g., Fig. \ref{fig:worboys-object-change-hist1} illustrates a typology for object level changes).

\subsection*{Spatial Changes at a Qualitative Level.}
\noindent In so far as a general qualitative theory of spatial change is concerned, there is only one type of occurrence, viz - a transition from one qualitative state (relation) to another state (relation) as (possibly) governed by the continuity constraints of the relation space. At this level, the only identifiable notion of an occurrence is that of a qualitative spatial transition that the primitive objects in the theory undergo, e.g.,  the transition of an object ($o_{1}$) from being \emph{disconnected} to another object ($o_{2}$) to being a \emph{tangential-proper-part} (see again Fig.~\ref{fig:rcc8}). At the level of a spatial theory, it is meaningless to ascribe a certain spatial transition as being an event or action; such distinctions demand a slightly higher level of abstraction. For instance, the example of a transition from disconnected to tangential-proper-part could either coarsely represent the volitional movement of a person into a room or the motion of a ball. Whereas the former is an action performed by an agent, the latter is a deterministic event that will necessarily occur in normal circumstances. Our standpoint here is that such distinctions can only be made in a domain specific manner; as such, the classification of occurrences into actions and events will only apply at the level of the domain with the general spatial theory dealing only with one type of occurrence, namely primitive spatial transitions that are definable in it.

\begin{figure}[t]
\centering
\includegraphics[width=0.99\columnwidth]{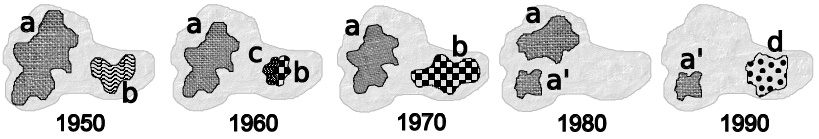}
\caption{Abduction in GIS. Source: \citep{Bhatt:RSAC:2012}}
\label{fig:event-gis}
\end{figure}

\subsection*{Typology of Events and Patterns.}
At the domain independent level, the explanation may encompass behaviours such as \emph{emergence}, \emph{growth} \& \emph{shrinkage}, \emph{disappearance}, \emph{spread}, \emph{stability}, etc., in addition to the sequential/parallel composition of the behavioural primitives aforementioned, e.g., \emph{emergence} followed by \emph{growth}, \emph{spread / movement}, \emph{stability} and \emph{disappearance} during a time-interval. Certain kinds of typological elements, e.g., \emph{growth} and \emph{shrinkage}, may even be directly associated with spatial changes at the qualitative level. Appearance of new objects and disappearance of existing ones, either abruptly or explicitly formulated in the domain theory, is also characteristic of non-trivial dynamic (geo)spatial systems. Within event-based GIS, appearance and disappearance events are regarded to be an important typological element for the modelling of dynamic geospatial processes \citep{Claramunt_Theriault_95_Managing,journals/gis/Worboys05}. For instance, \citet{Claramunt_Theriault_95_Managing} identify the basic processes used to define a set of low-order spatio-temporal events which, among other things, include appearance and disappearance events as fundamental. Similarly, toward event-based models of dynamic geographic phenomena, \citet{journals/gis/Worboys05} suggests the use of the appearance and disappearance events at least in so far as single object behaviours are concerned (see Fig. \ref{fig:event-gis}). Appearance, disappearance and re-appearances are also connected to the issue of object identity maintenance in GIS \citep{DBLP:conf/kr/Bennett02,Hornsby_Egenhofer_00_Identity}.%Changes with regard to the identity states of objects have been investigated in \citep{Hornsby_Egenhofer_00_Identity}. Primitive identity states for a single object considered in this work are existing, non-existing without history and non-existing with history. Nine elementary identity-based change types are identified and then generalized to changes involving two objects resulting in 18 possible change types that form the basis for a change description language to represent sequences of identity-based changes.

\subsubsection*{\bfseries II. Domain-Specific Spatial Occurrences}
At a domain-dependent level, behaviour patterns may characterize high-level processes,  environmental / natural and human activities such as \emph{deforestation, urbanisation}, \emph{land-use transformations} etc. These are domain-specific occurrences that induce a transformation on the underlying spatial structures being modelled \citep{Couclelis-Cosit09}. Basically, these are domain specific events or actions that have (explicitly) identifiable occurrence criteria and effects that can be defined in terms of qualitative spatial changes, and the fundamental typology  of spatial changes. For instance, in the example in Fig. \ref{fig:event-gis},  we can clearly see that region $a$ has continued to \emph{shrink} during 1950 to 1990, eventually \emph{disappearing} altogether. The following general notion of a `spatial occurrence' is identifiable \cite{bhatt:scc:08}: %\commentJOW{thought: can we replace example + image (and maybe other examples in this section) by something from the Dubai/Vegas scenario?}
\begin{quote}\small
`\emph{Spatial occurrences are events or actions with explicitly specifiable occurrence criteria and/or pre-conditions respectively and effects that may be identified in terms of a domain independent taxonomy of spatial change that is native to a general qualitative spatial theory}'.
\end{quote}

\noindent As an example, consider an event that will \emph{cause} a region to \emph{split} or make it \emph{grow / shrink}. Likewise, an aggregate cluster of geospatial entities (e.g., in wildlife biology domain) may \emph{move} and change its orientation with respect to other geospatial entities. Thinking in agent terms, a spatial action by the \emph{collective / aggregate} entity, e.g., \emph{turn south-east}, will have the effect of changing the orientation of the cluster in relation to other entities. In certain situations, there may not be a clearly identifiable set of domain-specific occurrences with explicitly known occurrence criteria or effects that are definable in terms of a typology of spatial change, e.g., cluster of alcohol-related crime abruptly appearing and disappearing at a certain time. However, even in such situations, an analysis of the domain-independent events and inter-event relationships may lead to an understanding of spatio-temporal relationships and help with practical hypothesis generation \citep{Beller:1991:ST-Events}.

\subsection{Practical Abduction for GIS}

%\subsubsection{Explanation by Spatio-Temporal Abduction}
Explanatory reasoning requires the ability to perform abduction with spatio-temporal information. In the context of formal spatio-temporal calculi, and logics of action and change, this translates to the ability to provide \emph{scenario and narrative completion} abilities at a high-level of abstraction.

\noindent Consider the GIS domain depicted in Fig.~\ref{fig:event-gis}, and the basic conceptual understanding of spatial occurrences described in Section \ref{sec:occurrences}. At a domain-independent level, the scene may be described using topological and qualitative size relationships. Consequently, the only changes that are identifiable at the level of the spatial theory are \emph{shrinkage} and eventual \emph{disappearance} -- this is because a domain-independent spatial theory may only include a generic typology (\emph{appearance, disappearance, growth, shrinkage, deformation, splitting, merging}, etc.) of spatial change. However, at a domain-specific level, these changes could characterize a specific event (or process) such as \emph{deforestation}. The hypotheses or explanations that are generated during a explanation process should necessarily consist of the domain-level occurrences in addition to the underlying (associated) spatial changes (as per the generic typology) that are identifiable. Intuitively, the derived explanations more or less take the form of existential statements such as: ``{\upshape Between time-points $t_{i}$ and $t_{i}$, the process of deforestation is abducible as one potential hypothesis}''. Derived hypotheses / explanations that involve both domain-dependent and as well their corresponding domain-independent typological elements are referred to as being `\emph{adequate}' from the viewpoint of explanatory analysis for a domain. At both the domain-independent as well as dependent levels, abduction requires the fundamental capability to interpolate missing information, and understand partially available narratives that describe the execution of high-level real or abstract processes. In the following, we present an intuitive overview of the scenario and narrative completion process.

\begin{figure}[t]
\begin{center}
\includegraphics[width=0.7\columnwidth]{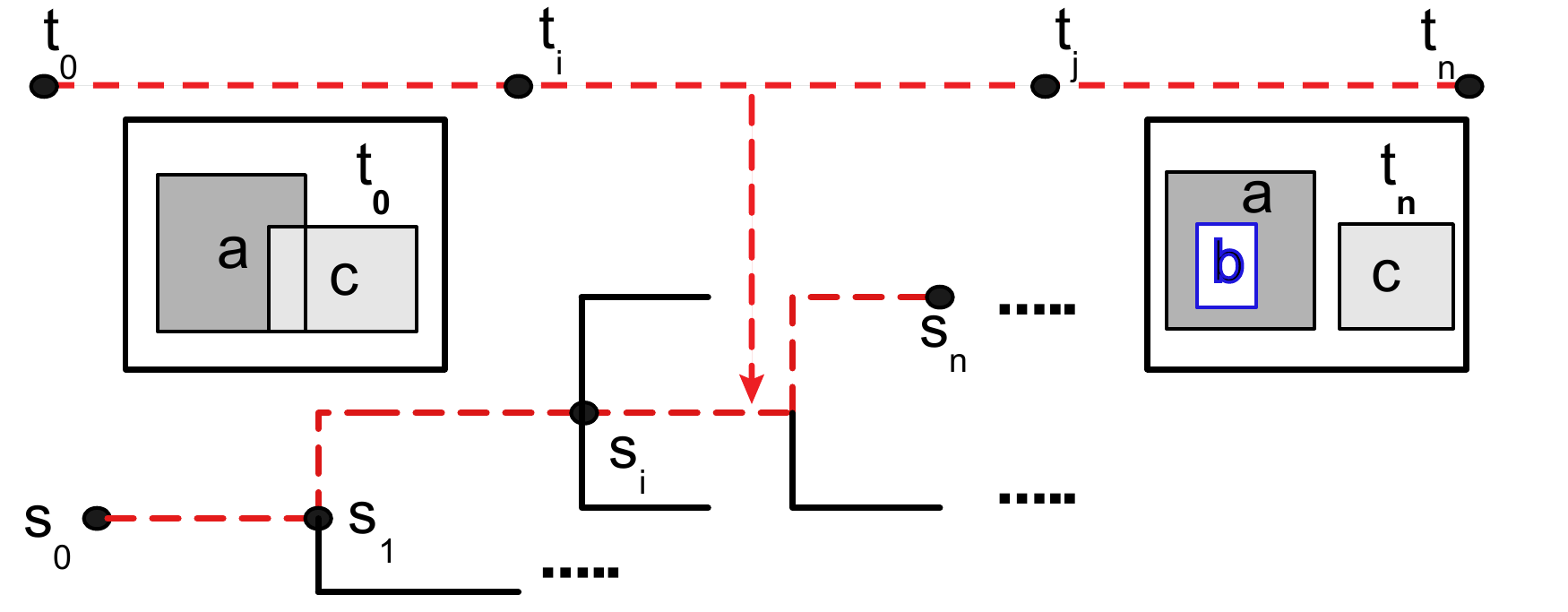}
\end{center}    
\vspace{-7pt}
\caption{Branching / Hypothetical Situation Space. Source: \citep{Bhatt:RSAC:2012}}
\label{fig:narrative-branching}
\end{figure}

%\noindent \bul\emph{Landscape Analysis Domain}:\commentMB{Explain domain here...before going into the intuitive overview of causal explanation}
% 

\subsubsection{ Scenario and Narrative Completion}
Explanation problems demand the inclusion of a narrative description, which from the logic-based viewpoint of this paper is essentially a distinguished course of actual events about which we may have incomplete information \citep{MillerS94,OccurNarraSC:Pinto:1998}. Narrative descriptions are typically available as \emph{observations} from the real / imagined execution of a system or process. Since narratives inherently pertain to actual observations, i.e., they are \emph{temporalized}, the objective is often to assimilate / explain them with respect to an underlying process model and an approach to derive explanations.

\noindent Given partial narratives that describe the evolution of a system (e.g., by way of temporally ordered scene observations in event-based GIS datasets) in terms of high-level spatio-temporal data, \emph{scenario and narrative completion} corresponds to the ability to derive completions that bridge the narrative by interpolating the missing spatial and action / event information in a manner that is consistent with domain-specific and domain-independent rules / dynamics. 

Consider the illustration in Fig. \ref{fig:narrative-branching} for a branching / hypothetical situation space that characterizes the complete evolution of a system. In Fig. \ref{fig:narrative-branching} -- the situation-based history $<s_{0},~s_{1},\ldots,~s_{n}>$ represents one path, corresponding to an actual time-line $<t_{0},~t_{1},\ldots,~t_{n}>$, within the overall branching-tree structured situation space. Given incomplete narrative descriptions, e.g., corresponding to only some ordered time-points in terms of high-level spatial (e.g., topological, orientation) and occurrence information, the objective of \emph{causal explanation}  \citep{bhatt:scc:08} in a spatio-temporal context  is to derive one or more paths from the branching situation space, that could best-fit the available narrative information. Of course, the completions that bridge the narrative by interpolating the missing spatial and action/event information have to be consistent with domain-specific and domain-independent rules/dynamics.

\noindent Explanation, in general, is regarded as a converse operation to temporal projection essentially involving reasoning from effects to causes, i.e., reasoning about the past \citep{Shanahan:1989:pred-explana}. Logical abduction is one inference pattern that can be used to realise explanation. In Section \ref{sec:geo-formal-framework}, we present a practical illustration of the concept of scenario and narrative completion (by abduction) for explanatory analysis in the GIS domain.
%  
%\noindent Many different formalizations of causal explanation with spatial knowledge, such as within a belief revision framework \citep{belief-revision-85}, nonmonotonic causal formalizations in the manner of \citep{non-mono-causal-th-04} are possible and the subject of ongoing study. Additionally, the suitability of event calculus \citep{Events:Kowalski:1986,DBLP:journals/cacm/Mueller09} vis-\`{a}-vis the situation calculus is also a topic that especially merits detailed treatment.

%Logical frameworks for performing causal explanation with spatial information
%generally require that the input information is consistent, meaning that the
%combined input data is compliant with the
%underlying logical spatial theory. 

\subsection{Temporal Partitioning, Qualitative Abstraction, and Integration}
\label{sec:qualabs}

\subsubsection{ Temporal Partitioning and Qualitative Abstraction}
\label{sec:q+sc}

In the geographic domain, the input data often stems from multiple sources, for instance from different sensors, 
remote sensing data, map data, etc., and the data itself is afflicted by measurement 
errors and uncertainty. To perform explanatory analysis on a level
of qualitative spatial relations, geo-referenced quantitative input data
about spatial objects from different sources needs to be translated into relations from several qualitative
spatial models or calculi dealing with different aspects of space, a process we refer to as \emph{qualitative abstraction}. A prerequisite for applying the qualitative abstraction procedure 
is that the input data is temporally partitioned such that each part is associated with a particular time point in an ordered sequence of time points.   
For each time point, the qualitative abstraction procedure takes the associated quantitative data and derives 
the spatial relations from the   given qualitative models holding between the involved objects.
The result is a static qualitative spatial description for each time point.
If uncertainty of quantitative information is explicitly represented, this needs to be
taken into account and may lead to disjunctions of relations on the qualitative level.

\subsubsection{ Integration and Spatial Consistency}

\begin{figure}[t]
\centering
\subfigure[]{\includegraphics[width=0.25\textwidth]{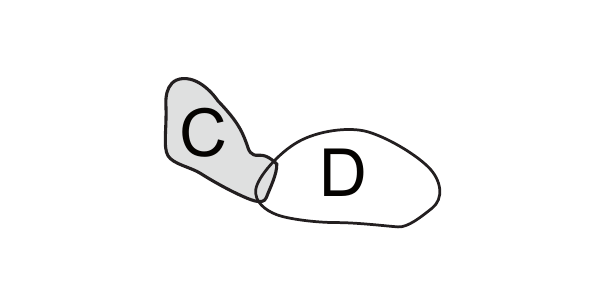}}
\subfigure[]{\includegraphics[width=0.25\textwidth]{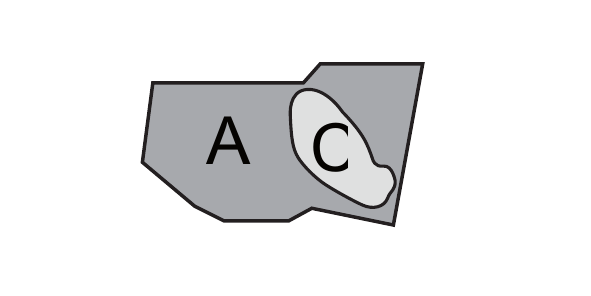}}
\subfigure[]{\includegraphics[width=0.25\textwidth]{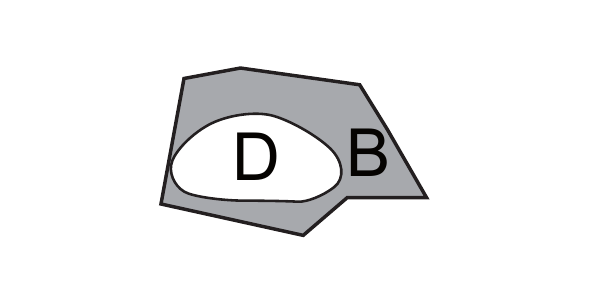}}
\subfigure[]{\includegraphics[width=0.2\textwidth]{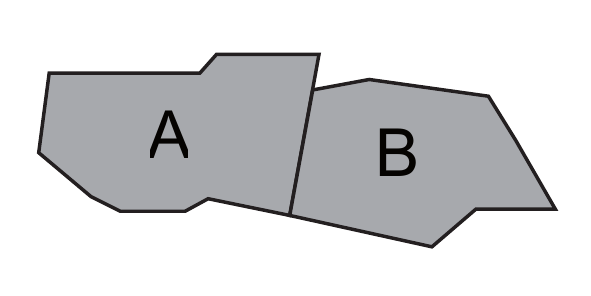}}
\caption{Information from four different sources which is inconsistent when combined.}
\label{fig:consistency}
\end{figure}

Due to the mentioned measurement errors and uncertainty of the quantitative
input data, the qualitative descriptions resulting from the qualitative abstraction for particular time points may contain contradictions or violate integrity constraints stemming from background knowledge about the domain. 
%Spatial inconsistencies, in particular, can easily arise from deviations in the perceived or computed
%spatial extensions of objects which can also lead to contradictions on the level of qualitative spatial relations. 
Fig.~\ref{fig:consistency} illustrates the case of a spatial inconsistency on the level of topological relations when combining the information from four different sources (all concerning the same time point): From combining the fact that objects $C$ and $D$ (e.g., two climate phenomena) are reported to \emph{overlap} by one source (a) with the reported relations $C$ is
completely \emph{contained in} $A$ (b) and $D$ is completely \emph{contained in} $B$ (c), it follows that the two regions $A$ and $B$ would need to overlap as well. This contradicts the information from the fourth source (d)---which could, for instance,
be a spatial databases containing boundaries of administrative regions---that says that $A$ and $B$  are \emph{externally connected}.
Instead of the fourth source, we could also have introduced a general integrity constraint
stating that administrative regions on the same level never overlap. This would have resulted in the same contradiction rendering
the given information inconsistent.

As a result of the possibility of inconsistent input information occurring in geographic
applications, frameworks for explanation and spatio-temporal analysis need the
ability to at least detect these inconsistencies in order to exclude the contradicting
information or, as a more appropriate approach, resolve the contradictions in a suitable
way. Removing logical inconsistencies is crucial in the context of a logic-based abductive reasoning
approach as we suggest in this paper, as otherwise,  incorrect conclusions can be abduced
from an inconsistency which will ultimately lead to incorrect results. While the view that
logical inconsistencies are undesirable has been challenged (see \cite{DBLP:conf/fair/GabbayH91}),
explanatory analysis with inconsistent information raises many challenges
going beyond the scope of this paper. In certain applications, it may be possible to derive that certain information  is irrelevant for the
explanatory task at hand and filter out this information  in advance
such that no removal of inconsistencies wrt.~ this information is required.

Deciding consistency of a set of qualitative spatial relations has been studied as
one of the fundamental reasoning tasks in qualitative spatial representation and reasoning \citep{Cohn_Renz_07_Qualitative}. The complexity of deciding
consistency varies significantly over the different existing qualitative calculi.
%While in the general case of the consistency decision problem the constraining
%qualitative relations between the objects can be disjunctions of several base
%relations of the involved calculus expressing uncertainty, this is usually
%not the case when the relations are ultimately computed from geo-referenced
%quantitative data. In these tasks it is usually sufficient to restrict oneself to descriptions in %which no disjunctions are allowed, so-called \emph{scenarios}, representing a concrete
%spatial configuration.
%\commentJOW{by restricting to scenarios here, I avoid having to talk about backtracking and
%tractable subsets}
For most common qualitative calculi such as  RCC-8, % \citep{Randell_Cui_Cohn_92_A},
the consistency can be decided in cubic time when the input description
is a scenario which means it does not contain disjunction of relations. This is achieved by the path consistency
or algebraic closure method \cite{Mackworth_77_Consistency} which is ultimately based on a set of composition
axioms that state which relation can hold between objects $A$ and $C$ given
the relations holding between object $A$ and $B$ and between $B$ and $C$.
%\footnote{in the case of a binary calculus} 
For general description including disjunctions
a more costly backtracking search has to be performed.

Integrity constraints have been  investigated in the (spatial) database literature \cite{DBLP:conf/icalp/FaginV84,springerlink:10.1023/A:1009754327059}. 
As the example above shows, integrity rules in a geographic context often come in the form 
qualitative spatial relations that have to be satisfied by certain types of spatial entities.
These kinds of spatial integrity constraints can be dealt with by employing terminological reasoning to determine whether a certain integrity rule has to be applied
to a given tuple of objects and feeding the resulting constraints into a standard qualitative consistency checker together with the qualitative relations coming from the input data.

%\commentJOW{removed the part that talked about integration of consistency checking into abduction process here}
%Consistency---in particular spatial consistency---is a crucial aspect in any symbolic
%approach to reason about actions, change and events. When predicting or simulating 
%future events or explaining past events from a limited number observations, it has to be
%guaranteed that considered hypothetical situations are spatially consistent, meaning
%that the symbolic relations describing the spatial configuration can be satisfied by
%the involved objects in the underlying spatial domain. Therefore, one either has to
%ensure that spatial consistency is already appropriately captured in the developed
%logical theory underlying the employed reasoning framework or a separate procedures
%for deciding consistency needs to be incorporated as a special predicate (for instance
%using a dedicated spatial reasoner such as the SparQ reasoning engine).
%\JOW{would you agree with the last sentence?}\commentMB{This approach could be do-able, in so far as practical problem solving by system development is concerned. however, this defeats the purpose of the sorts of logic+space integrations that i have been developing, and certainly goes against what we are selling in the paper. we need to pitch this differently.}

\subsubsection{ Conflict Resolution}
\label{sec:si+cr}

As indicated in the previous section, when conflicts arise during the integration
of spatial data, it is often desirable to not only detect the inconsistencies but also resolve conflicts in a reasonable manner
to still be able to exploit all provided information in 
%Examples for this are integration of information from different sensors or the combination of geo-coordinates
%for a given set of objects with qualitative spatial relations that have been derived from
%explicitly given change information (e.g., when an administrative region is subdivided
%into two new regions, it (usually) follows that the resulting regions are \emph{neighbored}).
the actual logical reasoning approach for explanation and analysis.
Methods for data integration and conflict resolution have for instance been studied under
the term information fusion  \cite{Gregoire_Konieczny_06_Logic}.
They are commonly classified into quantitative approaches and symbolic
approaches. Quantitative approaches mainly employ statistical methods such
as least-square adjustment to deal with multiple observations, while symbolic 
information fusion is concerned with the revision of logical theories under
the presence of new evidence. An important distinction here is that
between \emph{revision} and \emph{update}. In the case of revision, additional information about a particular state of the world becomes available and needs to be combined
with what was known before. In the case of update, one assumes that the state
may have changed and that the new information is more up-to-date than the
previous knowledge. These different information fusion settings have led to 
the formulation of different rationality criteria that corresponding computational approaches 
should satisfy such as the AGM postulates for belief change \cite{Alchourron_Gaerdenfors_Makinson_85_On}.
Such computational solutions often consist of merging operators that compute
a consistent model that is most similar to the inconsistent input data. In distance-based merging approaches this notion of similarity is described using a distance measure
between models. This idea has been applied to qualitative spatial representations
\cite{Condotta_Kaci_Schwind_08_A,Dylla_Wallgruen_07_Qualitative}
using the notion of conceptual neighborhood \cite{Freksa_91_Conceptual,Egenhofer_Al-Taha_92_Reasoning} to measure distance in terms of the number of neighborhood changes that need to be performed to get from inconsistent qualitative descriptions to consistent ones.

\begin{figure}[t]
\centering{
\subfigure[]{\label{fig:merging_a}\includegraphics[width=0.37\textwidth]{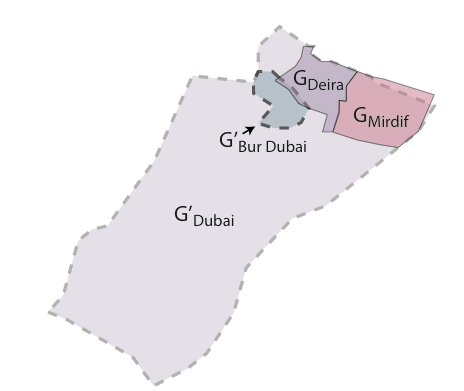}}%\subfigure[]{\label{fig:merging_a}\includegraphics[width=0.3\textwidth]{merging.pdf}}
\subfigure[]{\label{fig:merging_b}\includegraphics[width=0.3\textwidth]{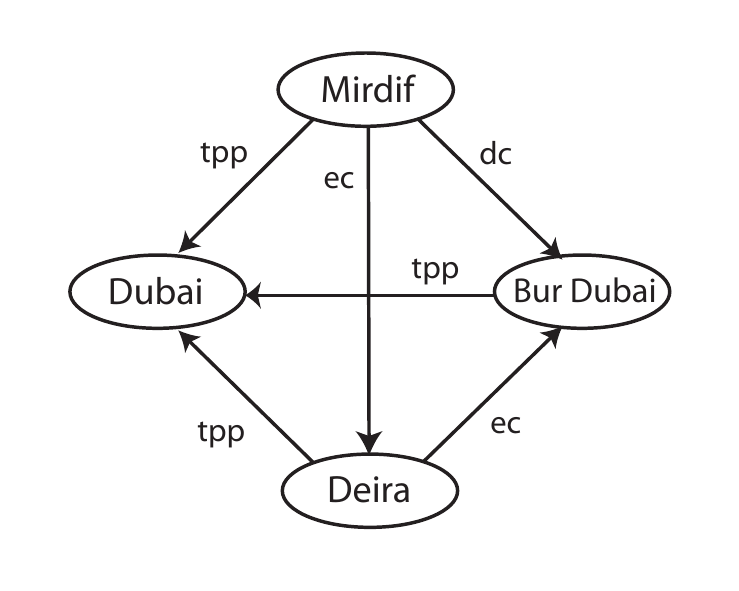}}
\subfigure[]{\label{fig:merging_c}\includegraphics[width=0.32\textwidth]{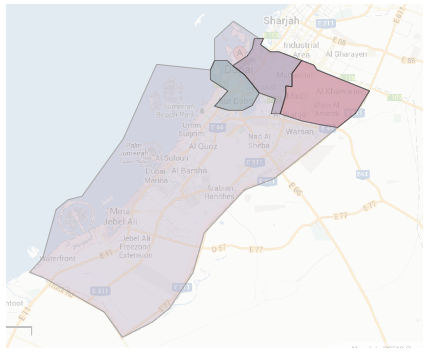}}%{resolution_example.pdf}}
\caption{Contradicting geometric information from two different sources (a), the qualitative merging result (b) based on domain-specific integrity constraints,
and the actual spatial configuration of the involved objects (c).}
\label{fig:merging}
}
\end{figure}

To illustrate the operation of a qualitative merging approach for conflict resolution, let us consider the example in Fig.~\ref{fig:merging} in which information
from two sources providing information about the city of Dubai (see again Fig.~\ref{fig:dubai}) at a given time
needs to be merged: Let us say the first source $G$ provides geometries for two districts, $G_{\text{Deira}}$ and $G_{\text{Mirdif}}$ (shown as polygons with fully drawn borders in Fig.~\ref{fig:merging_a}),  while the second source $G'$ provides geometries for another district, $G'_{\text{Bur Dubai}}$, and for the city of Dubai itself, $G'_{\text{Dubai}}$ (both shown as polygons with dashed boundaries in Fig.~\ref{fig:merging_a}.
Let us further assume that we also have two integrity constraints for the integration result in this scenario. The first one states that the geometries of districts cannot overlap and, thus,  have to be disjoint or touching
(disjunction $\{ \text{ec}, \text{dc} \}$ in terms of RCC-8 relations).  The second constraint demands that each geometry of a city district has to be completely contained in the geometry of 
the city. This corresponds to a disjunction of $\{  \text{ntpp}, \text{tpp} \}$ in terms of RCC-8 relations and applies to the relation of each of the three districts to $G'_{\text{Dubai}}$.
As Fig.~\ref{fig:merging_a} illustrates, superimposing the geometries of both sources
results in violations of both integrity constraints: 
$G_{\text{Deira}}$
and $G'_{\text{Bur Dubai}}$ overlap (RCC-8 relation po) and $G_{\text{Mirdif}}$ and $G'_{\text{Dubai}}$ overlap as well.
A qualitative merging approach would now resolve these conflicts on the qualitative level by computing the consistent scenario
that is closest to the qualitative interpretation of the input information. A possible result is shown in Fig.~\ref{fig:merging_b}. The
relation between $G_{\text{Deira}}$ and $G'_{\text{Bur Dubai}}$ has been changed to $\{\text{ec}\}$ and that of $G_{\text{Mirdif}}$ and $G'_{\text{Dubai}}$
to $\{\text{tpp}\}$, which corresponds to the actual spatial configuration depicted in Fig.~\ref{fig:merging_c}.

%\commentJOW{need to change RCC relations to lowercase}

%\commentJOW{removed the part about the spatio-TEMPORAL merging here as it doesnt to fit for this paper}
%Following this general line of distance-based conflict resolution
%approaches, a spatio-temporal merging approach is proposed in \cite{JOW-FD-ECAI}.
%In order to resolve conflicts when merging several snapshot-based spatio-temporal 
%databases, a sequence of qualitative scenarios is computed that accumulates the overall
%lowest distance from the static snapshots. The sequence is required to be continuous
%in the sense that only elementary conceptual neighborhood transitions are allowed between
%consecutive scenarios. It has to be pointed that the described approach treats space
%and time differently as it only considers modifications of the spatial relations to resolve
%conflicts but not modifications of time intervals associated with the snapshots.

\section{\textsc{Geospatial Analytics: A Narrative Based Formal Framework}}\label{sec:geo-formal-framework}

\begin{figure}[t]
\centering
\includegraphics[width=1\textwidth]{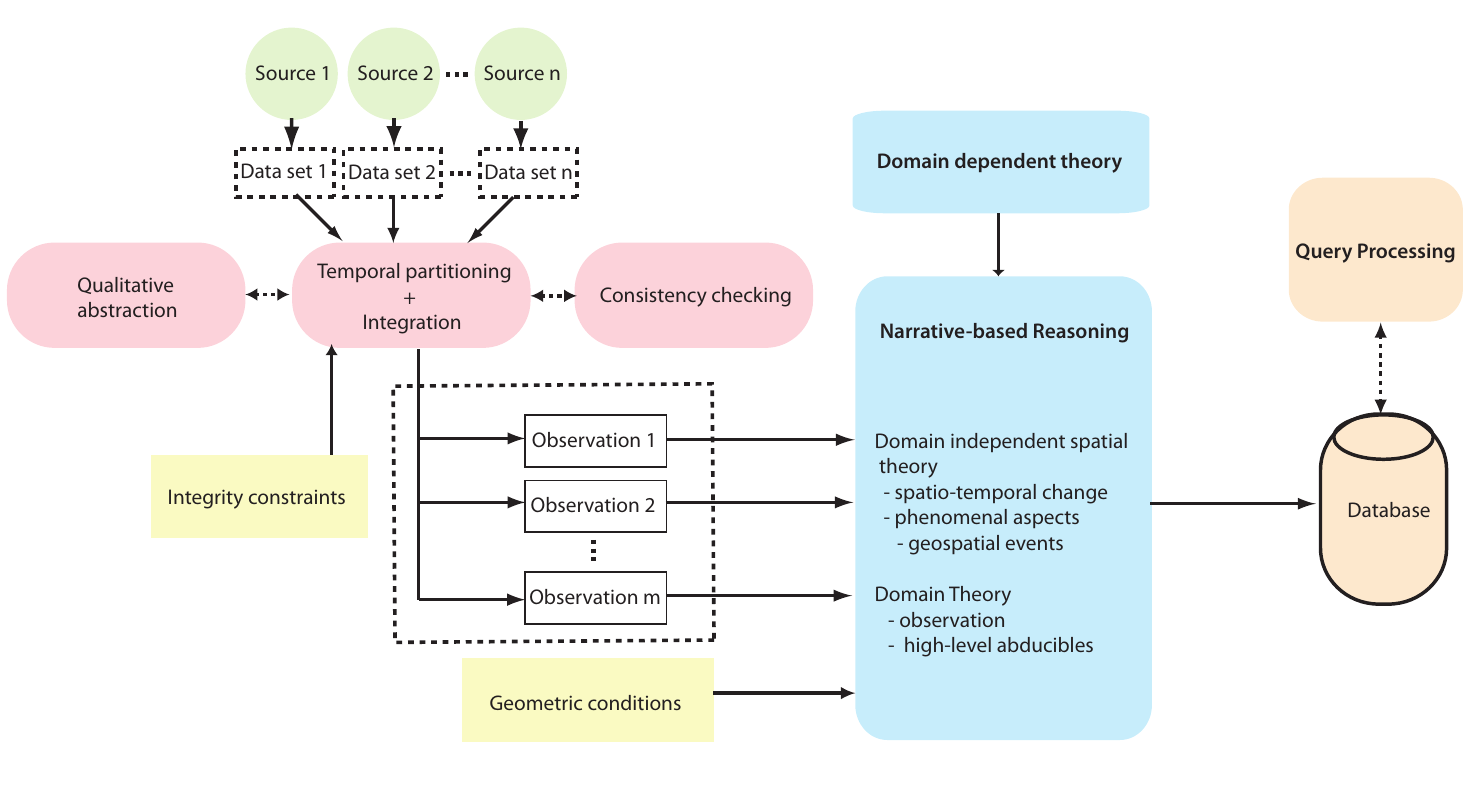}
\caption{Overview of narrative-based architecture for geospatial modelling, explanatory analysis, querying, and visualization}
\label{fig:architecture}
\end{figure}

The discussions in Section 3 encompassed a range of representation and computational challenges that accrue in the context of dynamic geospatial analysis. We now describe  our  formal framework, and its corresponding conceptual architecture, for high-level qualitative modeling and explanatory analysis for the domain of geospatial dynamics illustrated in Fig.~\ref{fig:architecture}.

%\commentMB{we will need to extend this with matching component...also, dataset 1, ...dataset n, might be grouped together, and the matching component could take its data from there, and some human-input as the other intake?}

\subsection{Overview of Architecture}

Our proposed architecture comprises the entire range of  steps required to perform explanatory analysis of geospatial dynamics on a qualitative level of abstraction
starting with the processing of the actual (typically quantitative) data to form a consistent qualitative description, to the usage of abductive reasoning for narrative completion, and to the recognition of high-level processes leading to a knowledge base that can be queried and utilized by application systems and decision makers. The main aspects of the proposed architecture are the following: % \commentJOW{we can give each bullet a name / heading too}

\begin{itemize}
\item \textbf{Input datasets}.\quad The input consists of \emph{data sets} from several \emph{sources} such as remote sensing data, spatial databases, sensor data, etc. 

\item \textbf{Preprocessing}.\quad These data sets are then processed to derive qualitative spatial observations associated with specific time points to hand over to the actual reasoning component. This preprocessing is done by the \emph{Temporal partitioning and Integration module}
responsible for partitioning the input data into time points and integrating data associated with the same time point including the resolution of spatial conflicts. 

\item \textbf{Qualitative abstraction}.\quad This module is itself supported by the \emph{Qualitative abstraction module} for performing the abstraction from quantitative to qualitative information and the \emph{Consistency checking module} for testing whether a qualitative spatial descriptions is consistent or contains logical contradictions. 

\item \textbf{Scenario and (Partial) Narrative Descriptions}.\quad The qualitative temporally-ordered observations generated by the  \emph{Temporal partitioning and Integration module} constitute the scenario and narrative descriptions, and serve as the input to the \emph{Reasoning module}, which embeds in itself one or more forms of (explanatory) reasoning capabilities. 

\item \textbf{Explanatory Reasoning}.\quad The reasoning component leads to the derivation of spatio-temporal knowledge that can be utilized by external services and application systems that directly interface with humans (e.g., experts, decision makers). 
Access can be provided by a \emph{Query Processing module} that allows for identifying high-level abducibles in the derived knowledge base.

\end{itemize}

%\noindent Building-up on the motivating scenario presented in Section \ref{sec:urban-dynamics}, we will 

In the following, we further explain the architecture and provide practical examples of the problems and solutions that we previously elaborated on in Section \ref{sec:spat-informatiks} in the context of a case-study. 

%Building-up on the motivating scenario presented in Section \ref{sec:urban-dynamics}, we now propose
%a conceptual architecture for high-level qualitative modelling and explanatory analysis in the
%geospatial domain and  provide practical examples of the problems / solutions that are presented in Section 3.

% 

%\noindent The architecture combines the different spatial information processing components addressed in Section 3. We explain the architecture together with running examples from the urban dynamics narrative of Section \ref{sec:urban-dynamics}. 

\newpage

\subsection{The Urban Dynamics Domain}

\begin{figure}[t]
\centering
\includegraphics[width=\textwidth]{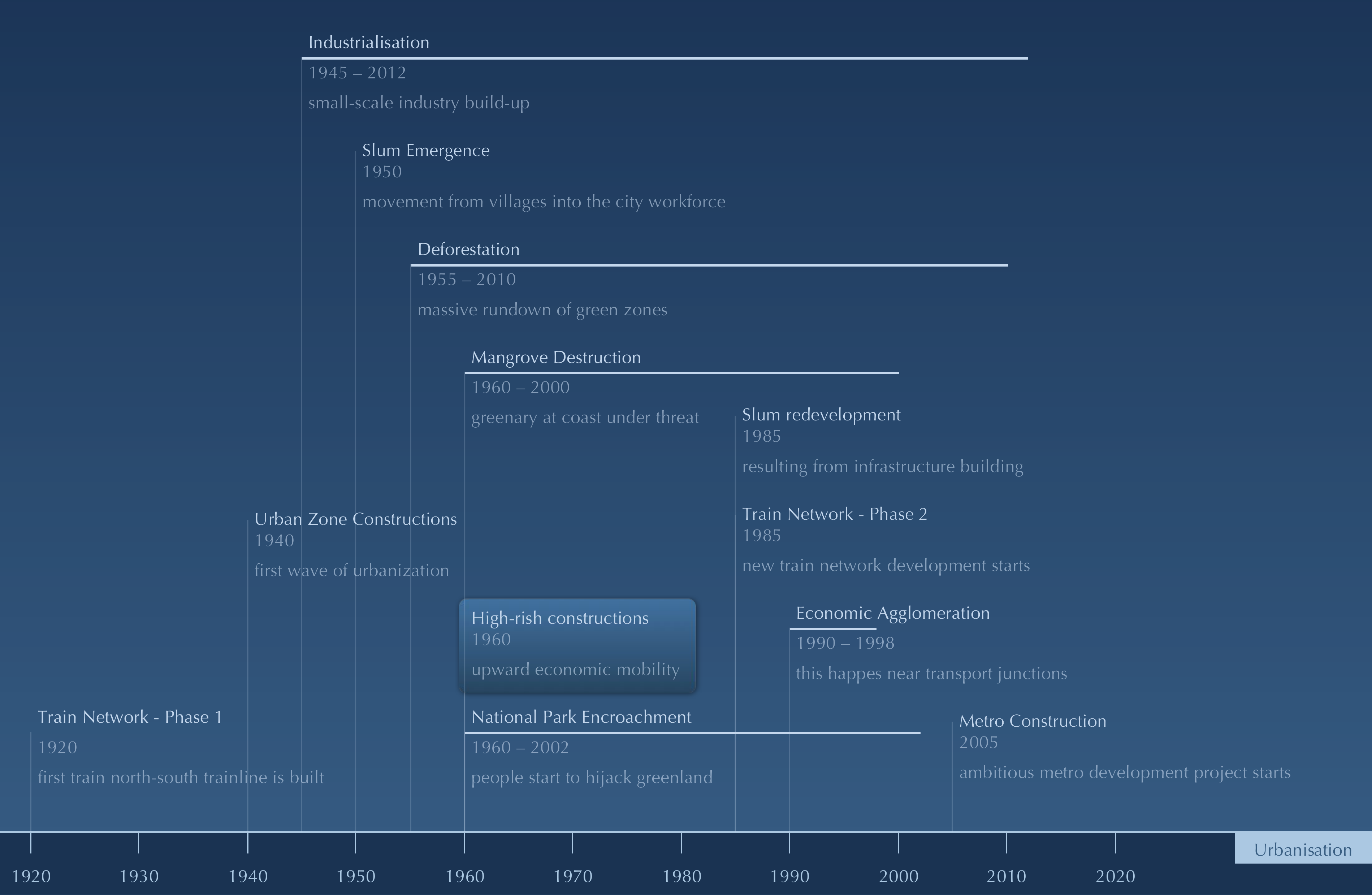}
\caption{{\sffamily An Urban Narrative} }
\label{fig:urban-narrative-bombay}
\end{figure}

\begin{figure*}[t]
\centering
\subfigure[]{\includegraphics[width=0.32\columnwidth]{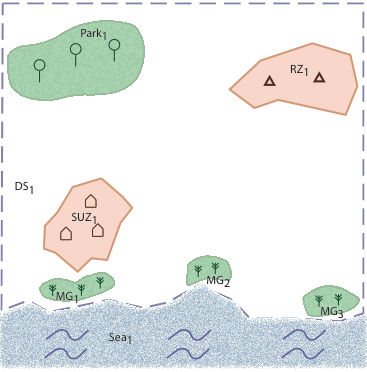}}%\hspace{0.5cm}
\subfigure[]{\includegraphics[width=0.32\columnwidth]{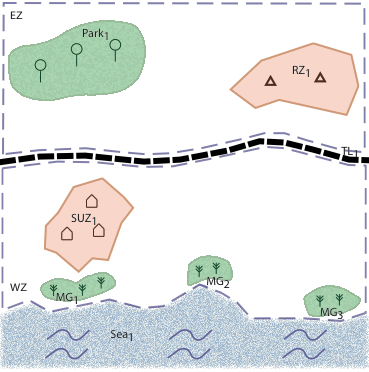}}
\subfigure[]{\includegraphics[width=0.32\columnwidth]{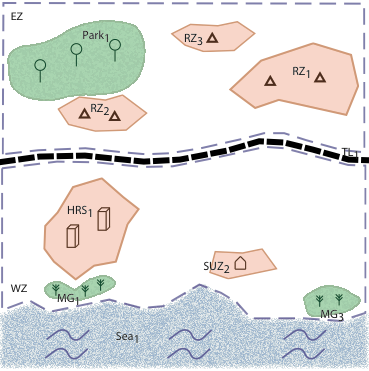}}%\hspace{0.5cm}

\subfigure[]{\label{fig:event-gis2_d}\includegraphics[width=0.32\columnwidth]{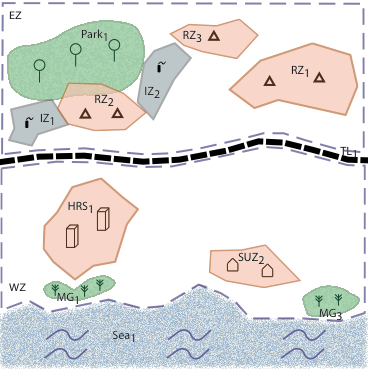}}
\subfigure[]{\includegraphics[width=0.32\columnwidth]{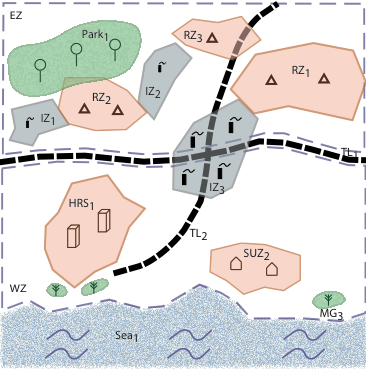}}%\hspace{0.5cm}
\subfigure[]{\includegraphics[width=0.32\columnwidth]{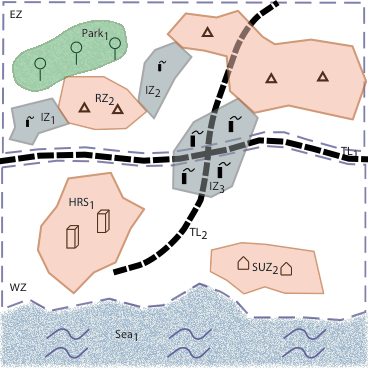}}

\subfigure{\includegraphics[width=0.45\columnwidth]{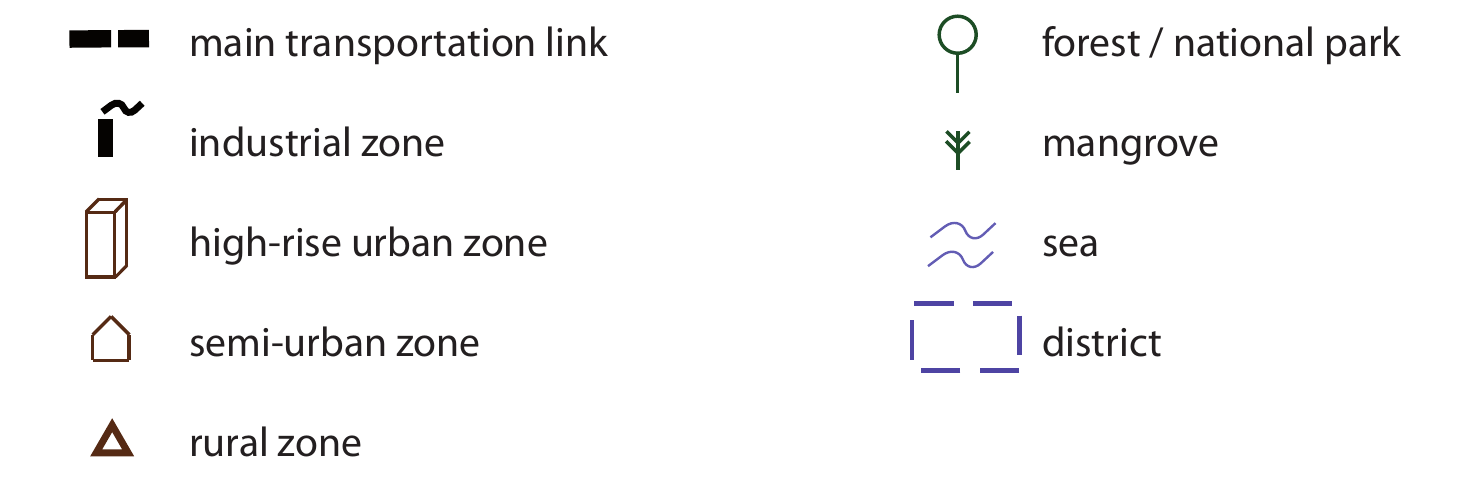}}
\caption{Abstract Spatio-Temporal Evolution of Urban Land-Use}
\label{fig:event-gis2}
\end{figure*}

Consider the following significantly trivialised \emph{urban narrative} (inspired from the dynamics of a real city); the textual description has also been illustrated as a timeline in Fig.~\ref{fig:urban-narrative-bombay}, and the object-level changes along with their temporal progression are illustrated in Fig.~\ref{fig:event-gis2}:

%the illustration in Fig. \ref{fig:event-gis2} provides one naive interpretation for the narrative described therein:

% \ref{fig:urban-narrative-bombay}:

\begin{center}
\colorbox{gray!18}{
\begin{minipage}{0.95\textwidth}
%\centering\small
{\footnotesize
\textbf{\small The Story of Bombaj}. The contemporary city of Mumbaj was once a collection of closely located tiny islands, surrounded by mangroves and other thick forests, along the coast of a huge landmass in the arabian sea. Guided by Bombaj's proximity to the sea and the Western world, humans deforested massive parts of the mangrove forests, and undertook reclamation of the islands to form one continuous entity connected to the huge landmass; this continuous entity came to be know as the city of Bombaj (subsequently Mumbaj).

\smallskip

\noindent \textbf{Migration}. Initially, there exists a thick Forest (subsequently becoming an endangered National Park) in the north-east, the Sea on the west,	and small pockets of human settlements by way of semi-urban / low-rise, and rural settlements. The idea of Bombaj ---its semantic characterization as a place--- is centred around these human settlements.

\smallskip

\noindent \textbf{Basic infrastructure setup}. Infrastructure gets established in an attempt to provide accessibility / coverage within the city: major transportation links gets established in addition to other initiatives. New conceptual zones gets established and place-names are formed based on the division created by the transportation link; primarily, two main zones that get created and persist even today come to be known as East-Bombaj and West-Bombaj.

 \smallskip

\noindent \textbf{Residential development}. New residential areas come up, and the West Zone, by virtue of its proximity to the sea, acquires a socio-economic privilege. Thereafter, powerful economic forces dictate that low income / low-rise areas, populated by recent immigrants (i.e., worker groups), come up in the lesser attractive East Zone. The city now starts to acquire its real character.

 \smallskip

\noindent \textbf{Industrialisation}. The East Zone, which is socio-economically perceived as being less attractive, starts to attract isolated pockets of industrialisation. New Industrial Zones get established in close proximity to the human settlements.

 \smallskip

\noindent \textbf{Infrastructure development}. Industrialisation, reinforced with further migration into the city, necessitates further infrastructure development. New transportation networks get built up, and major points of intersection / junctions get established / created -- these junctions acquire significance as point of \emph{economic agglomeration}. New industrial zones get established around these hubs of economic activity.

 \smallskip

\noindent \textbf{Rapid urban migration}. Large-scale deforestation of the thick forests and mangrove areas is undertaken as a result of high financial value of land in the West Zone, and massive population influx and re-development in the East Zone. Economic prosperity means that people in a lower income bracket are lifted, and there is a market for semi-urban settlements in the East Zone, which previously primarily consisted of rural settlements.

}
\end{minipage}}
\end{center}

%\begin{figure*}[t]
%\centering
%%\includegraphics[width=0.99\columnwidth]{pics/event-gis.pdf}
%\subfigure[]{\includegraphics[width=0.33\columnwidth]{casestudy_obs1.pdf}}%\hspace{0.5cm}
%\subfigure[]{\includegraphics[width=0.33\columnwidth]{casestudy_obs2.pdf}}
%\subfigure[]{\includegraphics[width=0.33\columnwidth]{casestudy_obs3.pdf}}%\hspace{0.5cm}
%
%\subfigure[]{\label{fig:event-gis2_d}\includegraphics[width=0.33\columnwidth]{casestudy_obs4.pdf}}
%\subfigure[]{\includegraphics[width=0.33\columnwidth]{casestudy_obs5.pdf}}%\hspace{0.5cm}
%\subfigure[]{\includegraphics[width=0.33\columnwidth]{casestudy_obs6.pdf}}
%
%\subfigure{\includegraphics[width=0.45\columnwidth]{legend.pdf}}
%\caption{Abstract Spatio-Temporal Evolution of Urban Land-Use}
%\label{fig:event-gis2}
%\end{figure*}

\begin{Example}[Process Analysis in the Urban Dynamics Domain]\label{exp:urban}
{\upshape
The urban dynamics domain consists of basic high-level processes such as industrialisation, migration, deforestation, relocation. The domain consists of the following entities: \msf{RuralZones}: \msf{rz_{1}}, \msf{rz_{2}}, \msf{rz_{3}}; \msf{ForestZones}: \msf{Park} (\msf{prk_{1}}), \msf{Mangroves}: \msf{mg_{1}}, \msf{mg_{2}}, \msf{mg_{3}}; \msf{SemiUrbanZones}: \msf{suz_{1}}, \msf{suz_{2}}; \msf{HighRiseZone}: \msf{hrs_{1}}; \msf{MunicipalZones}: \msf{district_{1}} (\msf{ds_{1}}), \msf{EastZone} (\msf{ez}), \msf{WestZone} (\msf{wz}), \msf{sea_{1}}; \msf{TransportationLinks}: \msf{tl_{1}}, \msf{tl_{2}};  \msf{IndustrialZones}: \msf{iz_{1}}, \msf{iz_{2}}, \msf{iz_{3}}.

%The domain consists of primitive spatial entities such as: {\sffamily ResidentialZones} ({\sffamily Rural, SemiUrban}, and {\sffamily HighRise}), {\sffamily ForestZones} ({\sffamily Parks, Mangroves}), {\sffamily IndustrialZones, TransportationLinks}, and {\sffamily MunicipalZones} (e.g., {\sffamily Districts}).

%
%
%\begin{itemize}
%	\item \msf{RuralZones}: \msf{rz_{1}}, \msf{rz_{2}}, \msf{rz_{3}}
%
%	 
%		
%	\item \msf{ForestZones}: \msf{Park} (\msf{prk_{1}}), \msf{Mangroves}: \msf{mg_{1}}, \msf{mg_{2}}, \msf{mg_{3}}
%
%	 
%	
%	\item \msf{SemiUrbanZone}: \msf{suz_{1}}, \msf{suz_{2}},
%	
%	 
%
%	\item \msf{HighRiseZone}: \msf{hrs_{1}}
%	
%	 
%	
%	\item \msf{MunicipalZones}: \msf{district_{1}} (\msf{ds_{1}}), \msf{EastZone} (\msf{ez}), \msf{WestZone} (\msf{wz}), \msf{sea_{1}}
%	
%	 
%	
%	\item \msf{TransportationLink}: \msf{tl_{1}}, \msf{tl_{2}}, 
%
%	 
%		
%	\item \msf{IndustrialZone}: \msf{iz_{1}}, \msf{iz_{2}}, \msf{iz_{3}}
%
%	
%\end{itemize}

\vspace{0.1in}

\noindent From the viewpoint of high-level narrative reasoning, the components of the theory that need to be formally modeled include: (1) domain constraints, spatial relationships (based on observational data), and other existential properties concerning the (appearance and disappearance) of objects, (2) process dynamics, or the laws of the domain, that determine occurrence criteria and effects for domain-specific events, (3) high-level abducibles that provide the causal rules that may be used as a basis of process extraction from a logically abduced model consisting only of domain-independent events.

%\begin{enumerate}
%	\item Domain Constraints -- Spatial Relationships and Existential Properties
%	
%	 
%	
%	\item High-level urban processes
%	
%	 
%	
%				\begin{enumerate}
%					\item Industrialisation
%					
%					 
%					
%					\item Migration
%					
%					 
%					
%					\item Deforestation
%					
%					 
%					
%					\item Relocation
%				\end{enumerate}
%	
%	 
%	
%	\item The Abduction tasks
%\end{enumerate}

\noindent The domain constraints and the high-level abducibles together constitute the overall specification, referred to as the domain-theory, for the urban dynamics domain. The high-level abducibles do not play a direct role in the narrative completion process, but are only required during a post-processing stage (as a means to query abduced / derived knowledge).\eod
%We denote the conjunction of all elements of the domain-theory as $\Sigma_{urban}$ (Section X; \eqref{domain}).
}
\end{Example}

\noindent Within an object and event-based GIS system, one may imagine high-level symbolic information to be available from a range of data sources. Performing explanatory analysis with this information first requires temporal partitioning, qualitative abstraction, and integration capabilities, presented next.

\subsection{Temporal Partitioning, Qualitative Abstraction, Integration}\label{sec:qual-integ}

%\commentJOW{[extend this part to cover and illustrate temporal partitioning, etc.]}

To illustrate the role of temporal partitioning and integration with qualitative abstraction, consistency checking, and conflict resolution (discussed previously in Sect.~\ref{sec:qualabs}) in our example, let us assume that the input data (a) stems from different sources and (b) each piece of information is associated with a timestamp specifying when the underlying measurement or observation has been performed. More specifically, let us say that  Source 1 provides information about different land use zones including parks,  residential zones, industrial zones, which are derived by analyzing aerial images, while Source 2 provides information about natural reservoirs, that is about the park and mangroves, stemming from a spatial database. All other information in our example comes from additional sources but does not play a role in this here.
Let us furthermore assume that the land use types are defined in a mutual exclusive way such that two different zones cannot overlap. 

Since all geometries we get from Source 1 and Source 2 are timestamped, the first thing that has to happen is  a partitioning of the total covered time period into
time intervals and by this inducing groups of spatial facts that are associated with each interval based on their timestamps. Each interval is represented by a time point $t_i$ in an ordered sequence of time points. 

\begin{figure}[t]
\centering
\subfigure[]{\label{fig:qual_a}\includegraphics[width=0.32\textwidth]{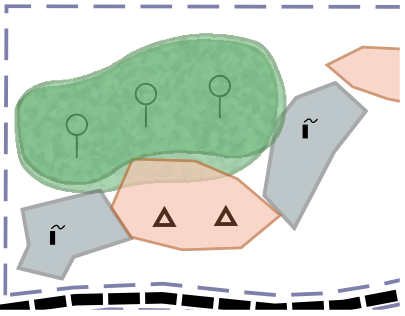}}%{qualification_example.pdf}}
\subfigure[]{\label{fig:qual_b}\includegraphics[width=0.32\textwidth]{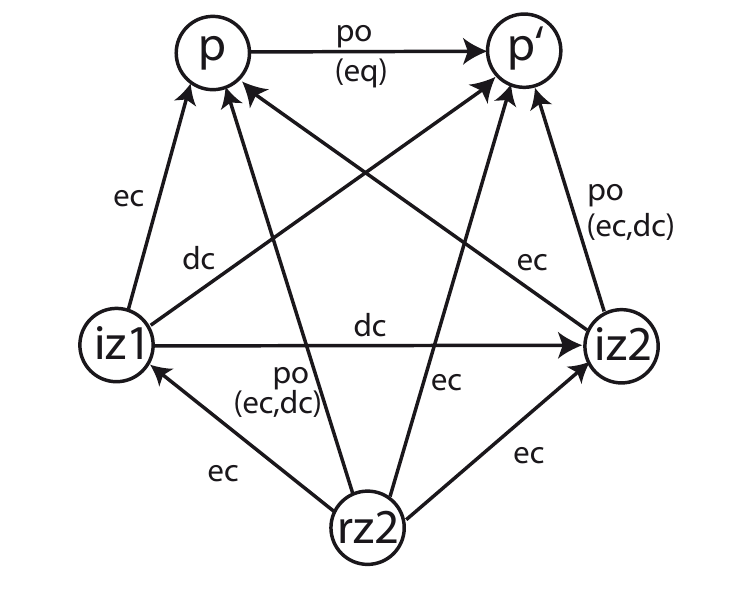}}
\subfigure[]{\label{fig:qual_c}\includegraphics[width=0.32\textwidth]{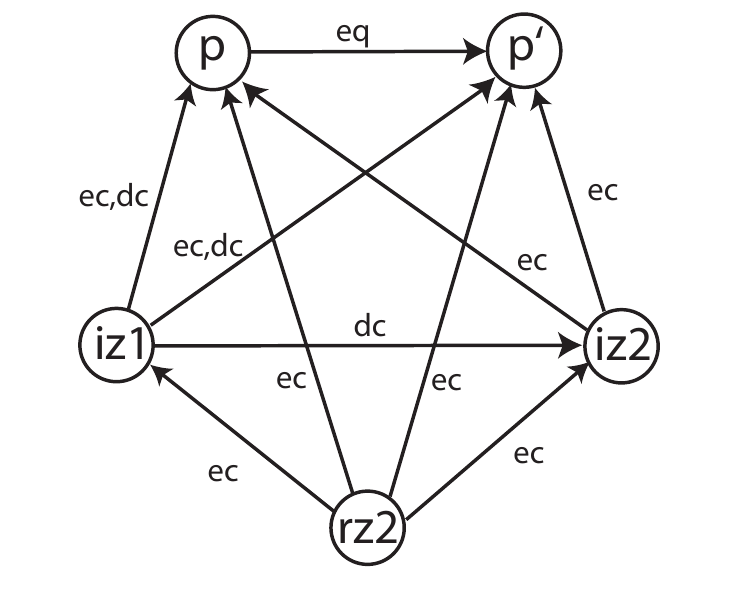}}
\caption{Qualitative abstraction of Fig.~\ref{fig:event-gis2_d} together with the integrity constraints results in an inconsistent qualitative model. The consistent model after resolving the conflicts.}
\label{fig:qual}
\end{figure}

Fig.~\ref{fig:qual_a} illustrates part of the combined information from all sources that after temporal partitioning fall into time period $t_4$ shown in Fig.~\ref{fig:event-gis2_d}. Source 1 and Source 2 both contain geo-referenced polygons for the park but this information does not match. To derive a consistent qualitative description for  time period $t_4$, 
the integration procedure follows Alg.~\ref{alg:merge} that takes the set of observed geometries $\mathcal{O}$ with  object identifiers  and a set 
of integrity constraints $\mathcal{IC}$ as input.
The first step is to use the Qualitative Abstraction module to translate the combined geometric data into qualitative spatial relations which results in a qualitative constraint network $Q$.\footnote{Alternatively, information for each data set could be qualified separately resulting in several constraint networks that have to be combined by a suitable merging operator.} Using the relations from the RCC-8 calculus this network looks as shown in Fig.~\ref{fig:qual_b} ($p$ and $p'$
represent the different geometries for the same park object). Next, the Consistency Checking module is used to test whether network $Q$ is consistent
and compliant with the integrity constraints. If this is the case, the result can directly be handed over as a qualitative observation for $t_4$ to the reasoning module. However,
as also shown in Fig.~\ref{fig:qual_b} this is not the case as integrity constraints are violated in three places. These violations are indicated by listing possible relations following from the integrity constraint in brackets below the original relation. The relation between $p$ and $p'$ should be eq simply because it is known that both represent the same object. The relation between $rz_2$ and $p$ should be either ec or dc because of the integrity constraints, and the same holds true for the relation between $p'$ and $iz_2$. Therefore, the qualitative conflict resolution component needs to be called to find a qualitative representation that is as close
as possible to the network from Fig.~\ref{fig:qual_b} but is overall consistent.

%Fig.~\ref{fig:qual_a} illustrates part of the combined information from all sources for time point $t_4$ shown in Fig.~\ref{fig:event-gis2_d}. Source 1 and source 2 both contain geo-referenced polygons
%for the park but this information does not match. The first step now is to qualify the geometric data from source 1 and 2 which results  in the qualitative constraint network $Q$.\footnote{Alternatively, information for each data set could be qualified separately resulting in several constraint networks that have to be combined by a suitable merging operator.} Using RCC-8 this network looks as shown in Fig.~\ref{fig:qual_b} ($p$ and $p'$
%represent the different geometries for the same park object). If network $Q$ is consistent
%and compliant with the integrity constraints, the result can directly be handed over as an observation to the reasoning module. However,
%as also shown in Fig.~\ref{fig:qual_b} this is not the case as integrity constraints are violated in three places indicated by listing possible relations following from the integrity constraint 
%in brackets below the original relation. The relation between $p$ and $p'$ should be eq simply because it is known that both represent the same object. The relation between $rz_2$ and $p$ should be either ec or dc because of our integrity constraint, and the same holds for the relation between $p'$ and $iz_2$. Therefore, the qualitative conflict resolution component needs to be called to find a qualitative representation that is as close
%as possible to the network from Fig.~\ref{fig:qual_b} but is overall consistent. 

To achieve the conflict resolution, an operator $\Lambda$ based on the idea of distance-based merging operators for qualitative spatial representations \cite{Condotta_Kaci_Schwind_08_A,Dylla_Wallgruen_07_Qualitative} is applied to $Q$.  
%and extended to take into account qualitative integrity constraints as well. This is typically straightforward as they simply add to the number of qualitative scenarios which are inconsistent. 
Our resolution operator $\Lambda$ is based on a distance measure $d(s,s')$ between
two scenarios over the same set of objects. It is computed
by simply summing up the distance of two base relations in the
conceptual neighborhood graph of the involved calculus given by
 $d_{B}(C_{ij}, C'_{ij})$ over all corresponding constraints  $C_{ij}, C'_{ij}$ in the input scenarios:

\begin{equation}
d(s,s')=\sum_{1 \leq i<j \leq m} \;d_{B}(C_{ij}, C'_{ij})
\label{eq:dss}
\end{equation}

The resolved network $\Lambda(Q)$ is then constructed by taking the union
of those scenarios that are consistent, compliant with the integrity constraints,
and have a minimal distance to $Q$ according
to $d(s,s')$\footnote{Taking the union here means we build a new network
by taking the union of all corresponding constraints.}:
\begin{equation}
\Lambda(Q) = \bigcup_{s \in S(Q)} s
\label{eq:def_delta}
\end{equation}
\noindent with
\begin{equation}
 S(Q) = \{s \in \cons{QCN} \;|\;  \forall s' \in \cons{QCN}:
 d(s',Q) \geq d(s,Q) \}
\label{eq:def_S}
\end{equation}
\noindent where $\cons{QCN}$ stands for the set of all scenarios that are consistent
and compliant with the integrity constraints. Following the approach described in \cite{Dylla_Wallgruen_07_Qualitative}, $\Lambda(Q)$ can be computed by incrementally
relaxing the constraints until at least one consistent scenario has been found.
This is illustrated in Alg.~\ref{alg:delta} where we assume that the function
relax(Q,i) returns the set of scenarios $s$ which have a distance $d(s',Q)=i$
to $Q$.

The result of applying the resolution operator to the network from Fig.~\ref{fig:qual_b} is shown in Fig.~\ref{fig:qual_c}: Both violations of integrity constraints have been resolved by assuming that instead of 'overlap' the correct relation is 'externally connected'. Interestingly, the resulting consistent qualitatively model contains two 
disjunctions basically saying that the relation between the park and $iz_1$ is either ec or dc.
This is a consequence of the fact that both qualitative models are equally close to the input
model such that it is not possible to decide between the two hypotheses. %\commentJOW{algorithm environment?}

%
%\noindent\begin{minipage}{\textwidth}
%  \centering
%    \begin{minipage}{.45\textwidth}
%    \centering
%    \captionof{algorithm}{test algorithm 1}
%    \label{alg:alg1}
%    \begin{algorithmic}
%    \WHILE{$N \neq 0$}
%    \IF{$N$ is even}
%    \STATE $X \Leftarrow X \times X$
%    \STATE $N \Leftarrow N / 2$
%    \ENDIF
%    \ENDWHILE
%    \end{algorithmic}
%  \end{minipage}
%  \begin{minipage}{.45\textwidth}
%    \centering
%    \captionof{algorithm}{test algorithm 2}
%    \label{alg:alg2}
%    \begin{algorithmic}
%    \WHILE{$N \neq 0$}
%    \IF{$N$ is even}
%    \STATE $X \Leftarrow X \times X$
%    \STATE $N \Leftarrow N / 2$
%    \ENDIF
%    \ENDWHILE
%    \end{algorithmic}
%  \end{minipage}
%  \captionof{figure}{Two algorithms side by side}
%  \label{fig:twoalg}
%\end{minipage}

%\commentMB{Delta typically denotes difference...also, i am using deltas throughout in my formalisation too....}

\begin{minipage}{\columnwidth}
%\centering
%\begin{algorithm}
%\colorbox{gray!18}{
\colorbox{gray!18}{
\begin{minipage}{0.4\columnwidth}\footnotesize
\captionof{algorithm}{\\Qualify+Merge($\mathcal{O}, \mathcal{IC}$)}
\begin{algorithmic}
\STATE $Q \gets \text{qualify}(\mathcal{O})$
\IF {$\neg$consistent$(Q, \mathcal{IC})$}
	\STATE $Q \gets \Lambda(Q, \mathcal{IC})$ \ENDIF
\RETURN $Q$
\end{algorithmic}
%\caption{QualifyAndMerge($\mathcal{O}, \mathcal{IC}$)}          % give the algorithm a caption
\label{alg:merge}
\vspace{0.45in} 

$ $
\end{minipage}}
%}
%\end{algorithm}
%\end{minipage}
%\begin{algorithm}
%\colorbox{gray!18}{
\colorbox{gray!18}{
\begin{minipage}{0.5\columnwidth}\footnotesize
\captionof{algorithm}{$\Lambda(Q, \mathcal{IC})$}
\begin{algorithmic}
\STATE $i \gets 0$, $N \gets \emptyset$
\WHILE {$N = \emptyset$}
\STATE $R \gets \text{relax}(Q,i)$
\FOR{$r \in R$}

\STATE \algorithmicif\ consistent$(r, \mathcal{IC})$ \algorithmicthen\ $N \gets N \cup r$ \algorithmicendif

%\IF {consistent$(r, \mathcal{IC})$}
%	\STATE $N \gets N \cup r $
%\ENDIF
\ENDFOR
\STATE $i \gets i +1$
\ENDWHILE
\RETURN $N$
\end{algorithmic}
%\caption{$\Lambda(Q, \mathcal{IC})$}          % give the algorithm a caption
\label{alg:delta}  
\end{minipage}}
%}
%\end{algorithm}
%}
\end{minipage}
%. and is supported for
%several qualitative relational formalisms in QSR toolboxes such as SparQ \cite{}
%and GQR \cite|{}. 

\subsection{Observation Data - Urban Narrative Description}

%\commentJOW{how can we make this part easier to follow for non-expert readers? more illustrations? real introduction to RAC? more examples in the text?}
%\commentJOW{would be good if the following part(s) would make more direct reference to the architecture to show where we are}

Given the qualitative abstraction, consistency detection and integration capabilities (as described so far), the objective now is to generate a temporally-ordered narrative of the processes that are reflected by the pre-processed data sets. Before exemplifying the narrative, some basic notation that we use from hereon follows:

%\commentMB{The conceptual model for our GIS can be clearly shown. This will include all the Sorts that we use in our logical axiomatisation., all arranged in a taxonomic form, whereever applicable.} 

\noindent \textbf{Notation}.\quad {We use a first-order many-sorted language ($\mathcal{L}$) with the following alphabet: $\{\neg,$ $\wedge,$ $\vee,$ $\forall,$ $\exists,$ $\supset,$ $\equiv\}$. There are sorts (and corresponding variables)  for: \emph{events} --- $\Theta$ $=$  $\{\theta_{1},~\theta_{2},~\dots,~\theta_{n}\}$, \emph{time-points} --- $\mathcal{T}$ $=$  $\{t_{1},~t_{2},~\dots,~t_{i}\}$, \emph{spatial objects} --- $\mathcal{O}$ $=$  $\{o_{1},~o_{2},~\dots,~o_{j}\}$, \emph{regions} of space --- $\mathcal{S}$ $=$  $\{s_{1},~s_{2},~\dots,~s_{k}\}$, and a function symbol $(extent$: $\mathcal{O}$ $\rightarrow$ $\mathcal{S})$ that determines the time-dependent spatial location of an entity. We only consider binary topological relationships of spatially extended regions in space. However, the theory encompasses point and line-segment based spatial calculi of arbitrary arity. 

{Let \RelSpace\  $=$ $\{r_{1},~r_{2},\ldots,r_{n}\}$ denote a reified $n$-ary qualitative spatial relationship space over an arbitrary qualitative spatial calculus. $\Phi$ $=$ $\{\phi_{1},~\phi_{2},~\dots,~\phi_{l}\}$ is the set of propositional and functional fluents, e.g., $\phi_{sp}(o_{i},~o_{j})~\in~\Phi$ is a functional fluent denoting the spatial relationship from \RelSpace\ between objects $o_{i}$ and $o_{j}$. The special event-predicate $tran(r_{i},~o_{i},~o_{j})$ $\in$  $\Theta$ denotes a transition to a spatial relation $r_{i}$ between objects $o_{i}$ and $o_{j}$. Finally, the ternary {$Holds(\phi,~r,~t)~\subset~[\Phi \times \RelSpace \times \mathcal{T}]$} predicate is used for temporal property exemplification, and {$Happens(\theta,t)$} denotes event occurrences. For notational convenience, we use the following syntactic sugar for fluent terms ($\phi$):$~P(\phi([~x{i},\ldots~x{n}~]))$ transforms to $[~P(\phi(x_{i}))~\wedge\ldots\wedge~P(\phi(x_{n}))~]$. The use of $[~x{i},\ldots~x{n}~]$ with events terms ($\Theta$) represents a vector argument, and is interpreted differently.}

%\commentJOW{That part probably needs to be explained more for non-expert readers (e.g. add examples)}
}

%\begin{enumerate}{\small
%	\item [(i)] $\Omega~\subset~\Omega^{'}$
%	
%	\item [(ii)] \textbf{if}: $\Omega[s]~\models~(\exists~o_{i},~o_{j},~\gamma).~[Holds(\phi_{space}(o_{i},~o_{j}),~\gamma,~s)]$\\\textbf{then}: $\Omega^{'}[s]~\models~(\exists~o_{i},~o_{j},~\gamma).~[Holds(\phi_{space}(o_{i},~o_{j}),~\gamma,~s)]$
%
%	\item [(iii)] \textbf{if}: $\Omega[s]~\models~(\exists~o_{i},~\gamma).~[Holds(\phi_{exists}(o_{i}),~\gamma,~s)]$\\\textbf{then}: $\Omega^{'}[s]~\models~(\exists~o_{i},~\gamma).~[Holds(\phi_{exists}(o_{i}),~\gamma,~s)]$
%	
%	\item [(iv)] \textbf{if}: $\Omega[s]~\models~(\exists~o_{i},~\gamma).~[Holds(\phi_{physical}(o_{i}),~\gamma,~s)]$\\$\textbf{then}: \Omega^{'}[s]~\models~(\exists~o_{i},~\gamma).~[Holds(\phi_{physical}(o_{i}),~\gamma,~s)]$
%	}
%	\end{enumerate} 

%
%Condition (i) in Definition \ref{def:monotonic-ext} stipulates that $\Omega^{'}$ contains additional facts than are present in $\Omega$. Conditions (ii-iv) express that none of the (spatial) entailments of $\Omega^{'}$ invalidate or refute the information contained in $\Omega$. Note that the notation $\Omega[s]$, read as the result of substituting the \emph{situational} argument in all instances of the $Holds$ predicate to situation $s$. This is necessary since the (monotonic) extension derivation is always applicable on a per-situation basis. 

\noindent The temporally-ordered (5) observations in \eqref{observations} represent {a formal description of the output of the Temporal Reasoning and Integration  module for time points $t_1$ to $t_6$ corresponding to the snapshots depicted in the six images in Fig.~\ref{fig:event-gis2}}.  They provide the grounding for a narrative description of the urban dynamics scenario under consideration:\footnote{To save space, the represented observations only show those facts that become true/false in a given observation. Also, some implicit facts are omitted.}

\begingroup
\everymath{\scriptstyle}
%\scriptsize
\begin{subequations}\label{observations}{%\footnotesize
\begin{equation}\label{abd-task-obs-t1}
\begin{aligned}
\Psi_{1}~\equiv~[~&Holds(\exts{[prk_{1},~rz_{1},~mg_{1},~mg_{2},~mg_{3},~\msf{ds_{1}},~\msf{suz_{1}},~\msf{sea_{1}}]},~true,~t_{1})~\wedge\\&\vec{o}=[\msf{prk_{1},~rz_{1},~mg_{1},~mg_{2},~mg_{3},~suz_{1}},~\msf{sea_{1}}],~Holds(\phi_{top}(\vec{o},~\vec{o}),~dc,~t_{1})~\wedge\\&Holds(\phi_{top}([\msf{prk_{1},~rz_{1},~mg_{1},~mg_{2},~mg_{3},~suz_{1}}],~\msf{ds_{1}}),~ntpp,~t_{1})~\wedge\\&Holds(\phi_{top}(\msf{ds_{1},~sea_{1}}),~ec,~t_{1})]
\end{aligned}
\end{equation}
\begin{equation}\label{abd-task-obs-t2}
\begin{aligned}
\Psi_{2}~\equiv~[~&Holds(\exts{[tl_{1},~ez,~wz]},~true,~t_{2})~\wedge\\&Holds(exists(\msf{ds_{1}}),~false,~t_{2})~\wedge~Holds(\phi_{top}(ez,~wz),~dc,~t_{2})~\wedge\\&Holds(\phi_{top}([\msf{prk_{1}},~\msf{rz_{1}}],~ez),~ntpp,~t_{2})~\wedge\\&Holds(\phi_{top}([\msf{suz_{1}},~\msf{mg_{1}},~\msf{mg_{2}},~\msf{mg_{3}}],~\msf{wz}),~ntpp,~t_{2})~\wedge\\&Holds(\phi_{top}([\msf{ez},~\msf{wz}],~\msf{tl_{1}}),~ec,~t_{2})]
\end{aligned}
\end{equation}
\begin{equation}\label{abd-task-obs-t3}
\begin{aligned}
\Psi_{3}~\equiv~[~&Holds(\exts{[rz_{2},~rz_{3},~suz_{2}]},~true,~t_{3})~\wedge\\&Holds(exists(\msf{mg_{2}}),~false,~t_{3})~\wedge~Holds(\phi_{top}([\msf{rz_{2}},~\msf{rz_{3}}],~ez),~ntpp,~t_{3})~\wedge\\&Holds(\phi_{top}(\msf{rz_{2}},~\msf{rz_{3}}),~dc,~t_{3})~\wedge~Holds(exists(\msf{suz_{2}}),~true,~t_{3})~\wedge\\&Holds(\phi_{top}(\msf{suz_{2},~wz}),~ntpp,~t_{3})~\wedge\\&Holds(\phi_{top}([\msf{suz_{1}},~\msf{mg_{1},~\msf{mg_{3}}}],~\msf{suz_{2}}),~dc,~t_{3})]
\end{aligned}
\end{equation}
\begin{equation}\label{abd-task-obs-t4}
\begin{aligned}
\Psi_{4}~\equiv~[~&Holds(\exts{[iz_{1},~iz_{2}]},~true,~t_{4})~\wedge\\&Holds(\phi_{top}([\msf{iz_{1}},~\msf{iz_{2}}],~\msf{ez}),~ntpp,~t_{4})~\wedge~Holds(\phi_{top}(\msf{rz_{2},~prk_{1}}),~po,~t_{4})]
\end{aligned}
\end{equation}
\begin{equation}\label{abd-task-obs-t5}
\begin{aligned}
\Psi_{5}~\equiv~[~Holds(\exts{[tl_{2},~iz_{3}]},~true,~t_{5})~~\wedge~Holds(\phi_{top}(\msf{[ez,~wz]},~\msf{iz_{3}}),~po,~t_{4})]
\end{aligned}
\end{equation}
\begin{equation}\label{abd-task-obs-t6}
\begin{aligned}
\Psi_{6}~\equiv~[~Holds(\exts{[mg_{3},~mg_{1}]},~false,~t_{6})~\wedge~Holds(\phi_{top}(\msf{rz_{2},~prk_{1}}),~ec,~t_{6})]
\end{aligned}
\end{equation}}
\end{subequations}
\begin{equation}\label{abd-task-obs-temp-ord}
\begin{aligned}
t_{1}~<~t_{2}~<~t_{3}~<~t_{4}~<~t_{5}~<~t_{6}
\end{aligned}
\end{equation}
\endgroup

%\commentJOW{This section either needs to be illustrated using a concrete example; alternatively: maybe this is a less important detail that we can leave out entirely?}

\subsubsection{ Partial Description and Extension}\label{sec:partial-desc-exten}
When spatial relationships ($\Phi_{space}$) between some objects are omitted, a complete description (with disjunctive labels) can be derived on the basis of the composition theorems (cmp.~Section \ref{sec:caxioms}), and other integrity constraints for the spatial domain under consideration. In the following, we elaborate the treatment for a partial situation description using the notion of a monotonic extension. Let {$\Omega$} denote a partial spatial state description consisting of facts expressed using the ternary {$Holds$} predicate. The following notion of a `monotonic extension' is necessary:

%{\small$O~\equiv~\{o_{i},~o_{j},~\ldots,~o_{n}\}$} denotes the set of spatial entities and {\small$\Phi^{arity}_{type}$ $\equiv$ $\{\phi^{1}_{exists},~\phi^{2}_{space},~\phi^{1}_{physical}\}$} denotes the set of spatial relationships that may exist between the spatial objects.

\begin{Definition}[Monotonic Extension]\label{def:monotonic-ext}{
The monotonic extension of a partial spatial state description $\Omega$ is another description $\Omega^{'}$ such that  $\Omega~\subset~\Omega^{'}$, and all semantic entailments with respect to the spatial information present in $\Omega$ are preserved in $\Omega^{'}$. For lack of space, we leave out the formal definition for the monotonicity condition.}\eod
\end{Definition}

\noindent The observation set ($\Psi$) constitutes the input to a high-level reasoning component. In the section to follow, we develop the formal domain-independent spatial theory that is used as the basis of reasoning, or in this specific case, for practical abductive reasoning in GIS.

%first idea:
%\begin{itemize}
%\item source1 provides all land use zone (park, mangroves, rural/semi-urban/high-rise/industrical zones)
%\item source 2 provides all natural zones (park, mangroves, sea)
%\item (optional source 3 provides administrative zones (westzone, eastzone))
%\item integrity constraint1: different land use zones cannot overlap
%\item integrity constraint2: nothings should overlap with the sea
%\item now take Fig. d and e from the example
%\item it should be possible to construct something about the overlap of the forest/ park and the rural zone which violate IC1 including different geometries given for the park in source1 and source 2
%\end{itemize}

%Fig.~\ref{fig:int}
%
%\begin{figure}[t]
%\centering
%\includegraphics[width=0.5\textwidth]{integration_example.pdf}
%\caption{... integration example ...}
%\label{fig:int}
%\end{figure}

%%%%%%%%%%%%%%%%%%%%%%%%%%%%%%%%%%%%%%%%%%%%%%%%
\subsection{A Domain-Independent Spatial Theory}
%%%%%%%%%%%%%%%%%%%%%%%%%%%%%%%%%%%%%%%%%%%%%%%%

From a dynamic spatial systems perspective, a domain-independent spatial theory (most crucially) consist of: (1) high-level axiomatic aspects that characterise a qualitative theory of spatial change; (2) phenomenal aspects inherent to dynamic (geo)spatial systems \citep{bhatt:scc:08}. We adapt this general notion for the domain of geospatial dynamics.

%% $=$ $\{dc,~ec,~po,~eq,~tpp,~ntpp,~tpp^{-1},~ntpp^{-1}\}$}. 
\subsubsection*{\bfseries I. Axiomatic Characterisation of a Spatial Theory} Many spatial calculi exist, each corresponding to a different aspect of space. Here, it suffices to think of one spatial domain, e.g., topology, with a corresponding mereotopological axiomatisation by way of the binary relationships of the RCC-8 calculus. From an axiomatic viewpoint, a spatial calculus defined with respect to an arbitrary relationship space \RelSpace\ has some general properties (described below in {\small(P1--P5)}). For any spatial calculus, it can be assumed that {\small(P1--P5)} are known apriori, i.e., these are the intensional properties that define the constitution of the calculus. To realize a domain-independent spatial theory that can be used for reasoning (e.g., spatio-temporal abduction) across different dynamic (geospatial) domains, it is necessary to preserve the high-level axiomatic semantics of these generic properties, and implicitly, the underlying algebraic properties, that collectively constitute a qualitative spatial calculus. A domain-independent spatial theory ({\small$\Sigma_{space}$}) may be obtained by axiomatising {\small(P1--P5)} as follows:\footnote{The variables $r_{1},...r_{n}$ correspond to qualitative spatial relationships pertaining to spatial calculus being modelled; e.g., in the current case, these could be interpreted as the RCC-8 topological relations \{$dc,~ec,~po,~eq,~tpp,~ntpp,~tpp^{-1},~ntpp^{-1}$\} }

\subsubsection*{(P1--P2). Basic Calculus Properties ($\Sigma_{cp}$).}\label{sec:MiscProp-gen}
$\mathcal{R}$ has the \emph{jointly exhaustive \& pair-wise disjoint} {\small(JEPD)} property, i.e., for any two entities in $\mathcal{O}$, one and only one spatial relationship from $\mathcal{R}$ holds in a given situation. The joint-exhaustiveness can be expressed using $n$ ordinary state constraints of the form in (6a).%\eqref{JE-Const-gen}.

%{
%\begingroup
%\everymath{\scriptstyle}
%\scriptsize
\begin{subequations}\label{JE-PD-CONST}
\begin{equation}\label{JE-Const-gen}
\begin{aligned}
&(\forall t).~\neg[Holds(\phi_{sp}(s_{1},~s_{2}),~r_{1},~t)~\vee~Holds(\phi_{sp}(s_{1},~s_{2}),~r_{2},~t)~\vee\cdots\\&\vee~Holds(\phi_{sp}(s_{1},~s_{2}),~r_{n-1},~t)]~\supset~Holds(\phi_{sp}(s_{1},~s_{2}),~r_{n},~t)
\end{aligned}
\end{equation}
\begin{equation}\label{PD-Const-gen}
\begin{aligned}
(\forall t).~\neg~[Holds(\phi_{sp}(s_{1},~s_{2}),~r_{1},~t)~\wedge~Holds(\phi_{sp}(s_{1},~s_{2}),~r_{2},~t)]
\end{aligned}
\end{equation}
\end{subequations}

\begin{subequations}
\begin{equation}\label{symmetry-gen}
\begin{aligned}
(\forall t).~[Holds(\phi_{sp}(s_{i},~s_{j}),~r,~t)~\supset~Holds(\phi_{sp}(s_{j},~s_{i}),~r,~t)]
\end{aligned}
\end{equation}
\begin{equation}\label{asymmetry-gen}
\begin{aligned}
(\forall t).~[Holds(\phi_{sp}(s_{i},~s_{j}),~r,~t)~\supset~\neg~Holds(\phi_{sp}(s_{j},~s_{i}),~r,~t)]
\end{aligned}
\end{equation}
\end{subequations}
%\endgroup
%\normalsize}

\noindent Similarly, $[n(n~-~1)/2)]$ constraints of the form in (6b) are sufficient to express the pair-wise disjointness of $n$ relations. % \eqref{PD-Const-gen} 
Other miscellaneous properties such as the symmetry (7a) %\eqref{symmetry-gen} 
\& asymmetry (7b) %\eqref{asymmetry-gen} 
of the base relations too can be expressed using ordinary constraints.

\subsubsection*{(P3). Conceptual Neighbourhood ($\Sigma_{cn}$).}
As we mentioned, the primitive relations of a qualitative calculus  have a \emph{continuity structure}, referred to as its conceptual neighbourhood (CND) (see \cite{Freksa_91_Conceptual,Egenhofer_Al-Taha_92_Reasoning,Galton00:qualitative}), which determines the direct, continuous changes in the quality space (e.g., by deformation and / or translational motion). The binary (reflexive) predicate $neighbour(r,~r')$ denotes a continuity relation between relations $r$ and $r'$. 
{
%\tiny\setlength\abovedisplayskip{1pt plus 3pt minus 7pt}
\setlength\belowdisplayskip{1pt plus 3pt minus 7pt}
\setlength\abovedisplayshortskip{1pt plus 3pt}
\setlength\belowdisplayshortskip{1pt plus 3pt minus 4pt}
\begingroup
\everymath{\scriptstyle}
%\scriptsize
\begin{equation}\label{transition-axiom-gen-1}
\begin{aligned}
Poss(&tran(r,~o_{i},~o_{j}),~t)~\equiv~[\{extent(o_{i},~t)~=~s_{i}~\wedge~extent(o_{j},~t)\\=~s_{j}\}&\wedge~\{(\exists~r')~Holds(\phi_{sp}(~s_{i},~s_{j}),~r',~t)~\wedge~neighbour(r,~r')\}]\end{aligned}
\end{equation}
\endgroup
\normalsize}

\noindent Continuity constraints are only useful in scenarios involving spatio-temporal continuity (e.g., diffusive phenome, movement in (geo)space), and may serve a useful role in spatio-temporal interpolation, and prediction, especially in scenarios where the available data is incomplete and/or error-prone.

\subsubsection*{(P4). Composition Theorems ($\Sigma_{ct}$).}\label{sec:caxioms}
From an axiomatic viewpoint, a spatial calculus defined on $\mathcal{R}$ is (primarily) based on the derivation of a set of composition theorems between the JEPD set $\mathcal{R}$. In general, for a (spatial, temporal or spatio-temporal) calculus consisting of $n$ JEPD relationships (i.e., $n$ $=$ $|\mathcal{R}|$), $[n~\times~n]$ compositions are precomputed. Each of these composition theorems is equivalent to an ordinary state constraint (9), %(\ref{Ordinary-Constraint-gen})
which every spatial situation description should satisfy.

{
%\tiny
\setlength\abovedisplayskip{1pt plus 3pt minus 7pt}
\setlength\belowdisplayskip{1pt plus 3pt minus 7pt}
\setlength\abovedisplayshortskip{1pt plus 3pt}
\setlength\belowdisplayshortskip{1pt plus 3pt minus 4pt}
\begingroup
\everymath{\scriptstyle}
%\scriptsize
\begin{equation}\label{Ordinary-Constraint-gen}
\begin{aligned}
(\forall t).~[Holds(\phi_{sp}(s_{1},~s_{2})&,~r_{1},~t)~\wedge~Holds(\phi_{sp}(s_{2},~s_{3}),~r_{2},~t)\\&\supset~Holds(\phi_{sp}(s_{1},~s_{3}),~r_{3},~t)]
\end{aligned}
\end{equation}
\endgroup

}

\subsubsection*{(P5). Axioms of Interaction ($\Sigma_{ai}$).}
{
These are applicable when more than one spatial calculus is modelled in a non-integrated manner (i.e., with independent composition theorems). These axioms explicitly characterize the relative entailments between inter-dependent aspects of space, e.g., topology and size. For instance, a spatial relationship of one type may directly entail or constrain a spatial relationship of another type (10a). Such axioms could also possibly be compositional in nature, making it possible to compose spatial relations pertaining to two different aspects of space in order to yield a spatial relation of either or both spatial types used in the composition (10b). %(\ref{axiom-interaction-gen-2}).
}

{
%\tiny
\setlength\abovedisplayskip{1pt plus 3pt minus 7pt}
\setlength\belowdisplayskip{1pt plus 3pt minus 7pt}
\setlength\abovedisplayshortskip{1pt plus 3pt}
\setlength\belowdisplayshortskip{1pt plus 3pt minus 4pt}
\begingroup
\everymath{\scriptstyle}
%\scriptsize
\begin{subequations}\label{axioms-interaction-gen}
\begin{equation}\label{axiom-interaction-gen-1}
\begin{aligned}
(\forall t).~[Holds(\phi_{sp1}(s,~s^{'}),~r,~t)~\supset~Holds(\phi_{sp2}(s,~s^{'}),~r^{\prime},~t)]
\end{aligned}
\end{equation}
\begin{equation}\label{axiom-interaction-gen-2}
\begin{aligned}
(\forall  t).~[Holds(\phi_{sp1}(s_{i},~s_{j}),~r^{\prime}_{sp1}&,~t)~\wedge~Holds(\phi_{sp2}(s_{j},~s_{k}),~r^{\prime}_{sp2},~t)\\&\supset~Holds(\phi_{sp}(s_{i},~s_{k}),~r_{sp},~t)]
\end{aligned}
\end{equation}
\end{subequations}\endgroup
}

\subsubsection*{\bfseries II. Phenomenal Aspects -- Geospatial Events ($\Sigma_{ph}$)}
Here, we define our exemplary interpretation for the geospatial events based on the semantic characterisation in Section \ref{sec:occurrences}. The definitions also utilise additional (binary) boolean function symbols ---$merge\_cond$ and $split\_cond$--- that extralogically define (e.g., in a geometric sense) the conditions needed to check for events.\footnote{Outside of the logical theory, the merge and split conditions are basically geometric operations that may have an arbitrary characterization.}

\subsubsection*{II.1\quad  Appearance and Disappearance}
This is the simplest case where the existential status of an object undergoes a change \eqref{app-disapp-ext-1}. Here, we assume that identity is handled outside of the reasoning framework.

\begingroup
\everymath{\scriptstyle}
%\scriptsize
\begin{subequations}\label{app-disapp-ext-1}%\small
\begin{equation}\label{disappearance-effect-exist}
\begin{aligned}
(\forall o,~s).~[Oc&curs(disappearance(o),~s)~\supset\\&Caused(exists(o),~false,~Result(disappearance(o),~s))]
\end{aligned}
\end{equation}
\begin{equation}\label{appearance-effect-exist}
\begin{aligned}
(\forall o,~s).~[Oc&curs(appearance(o),~s)\supset\\&Caused(exists(o),~true,~Result(appearance(o),~s))]
\end{aligned}
\end{equation}
\begin{equation}\label{disappearance-precon-exist}
\begin{aligned}
(\forall o,~s).~[&Poss(disappearance(o),~s)~\equiv~Holds(exists(o),~true,~s)]
\end{aligned}
\end{equation}
\begin{equation}\label{appearance-precon-exist}
\begin{aligned}
(\forall o,~s).~[&Poss(appearance(o),~s)~\equiv~Holds(exists(o),~false,~s)]
\end{aligned}
\end{equation}
\end{subequations}
\endgroup

\subsubsection*{II.2\quad  Split}
A split involves an existing object that disintegrates into a set of $n$ previously non-existing objects \eqref{split-ext-1}:

\begingroup
\everymath{\scriptstyle}
%\scriptsize
\begin{subequations}\label{split-ext-1}%\small
\begin{equation}\label{split-effect}
\begin{aligned}
(\forall o,~s).~&[~Occurs(split(o,~[o_{i},\ldots,o_{n}]),~s)~\supset\\&Caused(exists([o_{i},\ldots,o_{n}]),~true,~Result(split(o,~[o_{i},\ldots,o_{n}]),~s))~\wedge\\&Caused(exists(o),~false,~Result(split(o,~[o_{i},\ldots,o_{n}]),~s))~]
\end{aligned}
\end{equation}
\begin{equation}\label{split-trigger}
\begin{aligned}
(\exists~s^{'})(\exists~&o,~[o_{i},\ldots,o_{n}]).~[~s^{'}~<~s~\wedge~Holds(exists([o_{i},\ldots,o_{n}]),~true,~s)~\wedge\\&Holds(exists([o_{i},\ldots,o_{n}]),~false,~s^{'})\\&\neg~Holds(exists(o),~false,~s^{'})~\wedge~Holds(exists(o),~false,~s)~\wedge\\&\wedge~r_{i}=extent([o_{i},\ldots,o_{n}],~s)~\wedge~r_{j}=extent(o,~s^{'})~\wedge\\&split\_cond(r_{i},~r_{j})~]~\supset~Occurs(split(o,~[o_{i},\ldots,o_{n}]),~s)
\end{aligned}
\end{equation}
\end{subequations}
\endgroup

\subsubsection*{II.3\quad  Merge}
A merge event \eqref{split-merge-ext-1} is (formalized as) a dual of a split event:
\begingroup
\everymath{\scriptstyle}
%\scriptsize
\begin{subequations}\label{split-merge-ext-1}%\small
\begin{equation}\label{merge-effect}
\begin{aligned}
(\forall o,~s).~&[Occurs(merge([o_{i},\ldots,o_{n}],~o),~s)~\supset\\&Caused(exists([o_{i},\ldots,o_{n}]),~false,~Result(merge([o_{i},\ldots,o_{n}],~o),~s))~\wedge\\&Caused(exists(o),~true,~Result(merge([o_{i},\ldots,o_{n}],~o),~s))]
\end{aligned}
\end{equation}
\begin{equation}\label{merge-trigger}
\begin{aligned}
(\exists~s^{'})(\exists~&[o_{i},\ldots,o_{n}],~o).~[~s^{'}~<~s~\wedge~Holds(exists([o_{i},\ldots,o_{n}]),~false,~s)~\wedge\\&Holds(exists([o_{i},\ldots,o_{n}]),~true,~s^{'})~\wedge\\&\neg~Holds(exists(o),~true,~s^{'})~\wedge~Holds(exists(o),~true,~s)~\wedge\\&\wedge~r_{i}=extent([o_{i},\ldots,o_{n}],~s^{'})~\wedge~r_{j}=extent(o,~s)~\wedge\\&merge\_cond(r_{i},~r_{j})~]~\supset~Occurs(merge([o_{i},\ldots,o_{n}],~o),~s)
\end{aligned}
\end{equation}
\end{subequations}
\endgroup

%\commentMB{comment bout: causes to effects, effects to causes -- these are needed since they are useful for the completeness of the underlying meta theory}

\noindent  {Let {\small$\Sigma_{space}$ $\equiv_{def}$ $[\Sigma_{cp}~\cup~\Sigma_{cn}~\cup~\Sigma_{ct}~\cup~\Sigma_{ai}~\cup~\Sigma_{ph}]$} denote a domain-independent spatial theory that is based on the axiomatisations encompassing {\small(P1--P5)}, and the phenomenal aspects in {\small$\Sigma_{ph}$}.}

\subsubsection*{\bf Physically Plausible Scenarios.}\label{sec:global-const}%\subsection{Complete N-Clique Descriptions}
Corresponding to each spatial situation (e.g., within a hypothetical situation space; Fig. \ref{fig:narrative-branching}), there exists a situation description that characterizes the spatial state of the system. It is necessary that the spatial component of such a state be a `complete specification', possibly with disjunctive information. For $k$ (binary) spatial calculi being modelled, the initial situation description involving $n$ domain objects requires a complete specification with {\small$[n(n~-1)/2]$} spatial relationships for each calculus.\footnote{Precisely, under a unique names assumption for the fluents in {\small$\Phi$} (i.e., {\small$[\phi_{sp}(o_{i},~o_{j})$ $\neq$ $\phi_{sp}(o_{j},~o_{i})]$}), static spatial configurations actuality consist of {\small$[(k~\times~[n(n~-~1)~/~2])~\times~2]$} unique functional fluents.}

\begin{Definition}[$\mathcal{C}$-Consistency]\label{def:composition-consistency}
{A scene description is $\mathcal{C}$-Consistent, i.e., compositionally consistent, if the state or spatial situation description corresponding to the situation satisfies all the composition constraints of every spatial domain (e.g., topology, orientation, size) being modelled, as well as the relative entailments as per the axioms of interaction among inter-dependent spatial calculi when more than one spatial calculus is modelled.}
\end{Definition}

\noindent From the viewpoint of model elimination of narrative descriptions during an (abductive) explanation process,~$\mathcal{C}$-\emph{Consistency} of scenario descriptions is a key (contributing) factor determining the commonsensical notion of the \emph{physically realizability} of the (abduced) scenario completions.\footnote{\citet{bhatt:scc:08} show that a standard completion semantics with \emph{causal minimization} in the presence of frame assumptions and ramification constraints, either using circumscription or predicate completion, preserves this notion of $\mathcal{C}$-Consistency for {\small$\Sigma_{space}$} within a general class of action theories. Details are unessential here.}

\subsection{Practical Abduction in GIS with $\Sigma_{space}$}

%\subsubsection*{The Abductive Approach of Shanahan}
Let $\Sigma$ be the background theory and $\Phi$ be an observation sentence whose assimilation demands some explanation. According to the abductive approach to computing explanations, the task of assimilating $\Phi$ involves finding formulae $\Delta$ that when conjoined to $\Sigma$ yield $\Phi$ as a logical consequence (i.e., $\Sigma~\cup~\Delta~\models~\Phi$).  

Appendix A provides details of the precise abductive approach for computing explanations, as the details are not central for this paper. Instead, we focus on illustrating the nature of the high-level domain-independent abducibles that are generated as a result of the reasoning process in (A1--A3).

\subsubsection*{{\bf A1.}\quad \bf Abducing Appearances and Disappearances}
The following is with respect to the illustration in Fig. \ref{fig:narrative-branching}:\footnote{In (14-15), % (\ref{abd-app-disapp}--\ref{abd-exp}), 
$\Sigma_{change}$ corresponds to a general class of actions theories (e.g., in the manner described in \citep{bhatt:scc:08})  capable of handling the \emph{frame} and \emph{ramification} problems: general laws determining what does and does not change within a dynamically changing system. Details are not necessary to understand the result of the abduction methods.}

\begingroup
\everymath{\scriptstyle}
%\scriptsize
\begin{equation}\label{abd-app-disapp}%\small
\left\{
\begin{aligned}
&\Psi_{1}~\equiv~Holds(\phi_{top}(a,~c),~po,~t_{1})\\&\Psi_{2}~\equiv~Holds(\phi_{top}(a,~c),~ec,~t_{2})~\wedge~Holds(exists(b),~true,~t_{2})\\&~~~~~~~~~~~~\wedge~Holds(\phi_{top}(b,~a),~ntpp,~t_{2})\\&[\Sigma_{change}~\wedge~\Sigma_{space}~\wedge~\Psi_{1}~\wedge~\Delta_{ap-ds}]~\models~\Psi_{2}, where\\&\Delta_{1}~\equiv~(\exists~t_{i},~t_{j},~t_{k}).[~t_{1}~\leq~t_{i}~<~t_{2}~\wedge~Happens(appearance(b),~t_{i}    )]\\&\wedge~[~t_{i}~<~t_{j}~<~t_{2}~\wedge~Happens(tran(b,~a,~tpp),~t_{j})]~\wedge\\&[t_{k}~<~t_{2}~\wedge~Happens(tran(a,~c,~po),~t_{k})]~\wedge~[~t_{k}~\neq~t_{i}~\wedge~t_{k}~\neq~t_{j}]
\end{aligned}\right\}
\end{equation}
\endgroup

The derivation of $\Delta$ primarily involves non-monotonic reasoning in the form of minimising change (`$Caused$' and `$Happens$' predicates), in addition to making the usual default assumptions about inertia and indirect effects; the details are beyond the scope of this paper, and may be referred to in \citep{bhatt:scc:08}.

\subsubsection*{{\bf A2.}\quad\bf  Abducing Splits and Merges}
Below, $\Delta_{2}$ represents a subset of the minimal explanations that is derivable with respect to the observations in $\Psi_{e}$ and $\Psi_{f}$:

%\footnote{For instance, the a disappearance event is omitted since its treatment is similar to (14).} %\eqref{abd-app-disapp} 

\begingroup
\everymath{\scriptstyle}
%\scriptsize
\begin{equation}\label{abd-exp}%\small
\left\{
\begin{aligned}
&\Psi_{ini}~\equiv~\Psi_{e}~\equiv~[~Holds(\exts{[tl_{2},~iz_{3}]},~true,~t_{5})~~\wedge~Holds(\phi_{top}(\msf{[ez,~wz]},~\msf{iz_{3}}),~po,~t_{4})]\\&\Psi_{f}~\equiv~[~Holds(\exts{rz_{new}},~true,~t_{6})~\wedge~Holds(\exts{[rz_{1},~rz_{3},~mg_{3},~mg_{1}]},~false,~t_{6})~\wedge\\&Holds(\phi_{top}(\msf{rz_{2},~prk_{1}}),~ec,~t_{6})]\\&[\Sigma_{change}~\wedge~\Sigma_{space}~\wedge~\Psi_{e}~\wedge~\Delta]~\models~\Psi_{f}, where\\&\Delta_{2}~\equiv~(\exists~t_{i},~t_{j},~t_{k}).[~(t_{i},~t_{j})~<~t_{6}~\wedge~t_{i}~\neq~t_{j}~\wedge~Happens(disappearance(\msf{rz_{1}}),~t_{i})~\wedge\\&Happens(disappearance(\msf{rz_{3}}),~t_{j})]~\wedge~[t_{k}~\leq~t_{6}~\wedge~Happens(appearance(\msf{rz_{new}}),~t_{k})~]~\wedge\\&[~t_{k}~>~(~t_{i},~t_{j}~)~\wedge~Happens(merge([\msf{rz_{1},~rz_{3}}],~\msf{rz_{new}}),~t_{k})~]
\end{aligned}\right\}
\end{equation}
\normalsize%
\endgroup

\noindent In a manner methodologically similar to the case of geospatial events characterised so far, events such as \emph{growth}, \emph{shrinkage}, and basic \emph{transformation} and \emph{cloning} events may be subjected to particular concrete interpretations as well.

\noindent As a next step in the reasoning process, we turn to the issue of extracting high-level, domain-specific knowledge from the result of the scenario and narrative completion task.

%\begin{equation}\label{abd-exp}\small
%\left\{
%\begin{aligned}
%&[\Sigma_{change}~\wedge~\Sigma_{space}~\wedge~\Psi_{e}~\wedge~\Delta]~\models~\Psi_{f}, where\\&\Delta~\equiv~(\exists~t_{i},~t_{j},~t_{k}).[~t_{5}~\leq~t_{i}~<~t_{6}~\wedge~Happens(disappearance(\msf{rz_{1}}),~t_{i})]\\&\wedge~[~t_{j}~<~t_{6}~\wedge~Happens(merge([\msf{rz_{1},~rz_{3}}],~\msf{rz_{new}}),~t_{j})]~\wedge\\&[t_{k}~<~t_{2}~\wedge~Happens(appearance(\msf{rz_{new}}),~t_{k})~]~\wedge~[~t_{k}~\neq~t_{i}~\wedge~t_{k}~\neq~t_{j}]
%\end{aligned}\right\}
%\end{equation}
%\normalsize%

\subsubsection*{{\bf A3.}\quad\bf  Inferring High-Level (Domain-Dependent) Urbanization Processes}
The discussion so far focused on the domain-independent machinery needed to qualitatively represent and reason about certain aspects of dynamic (geo)spatial phenomena. Now, we turn to the urban dynamic domain, which is the focus of our running example. High-level urbanisation processes (e.g., natural, human, economic) in this domain may be characterised via a combination of low-level domain-independent qualitative spatial changes and geospatial events identifiable as per a certain event taxonomy. In the domain of Example \ref{exp:urban}, these correspond to urbanisation processes such as:  deforestation, migration, urban / rural re(construction) and relocation, industrialisation and infrastructure development, etc. Given the primary / domain-independent scenario and narrative completions (obtained by abduction) in [~$\Delta_{i},~\Delta_{j},\ldots,\Delta_{n}~$], high-level abducibles may be used in a domain-specific manner to infer the processes of interest (e.g., these abducibles may be constructed within standard query-based environment over a conventional GIS data set). For instance, \emph{high-level abducibles} (16)
%\eqref{high-abducibles} 
(referring to high-level processes) may be inferred given the primary abductions in (14--15):

%\begin{figure*}
\begingroup
\everymath{\scriptstyle}
%\scriptsize
\begin{equation}\label{high-abducibles}%\nonumber
\left[
\begin{aligned}
geosp&atial\_process({\color{blue!85!black}rural\_expansion}, t,~t^{\prime})~\longleftarrow~\\&[~(\exists~\msf{rz_{j}},\ldots,\msf{rz_{m}},~\msf{rz_{n}}).~RuralZones([~\msf{rz_{j}},\ldots,\msf{rz_{m}},~\msf{rz_{n}}~])~]~\wedge\\&\{[~(\exists~t_{i},).~during(t_{i},~t,~t^{\prime})~\wedge~Happens(merge([~\msf{rz_{j}},\ldots,\msf{rz_{m}}~],~\msf{rz_{n}}),~t_{i})~]&\mymodels~{\color{red!85}\Delta_{1}}\}\\geosp&atial\_process({\color{blue!85!black}mangrove\_deforestation},~t,~t^{\prime})~\longleftarrow~\\&[~(\exists~\msf{mg}).~MangroveZones(\msf{mg})~]~\wedge\\&\{[~(\exists~t_{i}).~during(t_{i},~t,~t^{\prime})~\wedge~Happens(shrinkage(\msf{mg}),~t_{i})~\vee\\&~~~~~~~~~~~~Happens(disappearance(\msf{mg}),~t_{i})~]&\mymodels~{\color{red!85}\Delta_{2}}\}\\&\vdots\\geosp&atial\_process({\color{blue!85!black}park\_encroachment},~t,~t^{\prime})~\longleftarrow~\\&[~(\exists~\msf{rz},~\msf{prk}).~RuralZone(\msf{rz})~\wedge~Park(\msf{prk})~]~\wedge\\&\{[~(\exists~t_{i}).~during(t_{i},~t,~t^{\prime})~\wedge~Happens(trans(rz,~prk,~[overlap,~inside]),~t_{i})~]&\mymodels~{\color{red!85}\Delta_{n}}\}
\end{aligned}
\right]
\end{equation}
\endgroup
%\end{figure*}

\noindent The set of high-level (domain-dependent) abducibles may be either dynamically constructed within a query-based environment, or may be pre-specified and invoked via some interfacing mechanism that connects the analytical capability with the real stake-holders in the analytical process. This would enable users and software services that utilise the narrative-based GIS architecture to independently define the semantics of the spatio-temporal phenomena in domain-specific ways.

For instance (building on the argument provided by one of the referees of this paper), a high-level domain-specific abducible could characterize the social and economic forces (e.g., \emph{gentrification, industrial agglomeration}) that drove such spatial expressions of urban development to occur per se. Such linking of high-level complex and / or subjectively interpreted geographic processes such as \emph{industrial agglomeration} to spatial-temporal data that capture readily observable properties (e.g., via \emph{satellite imagery} and \emph{land use}) depends on problem-specific considerations:

\begin{itemize}
	
	\item An analyst may decide to completely correlate observable spatio-temporal processes (e.g., \emph{shrinkage, splits, disappearance}) with complex socio-spatial phenomena such as \emph{urbanisation}.
	
	\item Spatio-temporal analysis (e.g., continual growth or shrinkage of a polygon) may be complemented with other data sources, and the influence of non-spatial datasets and quantitative analytical methods could be formally accounted for in the narrative framework such that the abductive explanation framework consists of both spatio-temporal as well as other kinds of abducibles (i.e., non-spatial evidences can be used to further enrich the interpretation of macro geospatial processes).
	
\end{itemize}

As discussed already in the paper, the focus of the narrative-centred model of this paper has been on the spatio-temporal aspects of the dynamic geospatial phenomena. A formal treatment of incorporating non-spatial datasets as evidences in the explanation process, albeit possible, is beyond the scope of this paper. Our focus has been on employing formal methods from the field of commonsense reasoning about space, actions, and change into the domain of dynamic GIS.

%but an abducible that characterizes the social and economic forces (e.g. gentrification, industrial agglomeration, etc.) that drove such spatial expressions of urban development to occur. How do we link hard-todefine and subjectively interpreted geographic processes such as industrial agglomeration to spatial-temporal data that capture readily observable properties such as land use, so that we can use GIS to effectively explore and reason about ÔmacroÕ processes such as urban development?"

\section{\textsc{Discussion and Conclusion}}\label{sec:disc-outlook}

The ability of semantic and qualitative analytical capability to complement and synergize with statistical and quantitiatively-driven methods has been recognised to be important within and beyond the range of GIS application domains (discussed in Section \ref{sec:application-areas}). Researchers in GIS and spatial information theory have investigated several fundamental ontological aspects concerning the modelling of events, processes, the practical development of taxonomies of events relevant to a geospatial context, and construction of formal methods in qualitative spatial information theory. 

As we have emphasised, event and object based explanatory analysis is especially important (e.g., in the context of a query-based GIS system) where the available data needs to be analysed for various purposes such as managerial decision making, policy formation and so forth. Indeed, the development of high-level analytical capability within the emerging \key{object}, \key{temporal} and \key{event-based} geographic information systems has been identified to present a range of fundamental representational and computational challenges -- it has been the objective of this paper to:
%\commentJOW{I would just make these normal sentences instead of using the quote environment, but that's just me}

\begin{quote} 
explicitly address some of these challenges from the viewpoint of the application of formal knowledge representation and reasoning methods concerning \emph{space, events, actions, and change}.
\end{quote}

The broad technical question that has been addressed in this paper is:

\begin{quote} 
what is it that constitutes the core spatial informatics underlying (specific kinds) of analytical capability within a range of dynamic geospatial domains?
\end{quote}

From a methodological viewpoint, the concrete goal of our research has been to: 

\begin{quote} 

investigate the theoretical foundations necessary to develop the computational capability for {high-level commonsense, qualitative} analysis of dynamic geospatial phenomena within next generation event and object based GIS systems. 
\end{quote}

We have presented an overarching framework for narrative-centred high-level modelling and explanatory analyses in the geospatial domain, and have provided a unified view of a consolidated architecture in the backdrop of an illustrated application scenario from the domain of urban dynamics. Building on existing foundations in the GIS community, and spatial information theory in particular, we have demonstrated fundamental challenges and presented solutions thereof encompassing aspects such as  \emph{qualitative abstraction and integration}, \emph{spatial consistency}, and \emph{practical geospatial abduction} within a logical setting. 

Most importantly, we believe that we have developed inroads from classical Knowledge Representation and Reasoning (KR) sub-disciplines in artificial intelligence, specifically formal methods in spatial and temporal reasoning, reasoning about action and change, and commonsense reasoning. We believe that these interdisciplinary inroads in GIScience open-up interesting possibilities toward the realization of next-generation analytical GIS software systems. From a topical viewpoint, we propose that this particularly demands a transdisciplinary scientific perspective that brings together Geography, Artificial Intelligence, and Cognitive Science.

%\textbf{CLP(QS)}.\quad CLP(QS) is a practically usable \emph{declarative spatial reasoning} system implemented as a general library within the context of Constraint Logic Programming (CLP) \citep{bhatt-et-al-2011,DBLP:conf/ecai/SchultzB12}. CLP(QS) is capable of modelling and reasoning about qualitative spatial relations pertaining to multiple spatial domains, i.e., one or more aspects of space such as \emph{topology}, and intrinsic and extrinsic  \emph{orientation}, \emph{distance}. Furthermore, users / application developers may freely mix object domains (i.e., \emph{points, line-segments}, and \emph{regions}) and with the available spatial domains. CLP(QS) also offers mixed geometric-qualitative spatial reasoning capabilities, and in its current form, a limited  \emph{quantification} support offering the means to go back from qualitative relations to the domain of precise quantitative information.
%
%CLP(QS) implements the semantics of qualitative spatial calculi in a manner that makes it possible to use spatial entities and relations between them as native first-class objects in a logic programming context. Furthermore it provides a declarative interface to qualitative and geometric spatial representation and reasoning capabilities such that these may be integrated with general knowledge representation and reasoning (KR) frameworks in artificial intelligence in particular, and more generally, within large scale spatial information systems, and cognitive systems and assistive technologies \citep{Bhatt-Schultz-Freksa:2013}. 

\newpage

\section*{\textsc{Acknowledgements}}

We would like to thank the anonymous reviewers for their valuable comments and feedback.

\smallskip

Mehul Bhattt gratefully acknowledges the funding and support of the German Research Foundation (DFG), {\upshape\footnotesize\texttt{www.dfg.de}}, via the Spatial Cognition Research Center (SFB/TR 8), and the SFB/TR 8 Project DesignSpace.

\section*{APPENDIX A}

\medskip

\subsection*{\bf Computing Explanations -- Logic-based Abduction in GIS}

%\subsubsection*{The Abductive Approach of Shanahan}
Let $\Sigma$ be a background theory and $\Phi$ be an observation sentence whose assimilation demands some explanation. According to the abductive approach, the task of assimilating $\Phi$ involves finding formulae $\Delta$ that when conjoined to $\Sigma$ yield $\Phi$ as a logical consequence (i.e., $\Sigma~\cup~\Delta~\models~\Phi$).  Additionally, a set of predicates are distinguished as being \emph{abducible} in order to avoid trivial explanations. It is essential that the explanation $\Delta$ must be in terms of predicates that have been designated as being \emph{abducible}. Finally, an approach is needed to incorporate the non-effects, and indirect effects of events and actions thereby overcoming the frame and ramification problems. This is achieved by the use of a relevant minimisation policy, which typically involves the use of \emph{circumscription} ($\mathsf{CIRC}$) \citep{McCarthy:1980:circumscription}. Furthermore, it is also necessary that the explanation be minimal, i.e., the derived explanation should not be subsumed by other explanations. Definition \eqref{def:explanation-gen} formalises the commonly-understood notion of explanation by logical abduction \citep{Shanahan:1989:pred-explana}.

\begin{Definition}[Explanation]\label{def:explanation-gen}{\upshape
A formula $\Delta$, essentially an existential statement, is an explanation of a ground observation sentence $\Phi_{obs}$ of language $\mathcal{L}$ in terms of the abduction policy $\eta^{*}$  given a background theory $[\Sigma~\equiv~\Sigma_{change}~\cup~\Sigma_{space}]$ and a circumscription policy that minimizes $\rho^{*}$ and allows $\sigma^{*}$ to vary if:

\begin{itemize}\upshape
	\item $\mathsf{CIRC}[\Sigma~\wedge~\Delta~;~\rho^{*}~;~\sigma^{*}]$ is consistent, and the models themselves are $\mathcal{C}$-Consistent (as per Definition \ref{def:composition-consistency})
	
	\item $\Delta$ mentions only predicates in $\eta^{*}$, and
	
	\item $\mathsf{CIRC}[\Sigma~\wedge~\Delta~;~\rho^{*}~;~\sigma^{*}]~\models~\Phi_{obs}$
	
%	\item $\Delta$ is an explanation of $\Phi$, and
	
	\item There is no explanation $\Delta^{'}$ of $\Phi_{obs}$ such that $\Delta~\models~\Delta^{'}$ and $\Delta^{'}~\nvDash~\Delta$ (i.e., the minimality criteria)
\end{itemize}
\eod
}
\end{Definition}

\newpage

\bibsep6pt

\renewcommand{\bibname}{ References} %for books
\renewcommand\refname{ References} %for articles

\bibliographystyle{abbrvnat}
%\bibliography{bibs/narrative-bib,bibs/clpqs-refs,bibs/QSR-bib,bibs/collab-literature,bibs/GIS-SpatialReasoning}

\begin{thebibliography}{0}
\providecommand{\natexlab}[1]{#1}
\providecommand{\url}[1]{\texttt{#1}}
\expandafter\ifx\csname urlstyle\endcsname\relax
  \providecommand{\doi}[1]{doi: #1}\else
  \providecommand{\doi}{doi: \begingroup \urlstyle{rm}\Url}\fi

\end{thebibliography}


\begin{thebibliography}{72}
\providecommand{\natexlab}[1]{#1}
\providecommand{\url}[1]{\texttt{#1}}
\expandafter\ifx\csname urlstyle\endcsname\relax
  \providecommand{\doi}[1]{doi: #1}\else
  \providecommand{\doi}{doi: \begingroup \urlstyle{rm}\Url}\fi

\bibitem[Abdelmoty et~al.(2009)Abdelmoty, Smart, El-Geresy, and
  Jones]{Abdelmoty_09_Supporting}
A.~I. Abdelmoty, P.~Smart, B.~El-Geresy, and C.~Jones.
\newblock Supporting frameworks for the geospatial semantic web.
\newblock In \emph{Proceedings of the 11th International Symposium on Advances
  in Spatial and Temporal Databases}, pages 355 -- 372, 2009.

\bibitem[Alchourron et~al.(1985)Alchourron, G{\"a}rdenfors, and
  Makinson]{Alchourron_Gaerdenfors_Makinson_85_On}
C.~E. Alchourron, P.~G{\"a}rdenfors, and D.~Makinson.
\newblock On the logic of theory change: Partial meet contraction and revision
  functions.
\newblock \emph{The Journal of Symbolic Logic}, 50\penalty0 (2):\penalty0
  510--530, 1985.

\bibitem[Allen(1983)]{IntervalCalc:Allen:1983}
J.~F. Allen.
\newblock Maintaining knowledge about temporal intervals.
\newblock \emph{Commun. ACM}, 26\penalty0 (11):\penalty0 832--843, 1983.
\newblock ISSN 0001-0782.

\bibitem[Barthes and Duisit(1975)]{Roland-1975-narrative-structural}
R.~Barthes and L.~Duisit.
\newblock {An Introduction to the Structural Analysis of Narrative}.
\newblock \emph{New Literary History}, 6\penalty0 (2):\penalty0 237--272, 1975.
\newblock ISSN 00286087.
\newblock \doi{10.2307/468419}.
\newblock URL \url{http://dx.doi.org/10.2307/468419}.

\bibitem[Beller(1991)]{Beller:1991:ST-Events}
A.~Beller.
\newblock Spatio/temporal events in a {GIS}.
\newblock In \emph{Proceedings of {GIS/LIS}}, pages 766--775. {ASPRS/ACSM},
  1991.

\bibitem[Bennett(2002)]{DBLP:conf/kr/Bennett02}
B.~Bennett.
\newblock Physical objects, identity and vagueness.
\newblock In D.~Fensel, F.~Giunchiglia, D.~L. McGuinness, and M.-A. Williams,
  editors, \emph{KR}, pages 395--408. Morgan Kaufmann, 2002.

\bibitem[Bhatt(2012)]{Bhatt:RSAC:2012}
M.~Bhatt.
\newblock Reasoning about space, actions and change: A paradigm for
  applications of spatial reasoning.
\newblock In \emph{Qualitative Spatial Representation and Reasoning: Trends and
  Future Directions}. IGI Global, USA, 2012.
\newblock ISBN ISBN13: 9781616928681.

\bibitem[Bhatt and Loke(2008)]{bhatt:scc:08}
M.~Bhatt and S.~Loke.
\newblock Modelling dynamic spatial systems in the situation calculus.
\newblock \emph{Spatial Cognition and Computation}, 8\penalty0 (1):\penalty0
  86--130, 2008.
\newblock ISSN 1387-5868.

\bibitem[Bhatt et~al.(2011{\natexlab{a}})Bhatt, Guesgen, W{\"o}lfl, and
  Hazarika]{bhatt2011-scc-trends}
M.~Bhatt, H.~Guesgen, S.~W{\"o}lfl, and S.~Hazarika.
\newblock Qualitative spatial and temporal reasoning: Emerging applications,
  trends, and directions.
\newblock \emph{Spatial Cognition \& Computation}, 11\penalty0 (1):\penalty0
  1--14, 2011{\natexlab{a}}.

\bibitem[Bhatt et~al.(2011{\natexlab{b}})Bhatt, Lee, and
  Schultz]{bhatt-et-al-2011}
M.~Bhatt, J.~H. Lee, and C.~Schultz.
\newblock {CLP(QS): A Declarative Spatial Reasoning Framework}.
\newblock In \emph{COSIT}, pages 210--230, 2011{\natexlab{b}}.

\bibitem[Claramunt and Th{\'e}riault(1995)]{Claramunt_Theriault_95_Managing}
C.~Claramunt and M.~Th{\'e}riault.
\newblock Managing time in {GIS}: {A}n event-oriented approach.
\newblock In J.~Clifford and A.~Tuzhilin, editors, \emph{Recent Advances on
  Temporal Databases}, pages 23--42. Springer, 1995.

\bibitem[Claramunt et~al.(1997)Claramunt, Th{\'e}riault, and
  Parent]{Claramunt_97_A}
C.~Claramunt, M.~Th{\'e}riault, and C.~Parent.
\newblock A qualitative representation of evolving spatial entities in
  two-dimensional spaces.
\newblock \emph{Innovations in GIS V, Carver, S. Ed.}, pages 119--129, 1997.

\bibitem[Clementini et~al.(1994)Clementini, Sharma, and
  Egenhofer]{Clementini_Sharma_Egenhofer_94_Modeling}
E.~Clementini, J.~Sharma, and M.~J. Egenhofer.
\newblock Modeling topological spatial relations : Strategies for query
  processing.
\newblock \emph{Computers and Graphics}, 18\penalty0 (92):\penalty0 815--822,
  1994.

\bibitem[Cockcroft(1997)]{springerlink:10.1023/A:1009754327059}
S.~Cockcroft.
\newblock A taxonomy of spatial data integrity constraints.
\newblock \emph{GeoInformatica}, 1:\penalty0 327--343, 1997.

\bibitem[Cohn and Renz(2007)]{Cohn_Renz_07_Qualitative}
A.~G. Cohn and J.~Renz.
\newblock Qualitative spatial reasoning.
\newblock In F.~van Harmelen, V.~Lifschitz, and B.~Porter, editors,
  \emph{Handbook of Knowledge Representation}. Elsevier, 2007.

\bibitem[Condotta et~al.(2006)Condotta, Saade, and
  Ligozat]{DBLP:conf/time/CondottaSL06}
J.-F. Condotta, M.~Saade, and G.~Ligozat.
\newblock A generic toolkit for n-ary qualitative temporal and spatial calculi.
\newblock In \emph{TIME}, pages 78--86. IEEE Computer Society, 2006.
\newblock ISBN 0-7695-2617-9.

\bibitem[Condotta et~al.(2008)Condotta, Kaci, and
  Schwind]{Condotta_Kaci_Schwind_08_A}
J.-F. Condotta, S.~Kaci, and N.~Schwind.
\newblock A framework for merging qualitative constraints networks.
\newblock In D.~Wilson and H.~C. Lane, editors, \emph{Proceedings of the
  Twenty-First International Florida Artificial Intelligence Research Society
  Conference, May 15-17, 2008, Coconut Grove, Florida, USA}, pages 586--591.
  AAAI Press, 2008.

\bibitem[Couclelis(2009)]{Couclelis-Cosit09}
H.~Couclelis.
\newblock The abduction of geographic information science: Transporting spatial
  reasoning to the realm of purpose and design.
\newblock In K.~S. Hornsby, C.~Claramunt, M.~Denis, and G.~Ligozat, editors,
  \emph{COSIT}, volume 5756 of \emph{Lecture Notes in Computer Science}, pages
  342--356. Springer, 2009.

\bibitem[Duckham et~al.(2010)Duckham, Jeong, Li, and
  Renz]{Duckham_10_Decentralized}
M.~Duckham, M.-H. Jeong, S.~Li, and J.~Renz.
\newblock Decentralized querying of topological relations between regions
  without using localization.
\newblock In \emph{Proc. 18th ACM SIGSPATIAL GIS}, pages 414--417, 2010.

\bibitem[Dylla and Wallgr{\"u}n(2007)]{Dylla_Wallgruen_07_Qualitative}
F.~Dylla and J.~O. Wallgr{\"u}n.
\newblock Qualitative spatial reasoning with conceptual neighborhoods for agent
  control.
\newblock \emph{Journal of Intelligent and Robotic Systems}, 48\penalty0
  (1):\penalty0 55--78, 2007.

\bibitem[Egenhofer and Mark(1995)]{citeulike:4160678}
M.~Egenhofer and D.~Mark.
\newblock {Naive Geography}.
\newblock In \emph{Spatial Information Theory A Theoretical Basis for GIS},
  pages 1--15. Springer Verlag, 1995.
\newblock \doi{10.1007/3-540-60392-1\_1}.
\newblock URL \url{http://dx.doi.org/10.1007/3-540-60392-1\_1}.

\bibitem[Egenhofer(1991)]{Egenhofer_91_Reasoning}
M.~J. Egenhofer.
\newblock Reasoning about binary topological relations.
\newblock In O.~Gunther and H.-J. Schek, editors, \emph{Advances in Spatial
  Databases, Second Symposium on Large Spatial Databases}, volume 525, pages
  143--160. Springer, 1991.
\newblock ISBN 3-540-54414-3.

\bibitem[Egenhofer and Al-Taha(1992)]{Egenhofer_Al-Taha_92_Reasoning}
M.~J. Egenhofer and K.~K. Al-Taha.
\newblock Reasoning about gradual changes of topological relationships.
\newblock In \emph{Proceedings of the International Conference GIS - From Space
  to Territory: Theories and Methods of Spatio-Temporal Reasoning on Theories
  and Methods of Spatio-Temporal Reasoning in Geographic Space}, pages
  196--219, London, UK, 1992. Springer-Verlag.

\bibitem[Egenhofer and Franzosa(1991)]{Egenhofer:1991:pointset}
M.~J. Egenhofer and R.~D. Franzosa.
\newblock {Point Set Topological Relations.}
\newblock \emph{International Journal of Geographical Information Systems},
  5\penalty0 (2):\penalty0 161--174, 1991.

\bibitem[Fagin and Vardi(1984)]{DBLP:conf/icalp/FaginV84}
R.~Fagin and M.~Y. Vardi.
\newblock The theory of data dependencies - an overview.
\newblock In J.~Paredaens, editor, \emph{ICALP}, volume 172 of \emph{Lecture
  Notes in Computer Science}, pages 1--22. Springer, 1984.

\bibitem[Fisher(1987)]{narrative-paradigm}
W.~R. Fisher.
\newblock \emph{Human communication as narration: Toward a philosophy of
  reason, value, and action}.
\newblock University of South Carolina Press, Columbia, SC, 1987.

\bibitem[Freksa(1991{\natexlab{a}})]{Freksa_91_Conceptual}
C.~Freksa.
\newblock Conceptual neighborhood and its role in temporal and spatial
  reasoning.
\newblock In M.~Singh and L.~Trav{\'e}-Massuy{\`e}s, editors, \emph{Decision
  Support Systems and Qualitative Reasoning}, pages 181 -- 187.
  1991{\natexlab{a}}.

\bibitem[Freksa(1991{\natexlab{b}})]{cosyFreksa1991a}
C.~Freksa.
\newblock Qualitative spatial reasoning.
\newblock In D.~Mark and A.~Frank, editors, \emph{Cognitive and linguistic
  aspects of geographic space}, pages 361--372. Kluwer, Dordrecht,
  1991{\natexlab{b}}.

\bibitem[Gabbay and Hunter(1991)]{DBLP:conf/fair/GabbayH91}
D.~M. Gabbay and A.~Hunter.
\newblock Making inconsistency respectable: a logical framework for
  inconsistency in reasoning.
\newblock In P.~Jorrand and J.~Kelemen, editors, \emph{FAIR}, volume 535 of
  \emph{Lecture Notes in Computer Science}, pages 19--32. Springer, 1991.
\newblock ISBN 3-540-54507-7.

\bibitem[Gahegan(1996)]{DBLP:journals/tgis/Gahegan96}
M.~Gahegan.
\newblock Specifying the transformations within and between geographic data
  models.
\newblock \emph{T. GIS}, 1\penalty0 (2):\penalty0 137--152, 1996.

\bibitem[Galton(2000)]{Galton00:qualitative}
A.~Galton.
\newblock \emph{{Q}ualitative spatial change}.
\newblock {S}patial information systems. Oxford Univ. Press, 2000.
\newblock ISBN 0198233973.

\bibitem[Galton and Hood(2004)]{galton04-interpolation}
A.~Galton and J.~Hood.
\newblock Qualitative interpolation for environmental knowledge representation.
\newblock In R.~L. de~M{\'a}ntaras and L.~Saitta, editors, \emph{ECAI}, pages
  1017--1018. IOS Press, 2004.
\newblock ISBN 1-58603-452-9.

\bibitem[Galton and Mizoguchi(2009)]{journals/ao/GaltonM09}
A.~Galton and R.~Mizoguchi.
\newblock The water falls but the waterfall does not fall: New perspectives on
  objects, processes and events.
\newblock \emph{Applied Ontology}, 4\penalty0 (2):\penalty0 71--107, 2009.

\bibitem[Goguen(2004)]{Goguen-course-compu-narratology}
J.~Goguen.
\newblock {CSE 87C Winter 2004 Freshman Seminar on Computational Narratology.
  }.
\newblock \emph{New Literary History}, 2004.
\newblock URL \url{http://cseweb.ucsd.edu/~goguen/courses/87w04/1.html}.

\bibitem[Gr{\'e}goire and Konieczny(2006)]{Gregoire_Konieczny_06_Logic}
{\'E}.~Gr{\'e}goire and S.~Konieczny.
\newblock Logic-based approaches to information fusion.
\newblock \emph{Information Fusion}, 7\penalty0 (1):\penalty0 4--18, 2006.

\bibitem[Grenon and Smith(2004)]{Grenon_Smith_04_SNAP}
P.~Grenon and B.~Smith.
\newblock Snap and span: Towards dynamic spatial ontology.
\newblock \emph{Spatial Cognition and Computation}, 4\penalty0 (1):\penalty0
  69--104, 2004.

\bibitem[Haarslev and M{\"o}ller(1997)]{DBLP:conf/dlog/HaarslevM97}
V.~Haarslev and R.~M{\"o}ller.
\newblock Spatioterminological reasoning: Subsumption based on geometrical
  inferences.
\newblock In R.~J. Brachman, F.~M. Donini, E.~Franconi, I.~Horrocks, A.~Y.
  Levy, and M.-C. Rousset, editors, \emph{Description Logics}, volume 410 of
  \emph{URA-CNRS}, 1997.

\bibitem[Herman et~al.(2005)Herman, Jahn, and Ryan]{narratology-marie-laure}
D.~Herman, M.~Jahn, and M.-L. Ryan.
\newblock \emph{Routledge Encyclopedia Of Narrative Theory}.
\newblock Routledge, Feb. 2005.
\newblock ISBN 0415282594.

\bibitem[Hornsby and Egenhofer(2000)]{Hornsby_Egenhofer_00_Identity}
K.~Hornsby and M.~J. Egenhofer.
\newblock Identity-based change: {A} foundation for spatio-temporal knowledge
  representation.
\newblock \emph{International Journal of Geographical Information Science},
  14\penalty0 (3):\penalty0 207--224, 2000.

\bibitem[Hornsby and Cole(2007)]{Stewart-Hornsby_Cole_07_Modeling}
K.~S. Hornsby and S.~J. Cole.
\newblock Modeling moving geospatial objects from an event-based perspective.
\newblock \emph{T. GIS}, 11\penalty0 (4):\penalty0 555--573, 2007.

\bibitem[Jiang and Worboys(2008)]{Jiang_Worboys_08_Detecting}
J.~Jiang and M.~Worboys.
\newblock Detecting basic topological changes in sensor networks by local
  aggregation.
\newblock In \emph{Proc. 16th ACM International Conference on Advances in
  Geographic Information Systems}, pages 1--10, 2008.

\bibitem[Khan and Schneider(2010)]{Kahn_Schneider_10_Topological}
A.~Khan and M.~Schneider.
\newblock Topological reasoning between complex regions in databases with
  frequent updates.
\newblock In \emph{18th ACM SIGSPATIAL Int. Conf. on Advances in Geographic
  Information Systems (ACM SIGSPATIAL GIS)}, pages 380--389, 2010.

\bibitem[Klippel et~al.(2008)Klippel, Worboys, and
  Duckham]{Klippel_Worboys_Duckham_08_Identifying}
A.~Klippel, M.~Worboys, and M.~Duckham.
\newblock Identifying factors of geographic event conceptualisation.
\newblock \emph{International Journal of Geographical Information Science},
  22\penalty0 (2):\penalty0 183--204, 2008.

\bibitem[Mackworth(1977)]{Mackworth_77_Consistency}
A.~Mackworth.
\newblock Consistency in networks of relations.
\newblock \emph{Artificial Intelligence}, 8\penalty0 (1):\penalty0 99--118,
  1977.

\bibitem[Mani(2012)]{CMN-Mani-2012}
I.~Mani.
\newblock Computational modeling of narrative.
\newblock \emph{Synthesis Lectures on Human Language Technologies}, 5\penalty0
  (3):\penalty0 1--142, 2012.

\bibitem[Mani(2013)]{Mani-comp-narratology}
I.~Mani.
\newblock {Computational Narratology}.
\newblock In P.~H\"{u}hn, J.~C. Meister, J.~Pier, and W.~Schmid, editors,
  \emph{{The Living Handbook of Narratology}}. Hamburg University Press, 2013.

\bibitem[McCarthy(1980)]{McCarthy:1980:circumscription}
J.~McCarthy.
\newblock Circumscription - a form of non-monotonic reasoning.
\newblock \emph{Artif. Intell.}, 13\penalty0 (1-2):\penalty0 27--39, 1980.

\bibitem[McCarthy(2000)]{McCarthy:2000:concepts-logical-AI}
J.~McCarthy.
\newblock Logic-based artificial intelligence.
\newblock chapter Concept of logical AI, pages 37--56. Kluwer Academic
  Publishers, Norwell, MA, USA, 2000.
\newblock ISBN 0-7923-7224-7.
\newblock URL \url{http://dl.acm.org/citation.cfm?id=566344.566348}.

\bibitem[McCarthy and Costello(1998)]{McCarthy-98-kr-combining-narrative}
J.~McCarthy and T.~Costello.
\newblock Combining narratives.
\newblock In \emph{KR}, pages 48--59, 1998.

\bibitem[Meister(2011)]{meister-narratology-hdbk-narrato}
J.~C. Meister.
\newblock {Narratology}.
\newblock In P.~H\"{u}hn, J.~C. Meister, J.~Pier, and W.~Schmid, editors,
  \emph{{The Living Handbook of Narratology}}. Hamburg University Press, 2011.

\bibitem[Mennis et~al.(2000)Mennis, Peuquet, and Qian]{Mennis00:conceptual}
J.~Mennis, D.~J. Peuquet, and L.~Qian.
\newblock {A} conceptual framework for incorporating cognitive principles into
  geographical database representation.
\newblock \emph{International Journal of Geographical Information Science},
  14\penalty0 (6):\penalty0 501--520, 2000.

\bibitem[Miller and Shanahan(1994)]{MillerS94}
R.~Miller and M.~Shanahan.
\newblock Narratives in the situation calculus.
\newblock \emph{J. Log. Comput.}, 4\penalty0 (5):\penalty0 513--530, 1994.

\bibitem[Mondo et~al.(2010)Mondo, Stell, Claramunt, and
  Thibaud]{journals/jucs/MondoSCT10}
G.~D. Mondo, J.~G. Stell, C.~Claramunt, and R.~Thibaud.
\newblock A graph model for spatio-temporal evolution.
\newblock \emph{J. UCS}, 16\penalty0 (11):\penalty0 1452--1477, 2010.

\bibitem[Mueller(2007)]{DBLP:journals/lalc/Mueller07}
E.~T. Mueller.
\newblock Modelling space and time in narratives about restaurants.
\newblock \emph{LLC}, 22\penalty0 (1):\penalty0 67--84, 2007.

\bibitem[{NIMA}(2000)]{NIMA:2000:GIS-Vision}
{NIMA}.
\newblock {National Imagery and Mapping Agency, The Big Idea Framework}, 2000.

\bibitem[Peuquet(1988)]{5906}
D.~J. Peuquet.
\newblock Representations of geographic space: Toward a conceptual synthesis.
\newblock \emph{Annals of the Association of American GeographersAnnals of the
  Association of American Geographers}, 78\penalty0 (3):\penalty0 373--394,
  1988.

\bibitem[Pinto(1998)]{OccurNarraSC:Pinto:1998}
J.~Pinto.
\newblock Occurrences and narratives as constraints in the branching structure
  of the situation calculus.
\newblock \emph{J. Log. Comput.}, 8\penalty0 (6):\penalty0 777--808, 1998.

\bibitem[Prince(1982)]{narratolog-prince-1982}
G.~Prince.
\newblock \emph{{Narratology: The Form and Function of Narrative}}.
\newblock Mouton, 1982.

\bibitem[Randell et~al.(1992)Randell, Cui, and Cohn]{Randell_Cui_Cohn_92_A}
D.~A. Randell, Z.~Cui, and A.~Cohn.
\newblock A spatial logic based on regions and connection.
\newblock In \emph{Principles of Knowledge Representation and Reasoning:
  Proceedings of the Third International Conference}, pages 165--176. Morgan
  Kaufmann, 1992.

\bibitem[Renolen(January 2000)]{Renolen_00_Modelling}
A.~Renolen.
\newblock Modelling the real world: Conceptual modelling in spatiotemporal
  information system design.
\newblock \emph{Transactions in GIS}, 4:\penalty0 23--42(20), January 2000.

\bibitem[Renz and Nebel(2007)]{renz-nebel-hdbk07}
J.~Renz and B.~Nebel.
\newblock Qualitative spatial reasoning using constraint calculi.
\newblock In \emph{Handbook of Spatial Logics}, pages 161--215. 2007.

\bibitem[Riessman(1993)]{narrat-analysis}
C.~K. Riessman.
\newblock \emph{{Narrative Analysis}}.
\newblock Newbury Park: Sage Publications, 1993.

\bibitem[Rodr\'{\i}guez et~al.(2010)Rodr\'{\i}guez, Brisaboa, Meza, and
  Luaces]{Rodriguez:2010:MCR:1869790.1869818}
M.~A. Rodr\'{\i}guez, N.~Brisaboa, J.~Meza, and M.~R. Luaces.
\newblock Measuring consistency with respect to topological dependency
  constraints.
\newblock In \emph{Proceedings of the 18th SIGSPATIAL International Conference
  on Advances in Geographic Information Systems}, pages 182--191, 2010.

\bibitem[Schultz and Bhatt(2012)]{DBLP:conf/ecai/SchultzB12}
C.~Schultz and M.~Bhatt.
\newblock {Towards a Declarative Spatial Reasoning System}.
\newblock In \emph{ECAI 2012}, pages 925--926, 2012.

\bibitem[Shanahan(1989)]{Shanahan:1989:pred-explana}
M.~Shanahan.
\newblock Prediction is deduction but explanation is abduction.
\newblock In \emph{IJCAI}, pages 1055--1060, 1989.

\bibitem[Wallgr{\"u}n et~al.(2007)Wallgr{\"u}n, Frommberger, Wolter, Dylla, and
  Freksa]{cosy:sparq-sc06}
J.~O. Wallgr{\"u}n, L.~Frommberger, D.~Wolter, F.~Dylla, and C.~Freksa.
\newblock Qualitative spatial representation and reasoning in the
  sparq-toolbox.
\newblock In T.~Barkowsky, M.~Knauff, G.~Ligozat, and D.~Montello, editors,
  \emph{Spatial Cognition V: Reasoning, Action, Interaction: International
  Conference Spatial Cognition 2006}, volume 4387 of \emph{LNCS}, pages 39--58.
  Springer-Verlag Berlin Heidelberg, 2007.

\bibitem[Westphal et~al.(2009)Westphal, Woelfl, and Gantner]{GQR:2009:AAAI-Sym}
M.~Westphal, S.~Woelfl, and Z.~Gantner.
\newblock {GQR}: {A} fast solver for binary qualitative constraint networks.
\newblock In \emph{AAAI Spring Symposium on Benchmarking of Qualitative Spatial
  and Temporal Reasoning Systems}, 2009.

\bibitem[Worboys(2005)]{journals/gis/Worboys05}
M.~F. Worboys.
\newblock Event-oriented approaches to geographic phenomena.
\newblock \emph{International Journal of Geographical Information Science},
  19\penalty0 (1):\penalty0 1--28, 2005.

\bibitem[Worboys and Hornsby(2004)]{DBLP:conf/giscience/WorboysH04}
M.~F. Worboys and K.~Hornsby.
\newblock From objects to events: Gem, the geospatial event model.
\newblock In M.~J. Egenhofer, C.~Freksa, and H.~J. Miller, editors,
  \emph{GIScience}, volume 3234 of \emph{Lecture Notes in Computer Science},
  pages 327--344. Springer, 2004.

\bibitem[Yuan(2001)]{citeulike:10114141}
M.~Yuan.
\newblock {Representing Complex Geographic Phenomena in GIS}.
\newblock \emph{Cartography and Geographic Information Science}, pages 83--96,
  Apr. 2001.
\newblock ISSN 1523-0406.

\bibitem[Yuan and Hornsby(2008)]{DBLP:books/daglib/0020516}
M.~Yuan and K.~Hornsby.
\newblock \emph{Computation and visualization for understanding dynamics in
  geographic domains - a research agenda}.
\newblock CRC Press, 2008.
\newblock ISBN 978-1-4200-6032-4.

\bibitem[Yuan et~al.(2004)Yuan, Mark, Egenhofer, and
  Peuquet]{RschAgendaUCGIS:2004}
M.~Yuan, D.~M. Mark, M.~J. Egenhofer, and D.~J. Peuquet.
\newblock \emph{Chapter 5, Extensions to Geographic Representations in `A
  Research Agenda for Geographic Information Science'}.
\newblock CRC Press, 2004.
\newblock ISBN 0849327288.

\end{thebibliography}

\end{document}